%% file: main.tex
\documentclass[runningheads]{llncs}

\usepackage{eccv}

\usepackage{svg}
\usepackage{eccvabbrv}

\usepackage{graphicx}
\usepackage{booktabs}
\usepackage[font=small]{caption}

\usepackage{floatrow}
\newfloatcommand{capbtabbox}{table}[][\FBwidth]
\usepackage[accsupp]{axessibility}  %

\usepackage[pagebackref,breaklinks,colorlinks]{hyperref}

\usepackage{orcidlink}

\begin{document}

\title{SuperLoRA: Parameter-Efficient Unified Adaptation of Multi-Layer Attention Modules} 

\titlerunning{SuperLoRA}

\author{Xiangyu Chen\inst{1,2}\orcidlink{0000-0002-9690-0067} \and
Jing Liu\inst{2} \orcidlink{0000-0002-1712-2966}\and
Ye Wang\inst{2}\orcidlink{0000-0001-5220-1830} \and
Pu (Perry) Wang\inst{2}\orcidlink{0000-0002-4718-3102} \and
Matthew Brand\inst{2} \and
Guanghui Wang\inst{3}\orcidlink{0000-0003-3182-104X} \and
Toshiaki Koike-Akino\inst{2}\orcidlink{0000-0002-2578-5372}
}

\authorrunning{X.~Chen et al.}

\institute{
University of Kansas, Lawrence, KS 66045, USA\\
\email{xychen@ku.edu}\\
 \and
Mitsubishi Electric Research Laboratories (MERL), Cambridge, MA 02139, USA\\
\email{\{xiachen, jiliu, yewang, pwang, brand, koike\}@merl.com}
\and
Toronto Metropolitan University,
 Toronto, ON M5B 2K3, Canada\\
\email{wangcs@torontomu.ca}
}

\maketitle

\begin{abstract}
Low-rank adaptation (LoRA) and its variants are widely employed in fine-tuning large models, including large language models for natural language processing and diffusion models for computer vision. 
  This paper proposes a generalized framework called SuperLoRA that unifies and extends different LoRA variants, which can be realized under different hyper-parameter settings. 
Introducing grouping, folding, shuffling, projecting, and tensor factoring, SuperLoRA offers high flexibility compared with other LoRA variants and demonstrates superior performance for transfer learning tasks especially in the extremely few-parameter regimes. 
 \and Parameter-efficient fine-tuning \and Tensor rank decomposition
\end{abstract}

\section{Introduction}
Large neural network models are dominating machine learning recently with the emergence of exceptional models, such as Vision Transformer (ViT)~\cite{dosovitskiy2020image}, ConvNeXt~\cite{liu2022convnet} and Stable Diffusion~\cite{ho2020denoising} for vision tasks, and large language models (LLMs) including GPT~\cite{achiam2023gpt}, PALM2~\cite{anil2023palm}, Gemini~\cite{team2023gemini} and LLaMA2~\cite{touvron2023llama} for natural language processing (NLP). 
However, the increased resource consumption and data requirement along with model size limits its generalization on downstream tasks. 
To solve this, Parameter-Efficient Fine-Tuning (PEFT) has been widely explored to fine-tune less parameters while retaining high performance. 
Among this, adapter-based technique like LoRA (Low-Rank Adaptation)~\cite{hu2021lora} demonstrates superiority in its convenience of plug-and-play nature. 

\begin{figure}[t]
    \centering 
    \includegraphics[width=\linewidth]{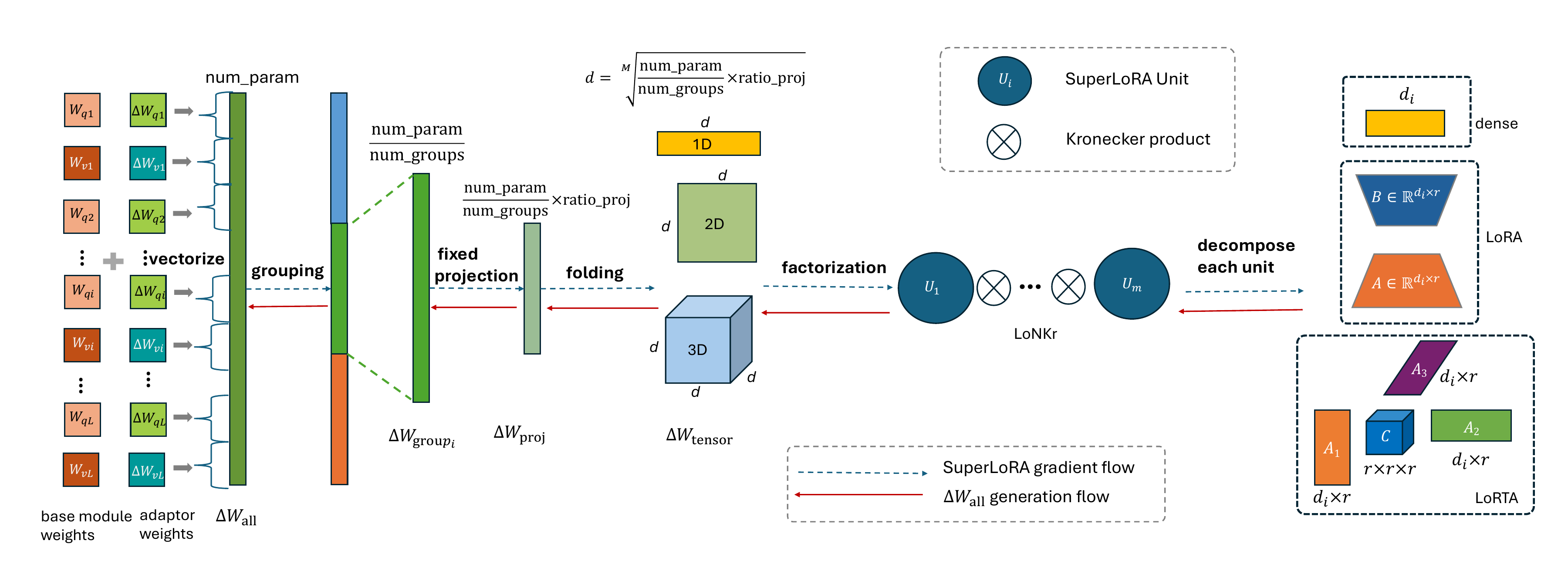}
    \caption{Schematic of SuperLoRA to fine-tune multi-layer attention modules at once with vectorizing, grouping, projection, folding, and factorization.}
    \label{fig:superlora}
\end{figure}

\begin{figure}[t]
\begin{floatrow}
\capbtabbox{%
    \begin{tabular}{c|l} \hline\toprule 
    notation & description \\ \hline
    $r$ & rank of factorization \\
    $\mathcal{F}$ & mapping function \\
    $\rho$ & compression ratio \\
    $G$ & number of groups \\
    $M$ & order of tensor modes \\
    $K$ & number of splits \\
    \bottomrule
    \end{tabular}
}{
    \caption{Hyperparameters and notation.}
    \label{tb:notation}
    }
\ffigbox{%
  \includegraphics[width=0.99\linewidth]{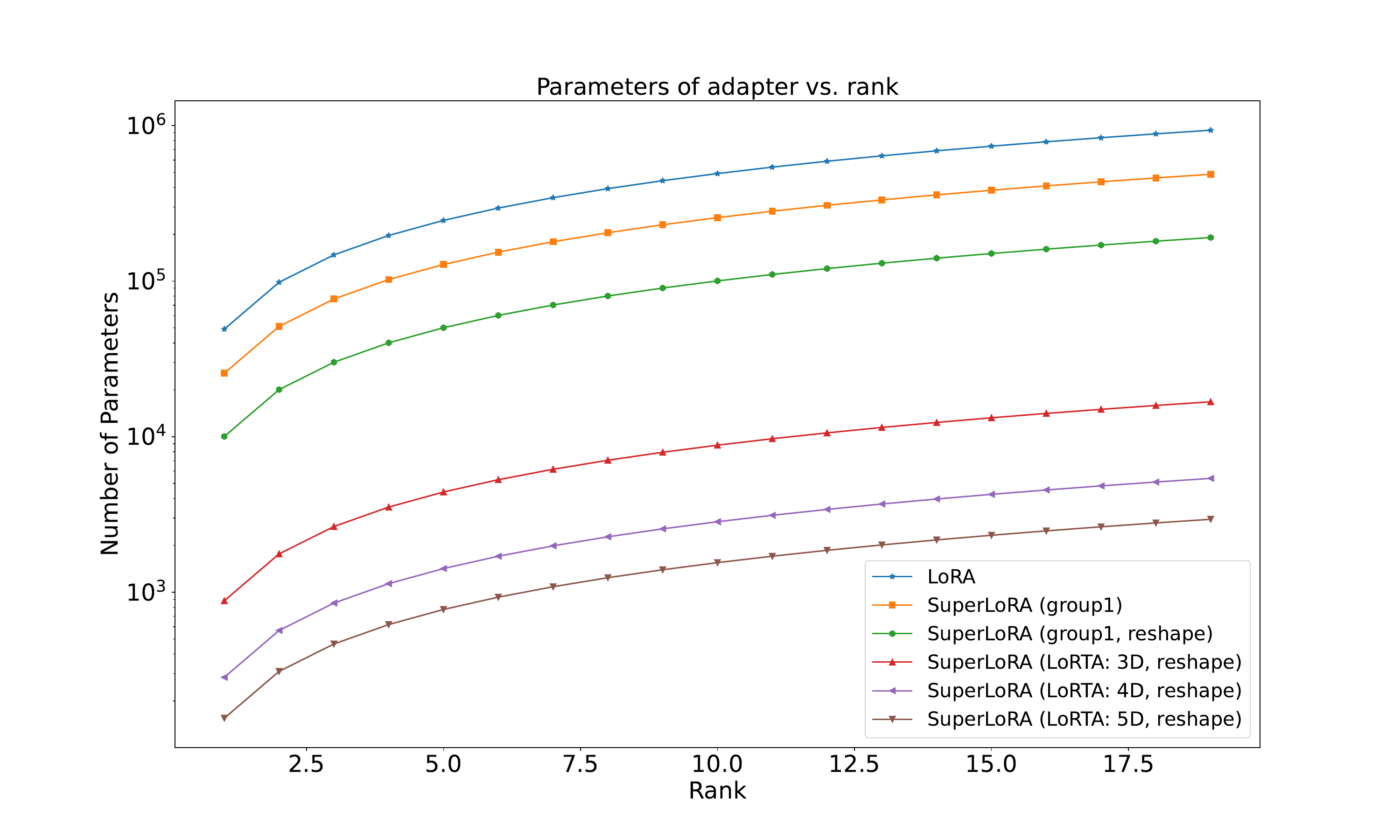}
}{%
  \caption{Required number of parameters.}%
  \label{fig:nparams}
}
\end{floatrow}
\end{figure}

LoRA~\cite{hu2021lora} approximates the weight updates of the base model by approximating the change $\Delta W$ of each weight matrix as the product of two low-rank matrices. 
This decreases the required parameters from $d^2$ to $2rd$ when $r\ll d$, where $d$ and $r$ are weight size and the rank, respectively. 
Most LoRA variants work on solving the inherent \textit{low-rank constraint} of matrix factorization, including LoHA (Low-rank Hadamard)~\cite{yeh2024navigating}, LoKr (Low-rank Kronecker)~\cite{yeh2024navigating}, and LoTR (Low Tensor Rank)~\cite{bershatsky2024lotr}. 
However, we find these variants can be unified within our framework---SuperLoRA---with different hyper-parameters as shown in~\Cref{tb:superlora}. 
Our proposed SuperLoRA framework is depicted in \Cref{fig:superlora}, which also yields to some new variants: LoNKr (\textbf{N}-split version of LoKr) and LoRTA (\textbf{T}ensor version LoRA). 
Additionally, we introduce two extended options, 1) reshaping $\Delta W$ to any arbitrary multi-dimensional tensor arrays before applying LoRA variants, and 2) splitting all $\Delta W$ into an arbitrary number of groups, which breaks the boundaries for $\Delta W$ across different weights.
Moreover, to further compress the number of trainable parameters, a projection layer $\mathcal{F}$ with fixed parameters is inserted to map $\Delta W_\mathrm{lora}$ to the actual $\Delta W$. 
Accordingly, SuperLoRA provides more flexibility and extended functionality, controlled by a set of hyper-parameters as shown in~\Cref{tb:notation}. 
The contributions of this paper are summarized as follows:
\begin{itemize}
\item We propose a new PEFT framework SuperLoRA which gracefully unifies and extends most LoRA variants.
\item With projected tensor rank decomposition, SuperLoRA can adapt all weights across layers jointly with a wide range of adjustable parameter amount.
\item We investigate the effect of tensor reshaping, grouping, random projection, and shuffling.
\item We demonstrate high parameter efficiency for large ViT and diffusion models in two transfer learning tasks: image classification and image generation.
\item Significantly reduced parameters by 3 to 10 folds can be achieved.
\end{itemize}

\begin{table}[t]
\caption{Hyper-parameter settings in SuperLoRA and the resultant LoRA variant}
\label{tb:superlora}
\centering
\begin{tabular}{l|l} \hline
\toprule
 hyper-parameters settings & method \\ \hline
 $\mathcal{F} = I$, weight-wise, $K = 1$, $C_{g1} = I$, $M = 1$, $A_{g11}\in \mathbb{R}^{d_\mathrm{in} {d_\mathrm{out} \times 1}}$  &  dense FT\\
$\mathcal{F} = I$, weight-wise, $K = 1$, $C_{g1} = I$, $M = 2$, $A_{g1m}\in \mathbb{R}^{d_m\times r}$   & LoRA~\cite{hu2021lora} \\
$\mathcal{F} = I$, weight-wise, $K = 2$, $C_{gk} = I$, $M = 2$, $A_{gkm}\in \mathbb{R}^{d_{m}\times r}$ & LoKr~\cite{yeh2024navigating} \\
$\mathcal{F} = I$, group-wise, $G=1$, $M>2$ & LoTR~\cite{bershatsky2024lotr} \\
$\mathcal{F} = I$, group-wise,  $K >2$,  $C_{gk} = I$, $M = 2$, $A_{gkm}\in \mathbb{R}^{d_{m}\times r}$& LoNKr \\
$\mathcal{F} = I$, group-wise, $K = 1$, $M > 2$, $A_{gkm}\in \mathbb{R}^{d_{m}\times r}$ & LoRTA \\
\bottomrule
\end{tabular}
\end{table}

\section{Related Work}
PEFT algorithms are widely explored in transfer learning tasks in both computer vision~\cite{jie2023revisiting,he2023parameter,jie2023fact} and NLP fields~\cite{li2021prefix,lester2021power,guo2021parameter,he2021towards,karimi2021compacter} as it not only saves memory and time at fine-tuning, but also requires much less data to fine-tune, making it feasible to borrow the capacity from large models in few-data tasks. 
Adapter-based methods~\cite{chen2022adaptformer,hao2022consolidator,houlsby2019parameter,pfeiffer2021adapterfusion}, that freeze the base model weights and fine-tune only the additional adapter parameters, stand out since their plug-and-play nature enables many downstream tasks to share the same large model, leaving the adapter to hold only the task-specific information. 
The widely used method LoRA~\cite{hu2021lora} and its extension~\cite{hayou2024lora+,zhu2024asymmetry} assume that the weight correction term can be estimated by low-rank decomposition under the low-dimensional manifold hypothesis.

Addressing the inherent \textit{low-rank constraint} of matrix factorization in LoRA, LoHA~\cite{yeh2024navigating} divides $\Delta W$ into two splits and combines them with Hadamard product, and KronA~\cite{edalati2022krona} combines the two splits with a Kronecker product to enlarge the overall rank. 
LoKr~\cite{yeh2024navigating}  further extended KronA to convolutional layers.
LoDA (Low-Dimensional Adaptation)~\cite{liu2023loda} extended LoRA by introducing nonlinearity.
Our SuperLoRA can nicely generalize and extend such variants. 

Instead of approximating weight-wise updates, LoTR~\cite{bershatsky2024lotr} jointly approximates all $\Delta W$ across the model with a careful handling to preserve the geometric meaning of each weight. 
Differently, SuperLoRA relaxes the geometrically meaningful boundaries by caring the total number of parameters and splitting it to any number of groups. 
For high-order tensor decomposition, LoTR employs more stringent Tensor Train Decomposition to deal with the core tensor explosion, while SuperLoRA coupled Tucker Decomposition with a fixed projection layer. 
Besides, their proposed methods are restricted to context when $\Delta W$ is the same high-order tensor, while with reshaping, SuperLoRA (LoRTA) can be applied to any weight shape. 

Most recent work~\cite{chen2024parameterefficient} decomposes each convolution kernel into a learnable filter atom and its non-learnable counterparts. 
The concept of filter atom is similar to the projection layer of SuperLoRA. 
However, it works on each convolutional kernels separately, resulting in a waste of parameters, while SuperLoRA considers the entire model jointly. 
Besides, the atom coefficients are obtained from matrix factorization, while SuperLoRA uses a fastfood projection~\cite{le2013fastfood}, which is faster, simpler and more theoretically justifiable to exploit intrinsic dimensionality~\cite{aghajanyan2020intrinsic}. 
In addition, SuperLoRA can control the size of atoms directly while atoms in their method are restricted in factorization.

Local LoRA~\cite{key2023local} aims to reduce memory consumption at fine-tuning by splitting large model into groups and then fine-tune group-by-group sequentially, but no adjustment on the LoRA structure was proposed. 
Instead, SuperLoRA focuses on how to split and assign LoRA for each group, which is a viable extension of Local LoRA.

\section{Methodology}
\subsection{Low-Rank Adaptation (LoRA)}
LoRA~\cite{hu2021lora} assumes the update $\Delta W$ of each weight matrix $W$ for fine-tuning can be approximated by low-rank mapping as $\Delta W = AB^\top$ ($[\cdot]^\top$ denotes matrix transpose), which is added to the frozen weight matrix as shown in \Cref{fig:lora}:
\begin{align}
   W' = W + \Delta W = W + AB^\top,
\end{align}
where $A\in \mathbb{R}^{d_\mathrm{in}\times r}$, $B\in \mathbb{R}^{d_\mathrm{out}\times r}$, and the rank $r$. 
With a smaller $r$ compared with the matrix dimensions, it only requires $(d_\mathrm{in}+d_\mathrm{out})r$ parameters for each weight matrix, while full fine-tuning (FT) for dense $\Delta W\in\mathbb{R}^{d_\mathrm{in}\times d_\mathrm{out}}$ results in $d_\mathrm{in}d_\mathrm{out}$ parameters. 
LoRA has been widely used in fine-tuning large models as much less trainable parameters save memory usage at training while retaining performance, making it easily adapted to downstream tasks with limited resources. 

\subsection{SuperLoRA}
\Cref{fig:superlora} shows the overview of SuperLoRA, which is a generalization of LoRA variants to allow high flexibility in the weight update $\Delta W$.
SuperLoRA can be formulated as below:
\begin{align}
\Delta W_{\mathrm{group}_{g}} = \mathcal{F}(\Delta W_{\mathrm{lora}_g}) &=
\mathcal{F} \left( \bigotimes_{k=1}^{K} \Bigg (C_{gk}\prod_{m=1}^{M}{}_{\times m}A_{gkm} \Bigg ) \right),
\label{eq:superlora}
\end{align}
where $\mathcal{F}(\cdot)$ is a simple projection function applied on the results of SuperLoRA modules. 
We denote $\prod_{m=1}^{M}{}_{\times m}$ as tensor products from mode-$1$ to mode-$M$.
Here, $M$ represents the order of the  reshaped $\Delta W_{\mathrm{lora}_g}$ tensor modes, and high-order Tucker decomposition is employed to formulate this high-order tensor. 
SuperLoRA units in \Cref{fig:superlora} are combined with Kronecker product across $K$ splits in a proper shape. 
Depending on reshaping, each split has multiple choices including a combination of dense FT (1D), LoRA (2D), and high-order Tucker decomposition~\cite{tucker1966some}, where $C_{gk}$ (core tensor) and all $A_{gkm}$ (plane factors) are combined with mode-$m$ tensor product. 

For SuperLoRA, we first concatenate all $\Delta W\in \mathbb{R}^{d_i\times d_i}$ across multiple layers to get the total correction of $\Delta W_\mathrm{all}\in \mathbb{R}^{\sum_i d_i^2}$. 
Then, $\Delta W_\mathrm{all}$ is divided into $g$ groups: $\{\Delta W_{\mathrm{group}_g}\}$ for $g\in \{1,2,\ldots, G\}$. 
Each LoRA module will then produce $\Delta W_{\mathrm{group}_g}$. 
Finally, stretch $\Delta W_{\mathrm{group}_g}$ to one dimension, fetch corresponding size of $\Delta W$ from those $\Delta W_{\mathrm{group}_g}$ and add it to candidate weight matrix, \eg, query and value projection weights for attention modules across layers.

\textbf{SuperLoRA and LoTR:} 
While LoRA estimates $\Delta W$ in a weight-wise independent way, SuperLoRA considers the whole weights $\Delta W_\mathrm{all}$ jointly.
It can relax the restriction of the weight shape and geometric meaning of weight axis unlike LoTR. 
Here, the number of groups can be adjusted to balance between parameter amount and fine-tuning performance. 
When the number of groups is the number of weights and the group boundary matches the weight boundary, it corresponds to weight-wise LoRA.
When the number of groups is $G=1$, SuperLoRA corresponds to LoTR~\cite{bershatsky2024lotr}, but with an additional projection mapping $\mathcal{F}$.

\textbf{Reshaping to regular tensor:} 
Grouping multiple layers together by concatenating $\Delta W$ along one axis results in skew $\Delta W_{\mathrm{group}_g}$, limiting the choice of ranks in LoRA modules and leading to worse approximation. 
For example, stacking query and value weight updates as $[\Delta W_\mathrm{q}, \Delta W_\mathrm{v}]$ will be of size $d_\mathrm{in} \times 2 d_\mathrm{out}$, which is less efficient for LoRA as $A$ and $B$ matrices have unbalanced sizes.
To solve this, we propose to reshape $\Delta W_{\mathrm{group}_g}$ to a regular tensor: \ie, square-like 2D matrix, cubic-like 3D tensor, or high-order hyper-cubic tensors having same dimension size across all axes.
This reshaping can reduce the dimension per axis in the order of $\mathcal{O}[N^{1/M}]$ for $N$ being the number of stacking weights, that in return can allow higher rank size per plane factors.
Several examples of grouping and reshaping are discussed in \Cref{appendix:grouping}, and its geometric analysis in \Cref{appendix:geometric}.

\begin{figure}[t]
  \centering
  \begin{subfigure}{0.24\linewidth}
    \includegraphics[width=\linewidth]{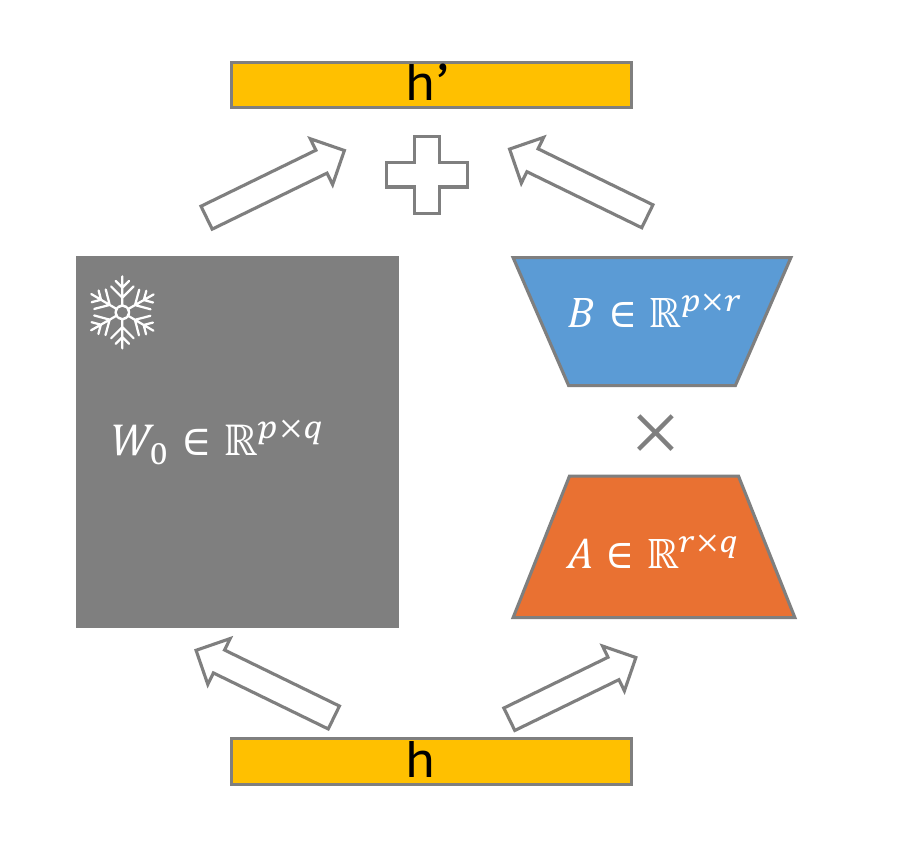}
    \caption{LoRA}
    \label{fig:lora}
  \end{subfigure}
  \begin{subfigure}{0.32\linewidth}
    \includegraphics[width=\linewidth]{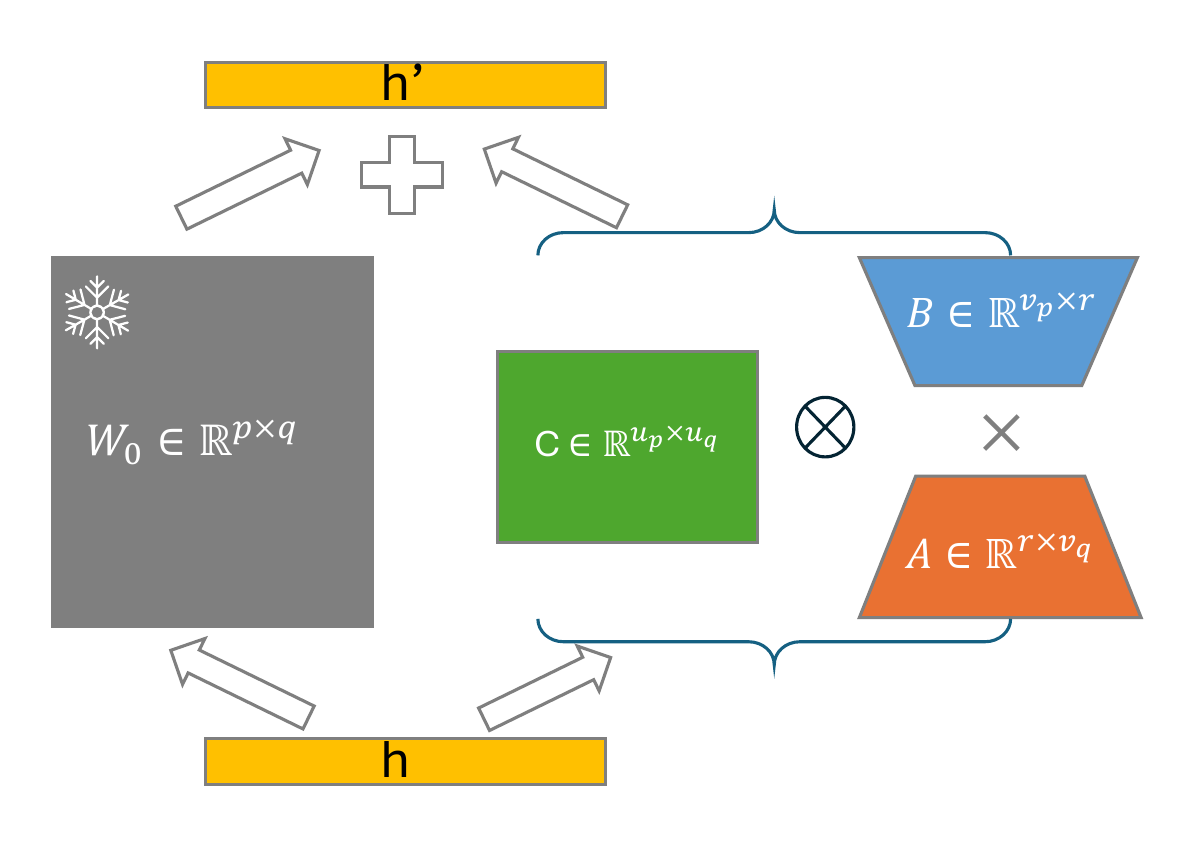}
    \caption{LoKr}
    \label{fig:lokr}
  \end{subfigure}
  \begin{subfigure}{0.4\linewidth}
    \includegraphics[width=\linewidth]{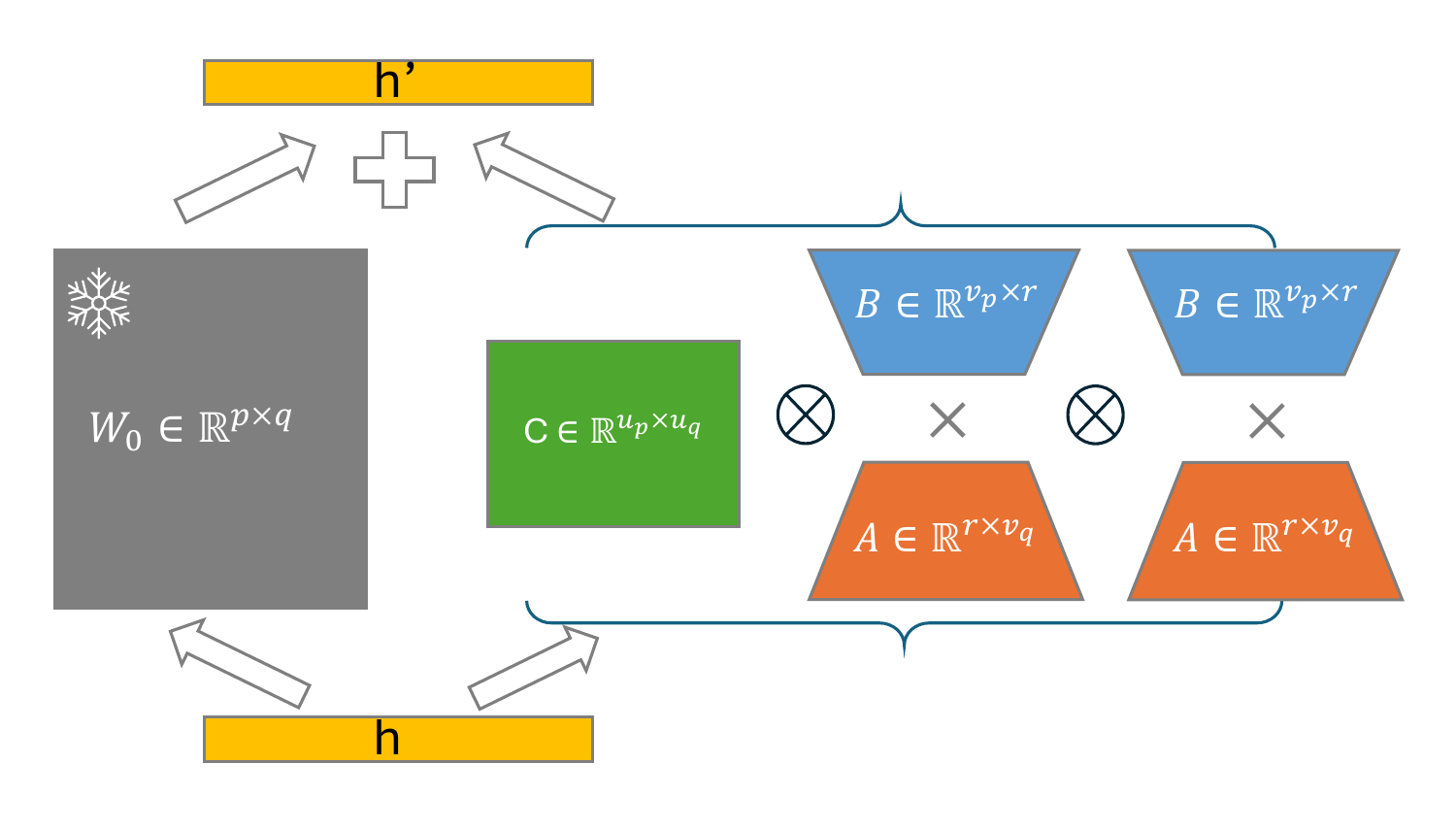}
    \caption{LoNKr (weight-wise version, ours)}
    \label{fig:lonkr}
  \end{subfigure}
  \caption{Overview of (a) LoRA; (b) LoKr; (c) LoNKr (weight-wise version, ours).}
  \label{fig:lodas}
\end{figure}

\textbf{SuperLoRA and LoKr/LoNKr:} 
LoKr is depicted in \Cref{fig:lokr}, which can be extended as shown in \Cref{fig:lonkr}.
We call it LoNKr, which combines $K$ splits composed of sub LoRA units through Kronecker products: \ie, $K>2$ in \Cref{eq:superlora}. 
When $K=2$, it reduces to LoKr but with an additional flexibility.
For example, LoNKr can still adapt multiple attention modules at once with an adjustable group size $G$, unlike weight-wise adaptation of LoKr.

\textbf{LoRTA:} 
Folding a matrix $\Delta W_{\mathrm{group}_g}$ into high-order tensor (\eg, 3D, 4D, 5D) can decrease parameters with tensor rank decomposition, like Tucker decomposition, where $\Delta W_{\mathrm{group}_g}$ is represented by $M$ 2D plane factors and one $M$D core tensor. 
We refer to this variant of SuperLoRA using Tucker decomposition as LoRTA.
For example, when $M = 4$ and $K = 1$, we have 4D tensor rank decomposition as follows:
\begin{align}
    \Delta W_{{\mathrm{group}_g}} = C_{gK} \times_1 A_{gK1}\times_2 A_{gK2}\times_3 A_{gK3}\times_4A_{gK4}\in\mathbb{R}^{d_1\times d_2\times d_3\times d_4},
\end{align}
where $C_{gK} \in \mathbb{R}^{r_1\times r_2\times r_3 \times r_4}$ is a reshaped core tensor, $A_{gKm}\in\mathbb{R}^{d_m\times r}$ is a mode-$m$ 2D plane factor, and $\times_m$ denotes mode-$m$ tensor product.
For simplicity, we set a rank $r = r_m$ for any mode $m\in\{1,2,\ldots, M\}$.

The core tensor may cause the explosion of parameters with larger rank as the number of parameters is exponential as $r^M$.
It may be resolved by restricting the core tensor to be strongly diagonal or identity.
For instance, $M=2$ with identity core tensor $C_{gK}=I$ corresponds to the original LoRA, and $M=r=1$ identity core tensor corresponds to the dense FT.
When using diagonal core tensor, it reduces to canonical polyadic (CP) decomposition.
\Cref{fig:nparams} shows the number of required parameters with CP decomposition. 
One can see that higher-order tensor decomposition can significantly reduce the total number of trainable parameters at a certain rank.
We provide another solution without limiting the core tensor by coupling with the projection layer $\mathcal{F}$ below.

\textbf{Projection:} 
Most LoRA variants assume the resultant $\Delta W_{\mathrm{lora}_g}$ from LoRA modules is the final $\Delta W$ added to $W$ directly. 
However, we can further modify the $\Delta W_{\mathrm{lora}_g}$ through a simple mapping: \eg, we can project much smaller $\Delta W_{\mathrm{lora}_g}$ into larger final $\Delta W_{\mathrm{group}_g}$ to improve the parameter efficiency. 
The compression ratio is $\rho=|\Delta W_{\mathrm{lora}_g}|/|\Delta W_{\mathrm{group}_g}|$, where $|\cdot|$ denotes the total number of elements of the tensor.
We consider a random projection layer based on the fastfood projection~\cite{le2013fastfood} to map $\Delta W_{\mathrm{lora}_g}$ to $\Delta W_{\mathrm{group}_g}$. 

Specifically, the fastfood projection is performed as follows:
\begin{equation}
 \Delta W_{\mathrm{group}_g} = \mathcal{F}(\Delta W_{\mathrm{lora}_g}) = \mathsf{vec}[\Delta W_{\mathrm{lora}_g}] \, \mathcal{H}' \, \mathsf{diag}[\mathcal{G}] \, \varPi \,  \mathcal{H} \, \mathsf{diag}[\mathcal{B}],
\end{equation}
where $\mathsf{vec}[\cdot]$ is a vectorization operator, $\mathsf{diag}[\cdot]$ denotes a diagonalization operator, $\mathcal{H}$ is Walsh--Hadamard matrix, $\mathcal{H}'$ is its truncated version, $\mathcal{G}$ is a random vector drawn from normal distribution, $\varPi$ is a random permutation matrix for shuffling, and $\mathcal{B}$ is a random vector drawn from Rademacher distribution. 
It is a fast Johnson--Lindenstrauss transform with log-linear complexity due to the fast Walsh--Hadamard transform, and no additional parameters are required when the random seed is predetermined.
Further, a nonlinear function such as tanhshrink can be added to make this layer nonlinear. 
To avoid introducing extra parameters for the projection layer, weights of this projection layer is reproduced on the fly with a known random seed and fixed during training and inference.

\textbf{Shuffling:} 
Another simple projection is to use a shuffling function without compression. 
It can be achieved by simplifying the fastfood projection without $\mathcal{H}$, $\mathcal{H}'$, $\mathcal{G}$, and $\mathcal{B}$ but with the random permutation $\varPi$ and projection ratio $\rho=1$.
As SuperLoRA updates all weights at once, we have a flexibility in a way to distribute $\Delta W_{\mathrm{group}_g}$ towards which element of $W$.
To understand how the weight assignment method impacts, we consider a random shuffling case for the projection function $\mathcal{F}$.
Several projection variants including shuffling are discussed in \Cref{appendix:projection}.

\section{Empirical Experiments}
We evaluate SuperLoRA on two different transfer learning tasks: image classification and image generation. 
Below, we describe the experiment settings and results for these tasks as well as visualization of the generated images.

\subsection{Classification transfer task}
\subsubsection{Settings:}
Transfer learning for image classification is conducted between ImageNet21k~\cite{deng2009imagenet} and CIFAR100~\cite{krizhevsky2009learning} based on a ViT-base~\cite{dosovitskiy2020image} model. 
Specifically, a ViT model\footnote{\url{https://huggingface.co/google/vit-base-patch16-224-in21k}} pretrained on ImageNet21k is loaded with a new classifier head\footnote{\url{https://github.com/bwconrad/vit-finetune}} for classifing CIFAR100 dataset. 
More details of the ViT model are described in \Cref{appendix:vit}.
The query and value projection layers in the attention modules are then fine-tuned with SuperLoRA. 
All layers of the pretrained ViT model are frozen except the SuperLoRA parameters and the new classifier head. 
The model is trained for 5,000 steps with the 
stochastic gradient descent (SGD) optimizer, with a batch size of 128 and a learning rate of $0.05$.
The OneCycleLR~\cite{smith2019super} scheduler is used. %

We evaluated SuperLoRA with grouping with/without reshaping to square-like for 2D $\Delta W_{\mathrm{group}_g}$, reshaping version for higher-order $\Delta W_{\mathrm{group}_g}$ including 3D, 4D and 5D. 
The fixed projection layers are inserted to SuperLoRA with reshaping (2D version) and also dense. 
Original weight-wise LoRA is also examined for comparison by setting the number of groups to the number of query and value weights (24 for 12-layer ViT-base) as all projection weights for ViT-base are equal size. 
Each correction weight is of size $768\times 768$ as the projection weight for query/value, resulting in 14M parameters. 
The number of groups are selected from $G \in \{1, 4, 8, 12, 24\}$, every two ranks are evaluated from 1 to 32 for most methods and LoRA $\alpha$ is set to be the same as rank. 
Except for most cases, more ranks are needed to span the parameter axis well, including larger ranks from 34 to 128 and smaller ranks below 8 for LoRTA. 
Projection compression ratio is from $\rho\in \{0.5, 0.25, 0.1, 0.01\}$, and the fixed projection parameters are shared across all groups in our experiments.  

\subsubsection{Results:}

\begin{figure}[t]
    \centering
    \includegraphics[width=\linewidth]{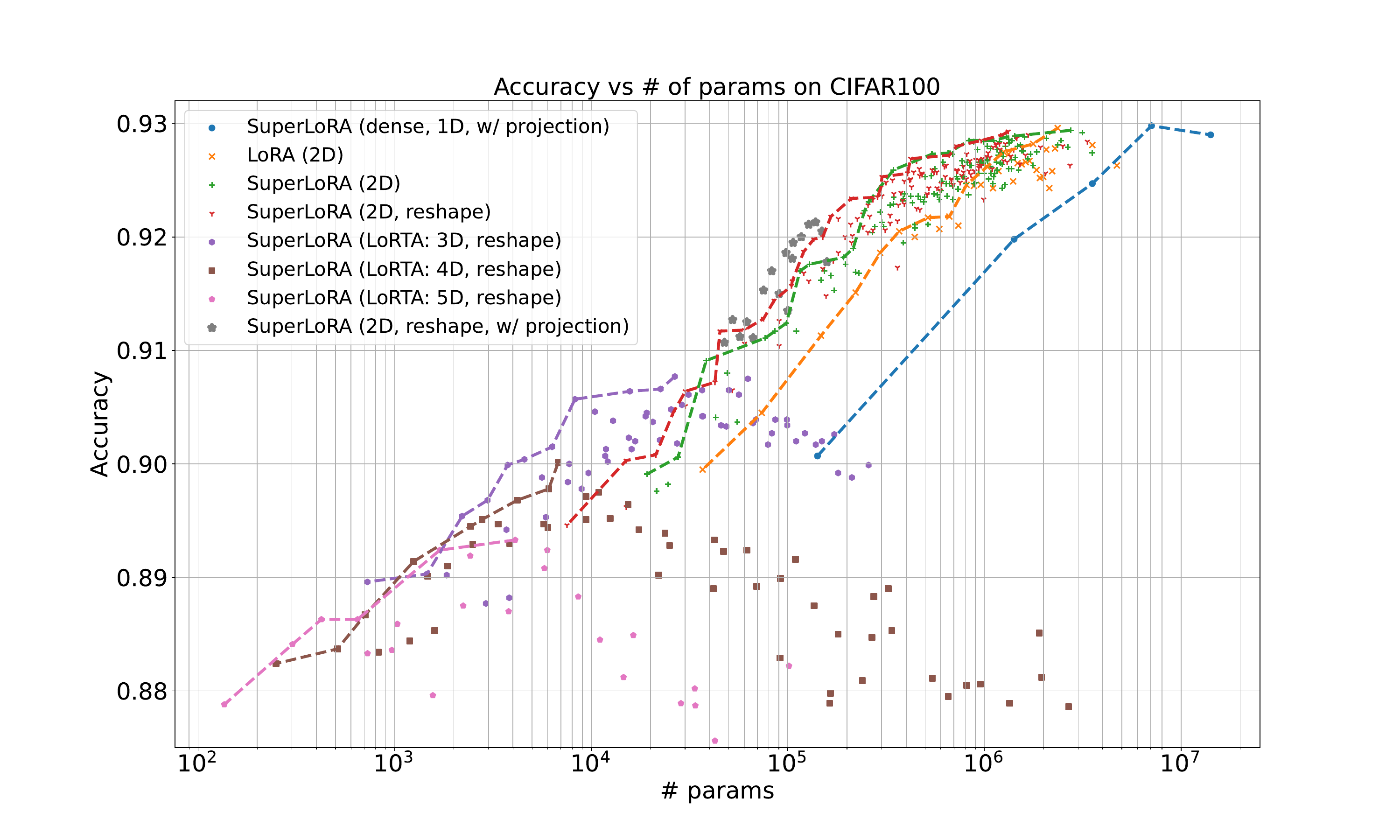}
    \caption{Classification on CIFAR100 dataset with SuperLoRA.}
    \label{fig:classificaiton}       
\end{figure}

Classification results versus the number of parameters are shown in \Cref{fig:classificaiton} with Pareto frontier lines.
Comparing group-wise SuperLoRA (2D with/without reshape) with weight-wise LoRA, we can find that SuperLoRA versions show better performance in terms of the trade-off between classification accuracy and the number of parameters. 
Noticeably, we observe three to four times advantage in terms of parameter efficiency for the same accuracy.
As the largest number of groups is set to 24 (\ie LoRA), it indicates smaller number of groups are superior. 
This may be because ViT model is excessively large for the CIFAR100 dataset, with much more redundant weights. 
Grouping weights and layers together can reduce noise brought by the redundancy. 
With reshaping $\Delta W_{\mathrm{group}_g}$ to a square matrix, classification accuracy further increases in the lower parameter regime and the range of parameters the model can cover becomes wider as higher rank can be used while maintaining a smaller number of parameters.

To discuss the effect of higher-order tensor folding, the order $M$ is set to be 3, 4 and 5 for SuperLoRA (\ie LoRTA) as well as 2. 
For $M=2$ cases with 2D tensor, we use identity core tensor like typical LoRA. 
With the increase of order from 2 to 5, higher order takes place lower-order at fewer-parameter regimes. 
Moreover, data points for high-order LoRTA show a hill-like trend with the increase of parameters. 
This may be caused by the inefficient core tensor, which increases parameters rapidly without benefiting the accuracy. 
When comparing the lowest rank LoRA (which achieves around $0.9$ accuracy with about $4\times 10^4$ parameters), our LoRTA (3D) significantly improves the accuracy by about $1\%$ at the comparable number of parameters, and more significantly reduces the number of parameters by 10 folds to keep the comparable accuracy of $0.9$.

Finally, we address the impact of the random but fixed projection layer $\mathcal{F}$. 
Fixed fastfoood projection is applied on SuperLoRA (1D, dense) and SuperLoRA (2D, reshape). 
For 1D dense with projection, the plot for a projection ratio of $\{1, 0.5, 0.25, 0.1, 0.01\}$ is placed from right to left in \Cref{fig:classificaiton}. 
The classification accuracy dropped less than $1\%$ from projection ratio 1 to $0.1$ (\ie $90\%$ less parameters), but it is worse than LoRA. 
For SuperLoRA (2D, reshape, w/projection), we evaluated a projection ratio of $\{0.5, 0.25, 0.1\}$. 
To get some results of projection for the number of parameters around $10^4$ and $10^5$, we select a few settings for SuperLoRA (2D, reshape) with $G=1$ as shown in the figure with a marker of dark stars. 
Most projection results demonstrate better accuracy compared with other SuperLoRA settings without projection in the same number of parameters level. 
This result shows a smaller adapter with fixed projection layer is a strong functionality to improve the parameter efficiency of SuperLoRA. 

Another experiment of transfer learning from ImageNet1k to CIFAR10 classification shows similar results, achieving $3$ to $10$-fold improved parameter efficiency, as discussed in \Cref{appendix:cifar10}.

\subsection{Image generation transfer task}
\subsubsection{Settings:}
For the image generation task, SuperLoRA is evaluated by transfer learning between SVHN~\cite{netzer2011reading} and MNIST datasets~\cite{lecun1998gradient}. 
Both datasets have 10 classes corresponding to images of the digits 0 to 9, where the SVHN images have a more complicated color background, while the MNIST images are nearly black-and-white with a plane black background.
We mainly work on the transfer learning from SVHN to MNIST.
The reverse transfer learning from MNIST to SVHN is discussed in \Cref{appendix:2svhn}.

The model we worked on is a classifier-free diffusion model~\cite{ho2021classifier} and the correction weights from LoRA variants are added to query and value projection matrices in the attention modules of U-Net backbone~\cite{ronneberger2015u}. 
Note that the size of projection weights differs across layers for this U-Net structure, which allows us to examine the performance of SuperLoRA after breaking the boundaries of different weight matrices. 
More details of the diffusion model are described in \Cref{appendix:unet}.
For comparison, the original weight-wise LoRA and dense FT are also evaluated. 
For SuperLoRA variant, LoRA, LoNKr and LoRTA consider three versions: weight-wise, group-wise and group-reshaped. 
The scaling factor $\alpha$ of LoRA is fixed to $2.0$ for all variants unless specified. 
40 epochs with a batch size of $32$ are carried out and results plotted are mainly from epoch 20 noticing convergence becomes stable around epoch 20. 
The maximum rank is set to $32$ by default and a constraint $r<\min(d_\mathrm{in}, d_\mathrm{out})$ is imposed. 
To evaluate the quality of images generated by the fine-tuned diffusion model, we consider several metrics including Inception Score (IS)~\cite{salimans2016improved}, Fr\'{e}chet Inception Distance (FID)~\cite{heusel2017gans}, Multi-Scale Intrinsic Distance (MSID)~\cite{tsitsulin2019shape}, Kernel Inception Distance (KID)~\cite{binkowski2018demystifying}, Recall and Precision~\cite{kynkaanniemi2019improved}.
Except for the recall and precision metrics, all metrics should be lower for higher-quality image generations.
As we found $\ell_1$-distance based IS is more consistent to the perceptual visual quality, we mainly focus on IS metric results in the main content, while the  results for other metrics can be found in \Cref{appendix:2mnist}. 
For following figures, Pareto frontier lines/dots are mainly shown to provide the limit of each method, while Appendix provides more complete figures with all data points.

\subsubsection{Grouping effect:}
\begin{figure}[t]
    \centering
    \begin{subfigure}{0.49\linewidth}
        \includegraphics[width=\linewidth]{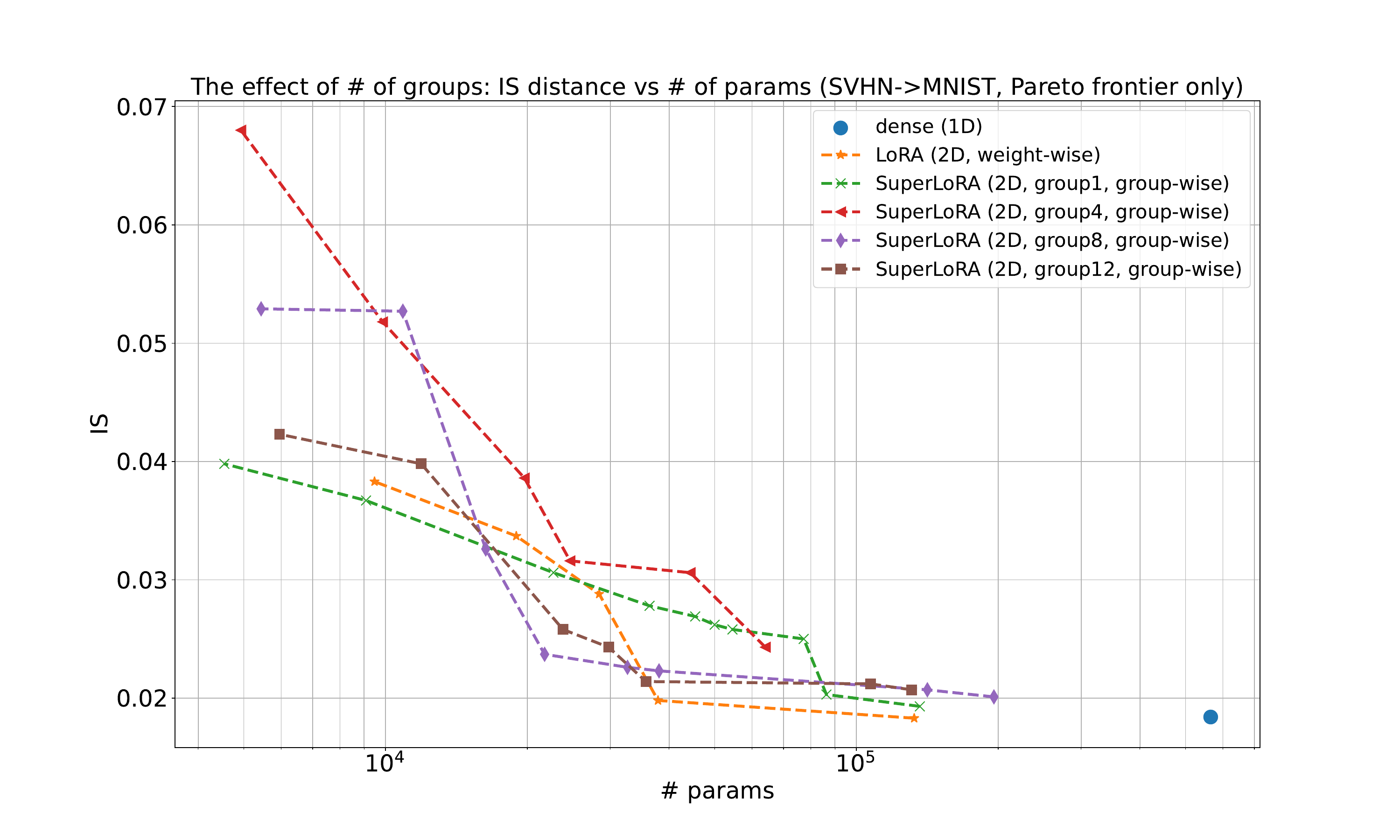}
        \caption{weight-wise \vs group-wise}
        \label{fig:fig_group_is}       
    \end{subfigure}
    \begin{subfigure}{0.49\linewidth}
        \includegraphics[width=\linewidth]{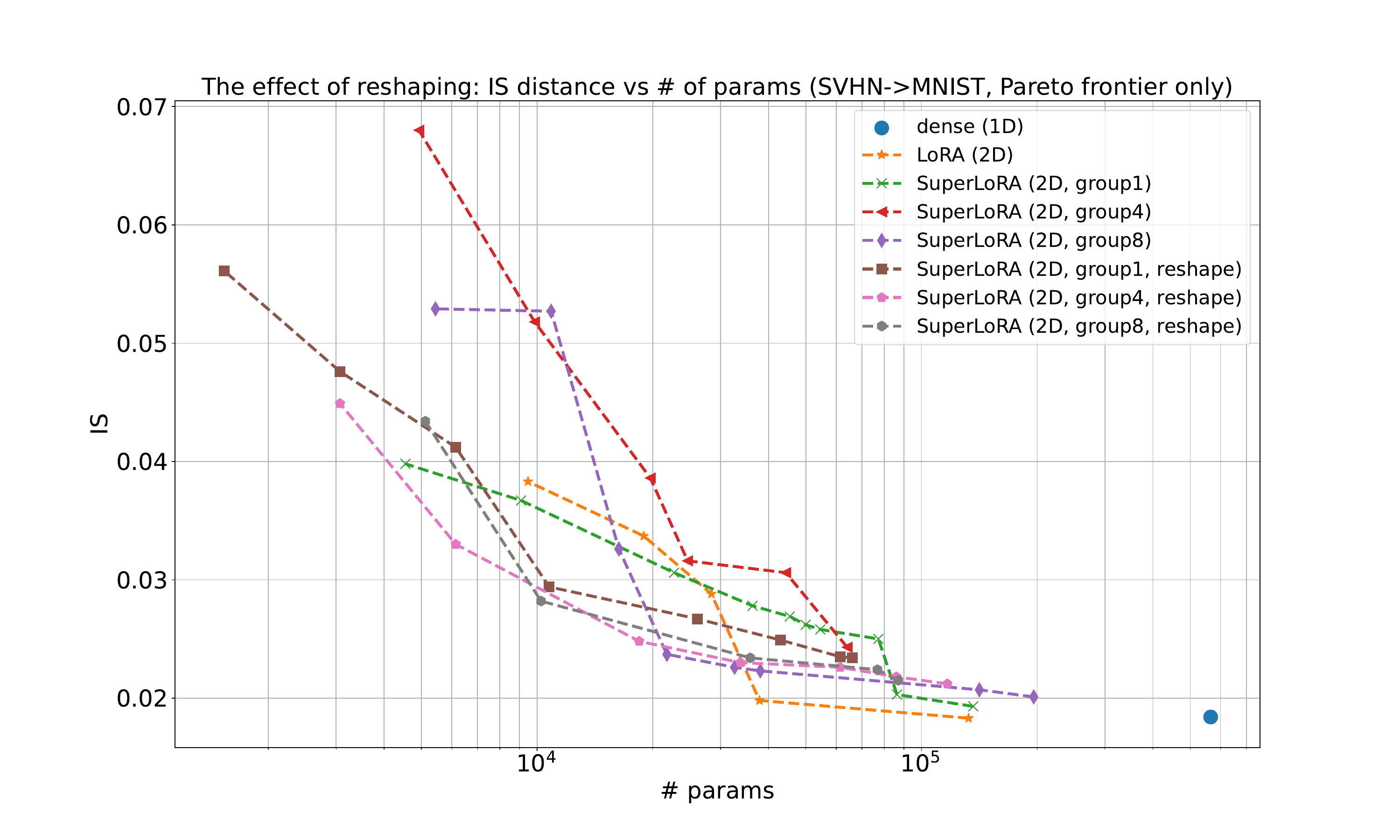}
        \caption{reshaping \vs non-reshaping}
        \label{fig:fig_reshape_is}       
    \end{subfigure}
    \caption{Comparison between weight-wise LoRA and group-wise SuperLoRA: (a) sweeping rank and group; (b) with/without reshaping to square-like.}
\end{figure}

First, we evaluated how splitting all $\Delta W_\mathrm{all}$ into multiple groups affects the performance. 
\Cref{fig:fig_group_is} shows the results of dense, original weight-wise LoRA and group-wise SuperLoRA with different number of groups. 
Sweeping the rank and the number of groups, we plot the image quality metrics in y-axis and the required number of trainable parameters in x-axis. 
Pareto frontier lines/data points are also shown in the figure. 

\Cref{fig:fig_group_is} shows that the dense FT for $\Delta W$ presents the best IS, while requiring most parameters. 
Original weight-wise LoRA is closest to dense, in terms of both IS and parameter amount. 
However, in low-parameter regimes, SuperLoRA (2D, group1) shows the best results compared with other grouping. 
While in the middle of parameter amount axis, other splittings including groups $G=8$ and $12$ show slightly better IS compared with LoRA. 
Besides, splitting $\Delta W_\mathrm{all}$ shows much more data points compared with both LoRA and dense, providing us higher flexibility to adjust the trade-off between quality and parameter efficiency especially when the memory resource is limited. 

\subsubsection{Reshaping effect:}
To evaluate the importance of reshaping, we compare group-wise SuperLoRA with and without reshaping in \Cref{fig:fig_reshape_is}. 
For weight-wise LoRA, most weight matrices corrected are square already. 
For all splitting with groups $G=1$, $4$ and $8$, we confirmed that reshaping shows smaller number of parameters and better IS compared with their corresponding non-reshaping counterparts. 
This indicates that reshaping $\Delta W$ to regular tensor array (square, cube, and hyper-cube) is vital for SuperLoRA fine-tuning to prevent unbalanced skew tensors when adapting multiple weights at once.

\subsubsection{LoKr vs. LoNKr:}
In 2D $\Delta W$, we also compared LoKr with our proposed extension LoNKr, a variant of SuperLoRA. 
We evaluated LoNKr when the number of splits is $K\in \{2,3,4\}$, where $K=2$ corresponds to the original LoKr. 
For the dense factor on the left in LoNKr/LoKr as shown in \Cref{fig:lonkr}, dimension is fixed to $6$, $8$ or $10$. 
\Cref{fig:fig_lonkr_is} shows that more splits provide us more choices in low-parameter regimes, especially for group-wise LoNKr. 
LoNKr shows much more data points and better IS when the number of parameters is less than $5{,}000$. 
And the least parameter for LoKr and LoNKr dropped greatly from 500 to 150. 

\subsubsection{LoRTA:}
\begin{figure}[t]
    \centering
    \begin{subfigure}{0.49\linewidth}
        \includegraphics[width=\linewidth]{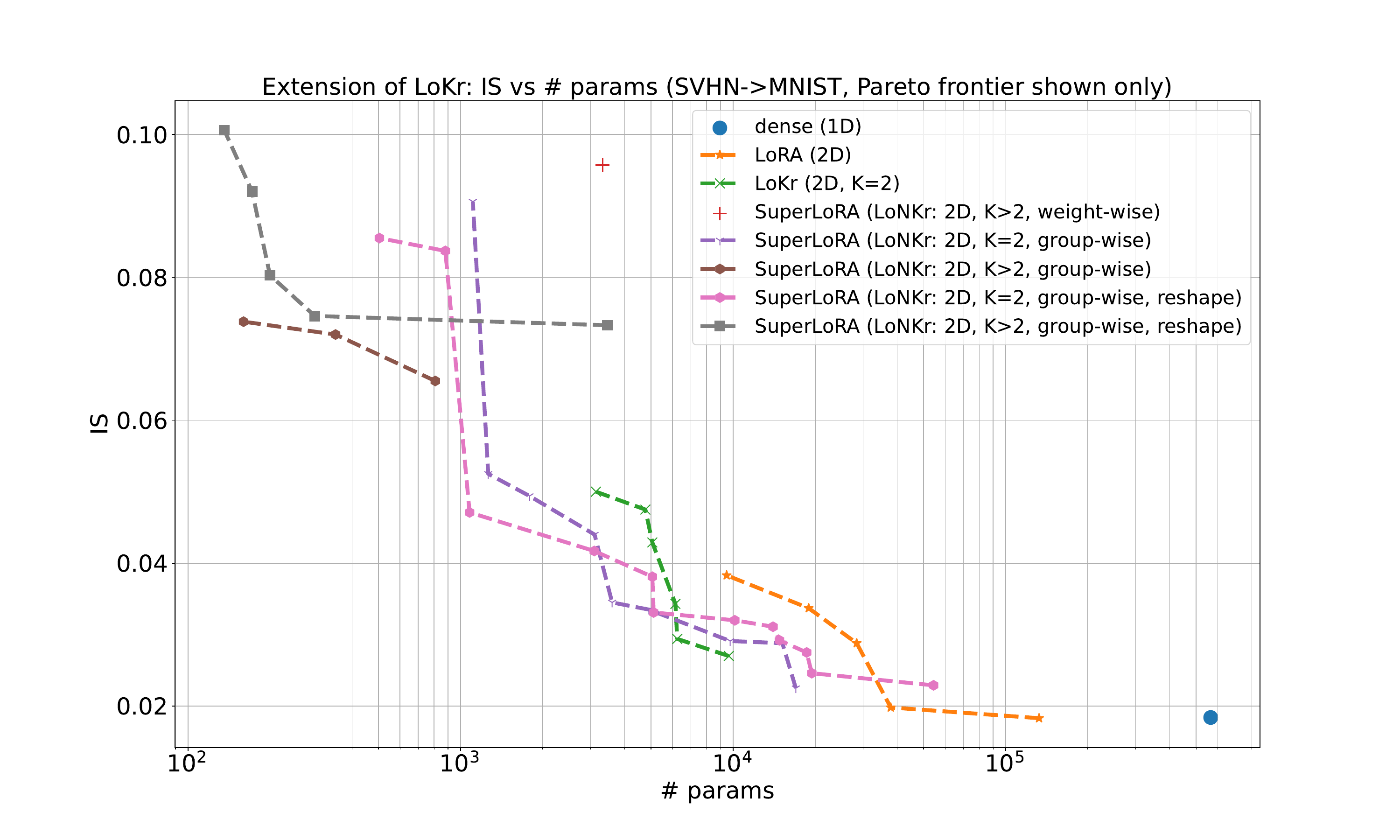}
        \caption{SuperLoRA (LoNKr)}
        \label{fig:fig_lonkr_is}       
    \end{subfigure}
    \begin{subfigure}{0.49\linewidth}
        \includegraphics[width=\linewidth]{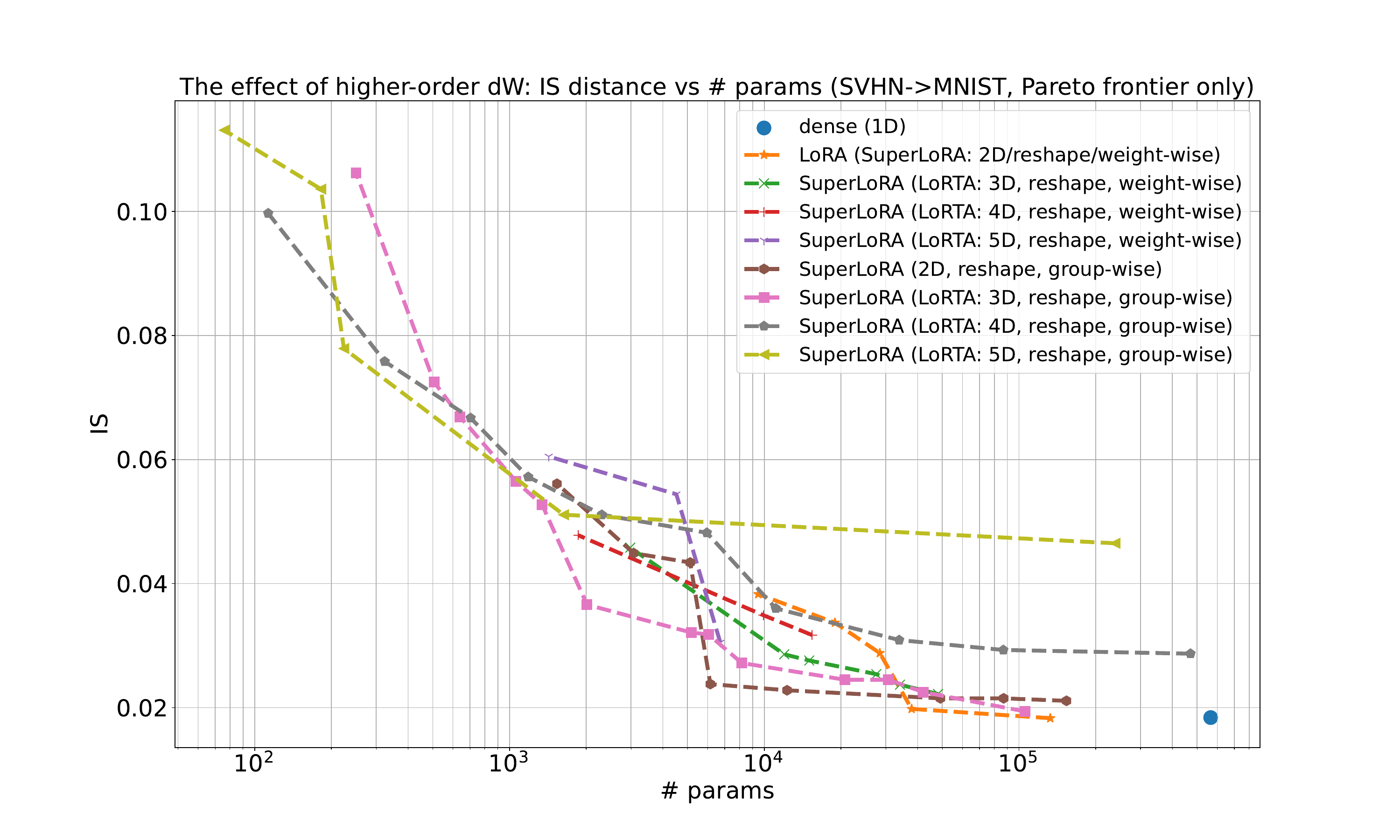}
        \caption{SuperLoRA (LoRTA)}
        \label{fig:fig_lorta_is}       
    \end{subfigure}
    \caption{Performance of SuperLoRA variants: (a) LoNKr as shown in \Cref{fig:lonkr}; (b) LoRTA when folding to high-order tensor.}
\end{figure}

LoRTA reshapes $\Delta W_\mathrm{all}$ to high-order tensor. 
We evaluated 3D, 4D and 5D, as data points become much less when the dimension is too small for all planes when order is larger than 5D. 
From \Cref{fig:fig_lorta_is}, the higher the order of tensor folding, the less data points we have. 
In both weight-wise and group-wise version, 5D LoRTA reduces the least parameter it requires. 
Especially for group-wise LoRTA, 5D LoRTA requires less than $80$ parameters to produce a result compared with beyond $1000$ for 2D LoRTA and beyond 200 for 3D LoRTA, while original LoRA needs about $10^4$ parameters, about 120-fold more parameters. 
To achieve a comparable IS of LoRA having $10^4$ parameters, LoRTA (3D) just needs $2\times 10^3$ parameters, \ie $5$-fold reduction.

\subsubsection{Projection effect:}
SuperLoRA can use a projection layer $\mathcal{F}$ which is randomly initialized but fixed at both finetuning and inference.
Linear fastfood projection and nonlinear projection with tanshrink applied after the linear projection matrix are evaluated. 
Besides, a modified version of fastfood projection with random Gaussian instead of random binary $\mathcal{B}$ is also tested for both linear and nonlinear versions, denoted as linear\textsubscript{v2} and nonlinear\textsubscript{v2} respectively. 
The projection matrix is shared across all groups. 
We evaluated number of groups $G\in\{1,4\}$, rank $r\in\{1,4,8\}$ and projection ratio $\rho\in \{0.01, 0.1, 0.5\}$ on SuperLoRA (2D, reshape) and SuperLoRA (LoRTA, reshape) for 3D, 4D and 5D tensor. 

\Cref{fig:fig_linear_is} demonstrates with smaller projection ratio, required parameters for both SuperLoRA (2D, reshape) and SuperLoRA (LoRTA, group-wise) are pushed to extremely low-parameter regimes. 
The least parameter required becomes only about 30, compared with 10,000 for original LoRA. 
Surprisingly, linear version for both methods shows better performance than nonlinear version which are attached in \Cref{appendix:projection}. 
Besides, in extremely low-parameter regimes, higher rank with projection layer for SuperLoRA (LoRTA, group-wise) works better than small ranks itself, showing promising direction to explore projection layer in extremely low-parameter regime. 
In terms of linear \vs linear\textsubscript{v2},  linear\textsubscript{v2} shows better performance in higher-parameter area while linear works better in lower-parameter area, even better than SuperLoRA (LoRTA) without projection.

\subsubsection{Shuffling effect:}
\begin{figure}[t]
    \centering
    \begin{subfigure}{0.49\linewidth}
        \includegraphics[width=\linewidth]{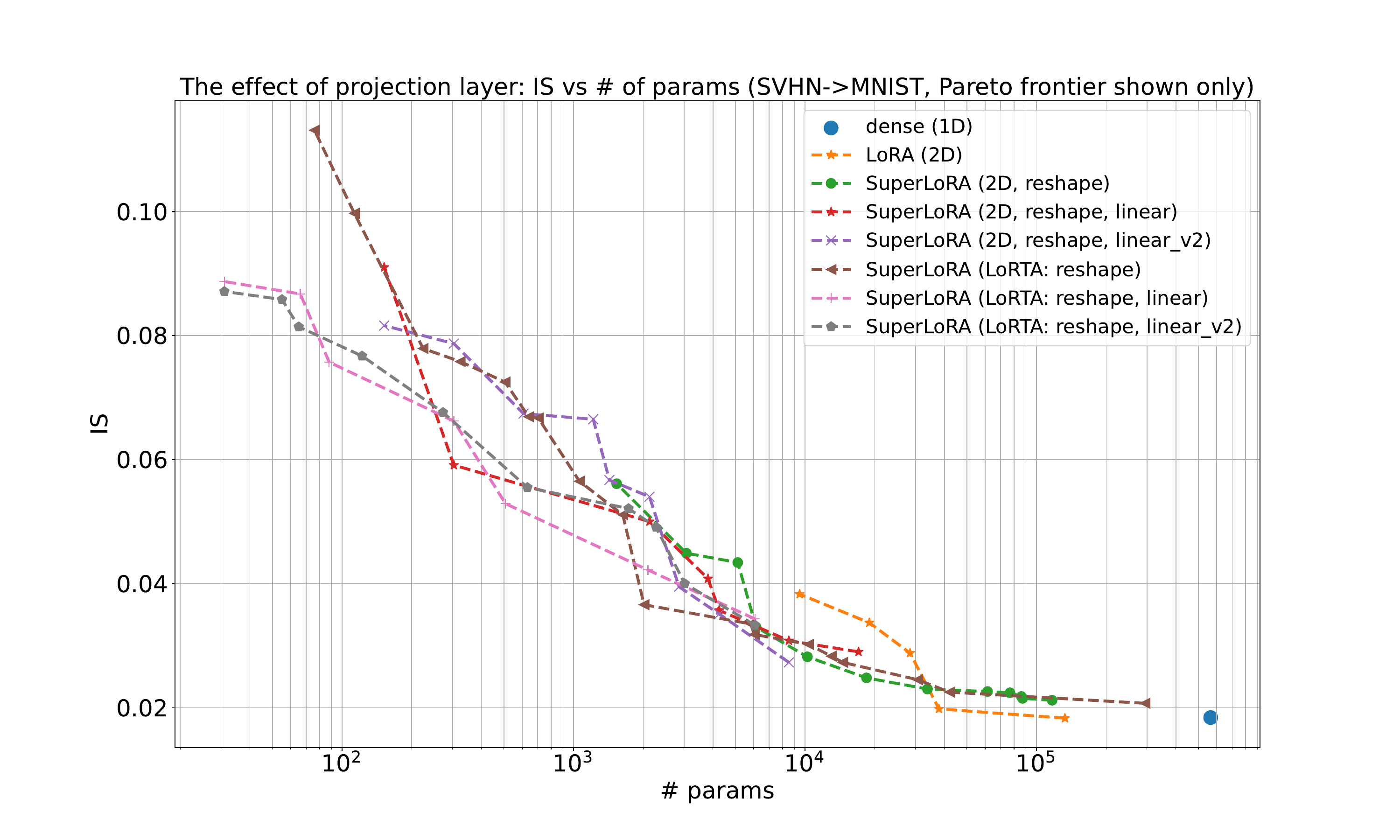}
        \caption{fixed random projection within group}
        \label{fig:fig_linear_is}       
    \end{subfigure}
    \begin{subfigure}{0.49\linewidth}
        \includegraphics[width=\linewidth]{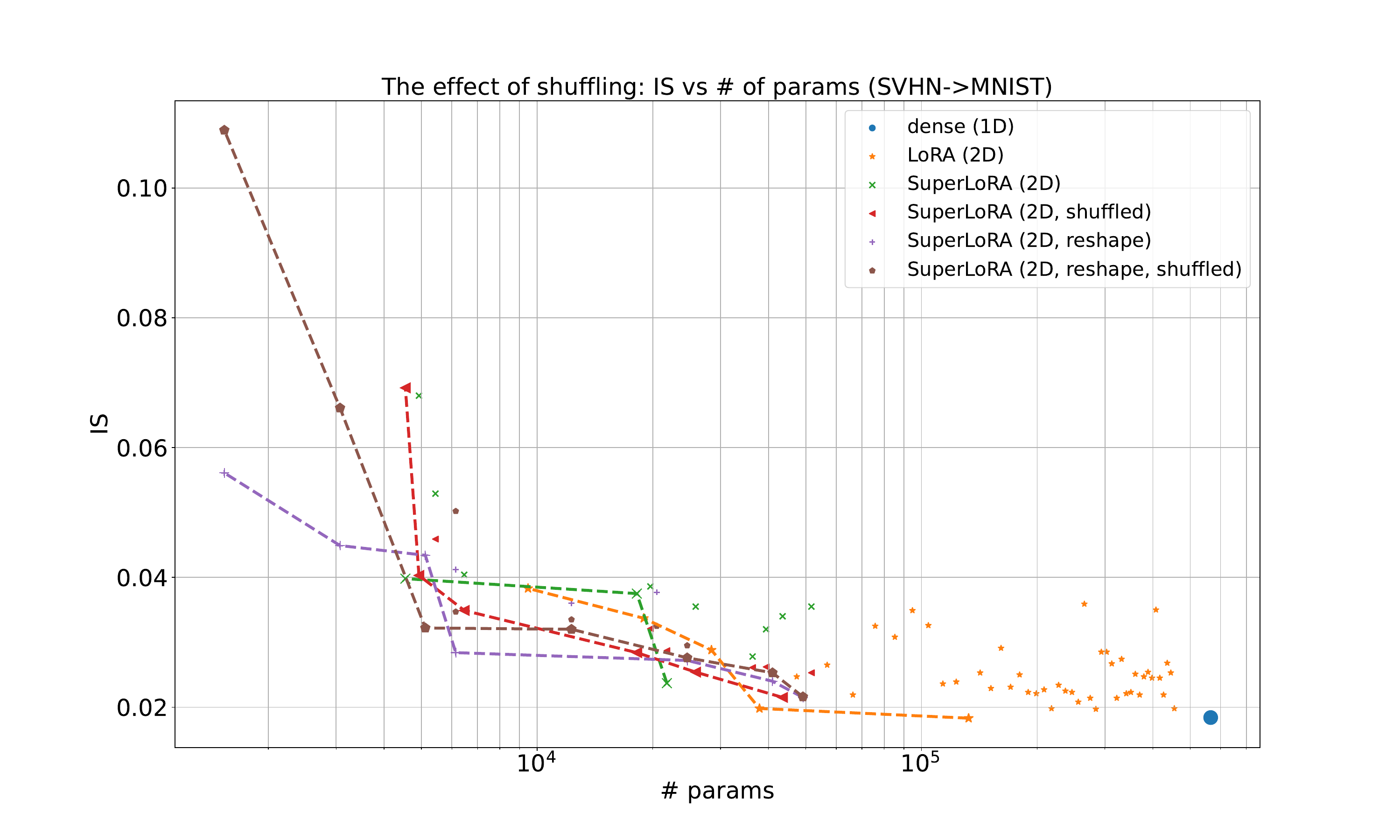}
        \caption{fixed random shuffling within group}
        \label{fig:fig_shuffle_is}       
    \end{subfigure}
    \caption{SuperLoRA under different projection modes: linear, linear\textsubscript{v2} and shuffling.}
\end{figure}

As another simple projection, we studied a random shuffling to distribute $\Delta W_\mathrm{group}$ before adding it to corresponding $W$. 
We evaluated SuperLoRA (2D) and SuperLoRA (2D, reshape) with/without shuffling for groups $G\in \{1, 4, 8, 16]\}$ and ranks $r\in \{1, 4, 8\}$, where the shuffled indexes are shared across all groups. 
The shuffling corresponds to one of fastfood projection modes by setting projection ratio to $\rho=1$ with only permutation matrix $\varPi$. 
As shown in \Cref{fig:fig_shuffle_is}, shuffling inside groups had no harm on IS. 
It even improved IS for SuperLoRA (2D) in most cases. 

\subsection{Visualization}
\begin{figure}[t]
    \centering
    \begin{subfigure}{0.22\linewidth}
        \includegraphics[width=\linewidth]{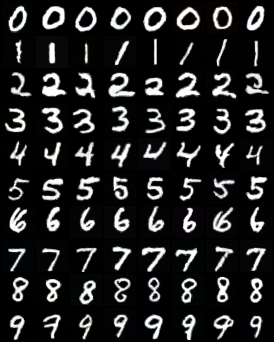}
        \captionsetup{font=tiny,justification=centering}
        \caption{dense \\$\#$param 565,248\\IS 0.0184}
    \end{subfigure}
    \begin{subfigure}{0.22\linewidth}
        \includegraphics[width=\linewidth]{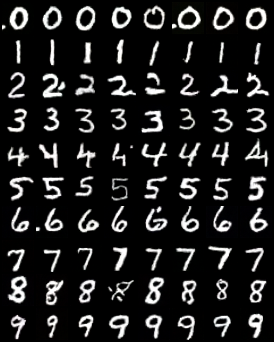}
        \captionsetup{font=tiny,justification=centering}
        \caption{LoRA $r=8$\\$\#$param 75,776\\IS 0.03025}
    \end{subfigure}
        \begin{subfigure}{0.22\linewidth}
        \includegraphics[width=\linewidth]{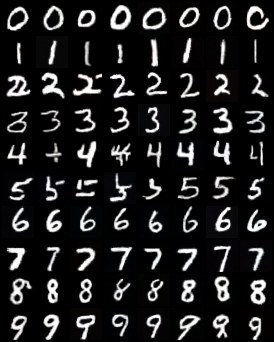}
        \captionsetup{font=tiny,justification=centering}
        \caption{SuperLoRA \\$\#$param 70,720\\IS 0.0305}
        \label{fig:loragroup_high}
    \end{subfigure}
        \begin{subfigure}{0.22\linewidth}
        \includegraphics[width=\linewidth]{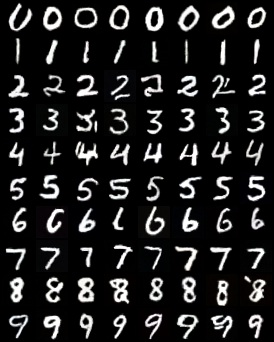}
        \captionsetup{font=tiny,justification=centering}
        \caption{SuperLoRA\\$\#$param 73,728\\IS 0.0263}
    \end{subfigure}
    \caption{Visualization of generated images under high-parameter level ($>70{,}000$).}
    \label{fig:visualization_high}
\end{figure}

To better understand the superiority of SuperLoRA, especially in low-parameter regimes, we visualize a set of generated images from SuperLoRA, as well as dense FT and LoRA, from a range of parameter setting: high-parameter ($>70{,}000$), middle-parameter (from 5,000 to 10,000), low-parameter (around 1,000) and extremely-low parameter ($<100$) regimes. 
We selected one image with the best IS for each hyper-parameter setting we have tested under same level of parameter amount. 
\Cref{fig:visualization_high} shows that all generated images by the transfer learning model from SVHN to MNIST are close to images from MNIST dataset itself with black-white background, removing most domain information of color SVHN. 
SuperLoRA (2D, group8, rank13) in \Cref{fig:loragroup_high} shows competitive results with LoRA (rank8) using $5{,}000$ less parameters. 

\begin{figure}[t]
    \centering
    \begin{subfigure}{0.18\linewidth}
        \includegraphics[width=\linewidth]{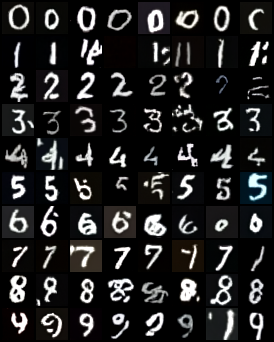}
        \captionsetup{font=tiny,justification=centering}
        \caption{LoNKr\\$\#$param 5,112\\IS 0.036}
    \end{subfigure}
        \begin{subfigure}{0.18\linewidth}
        \includegraphics[width=\linewidth]{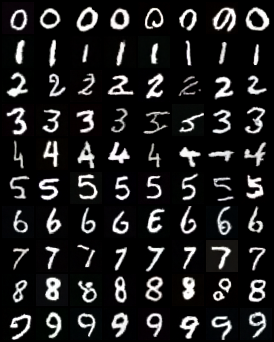}
        \captionsetup{font=tiny,justification=centering}
        \caption{SuperLoRA\\$\#$param 10,752\\IS 0.0294}
    \end{subfigure}
        \begin{subfigure}{0.18\linewidth}
        \includegraphics[width=\linewidth]{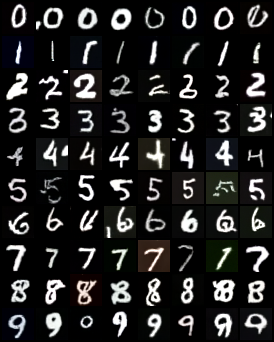}
        \captionsetup{font=tiny,justification=centering}
        \caption{LoRTA\\$\#$param 8,160\\IS 0.0272}
    \end{subfigure}
            \begin{subfigure}{0.18\linewidth}
        \includegraphics[width=\linewidth]{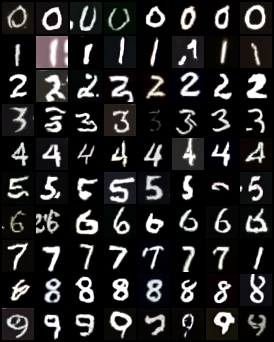}
        \captionsetup{font=tiny,justification=centering}
        \caption{LoRTA\\$\#$param 11,100\\IS 0.036}
    \end{subfigure}
    \begin{subfigure}{0.18\linewidth}
        \includegraphics[width=\linewidth]{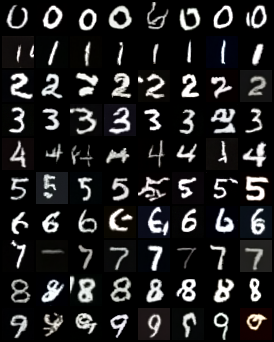}
        \captionsetup{font=tiny,justification=centering}
        \caption{SuperLoRA w/ p\\$\#$param 8,512\\IS 0.0273}
    \end{subfigure}
    \caption{Visualization of generated images under middle-parameter level ($[5{,}000, 20{,}000]$).}
    \label{fig:visualization_mid}
\end{figure}

For the middle-parameter regimes, \Cref{fig:visualization_mid} shows visualization of LoNKr, SuperLoRA (2D, reshape), LoRTA (3D, reshape), LoRTA (4D, reshape) and SuperLoRA (2D, reshape, projection). 
More domain information with colorful digits and background occur occasionally. 
There are also some missing digits presented in middle-parameter area.

\begin{figure}[t]
    \centering
    \begin{subfigure}{0.18\linewidth}
        \includegraphics[width=\linewidth]{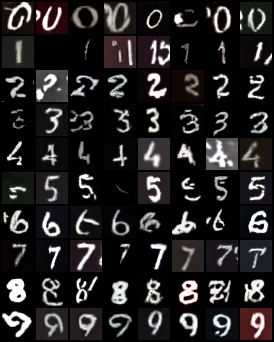}
        \captionsetup{font=tiny,justification=centering}
        \caption{LoNKr\\$\#$param 1,080\\IS 0.0471}
        \label{fig:visualization_low1}
    \end{subfigure}
    \begin{subfigure}{0.18\linewidth}
        \includegraphics[width=\linewidth]{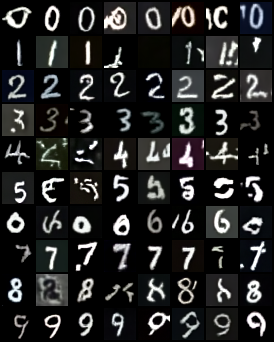}
        \captionsetup{font=tiny,justification=centering}
        \caption{LoRTA\\$\#$param 1,060\\IS 0.0565}
        \label{fig:visualization_low2}
        \end{subfigure}
    \begin{subfigure}{0.18\linewidth}
        \includegraphics[width=\linewidth]{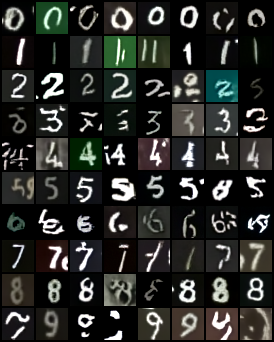}
        \captionsetup{font=tiny,justification=centering}
        \caption{SuperLoRA w/ p\\$\#$param 832\\IS 0.0607}
        \label{fig:visualization_low3}
    \end{subfigure}
        \begin{subfigure}{0.18\linewidth}
        \includegraphics[width=\linewidth]{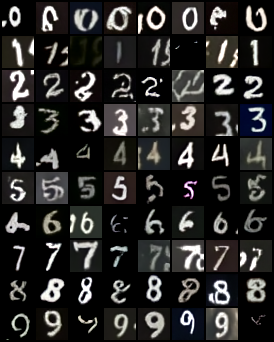}
        \captionsetup{font=tiny,justification=centering}
        \caption{LoRTA\\$\#$param 76\\IS 0.1131}
    \end{subfigure}
            \begin{subfigure}{0.18\linewidth}
        \includegraphics[width=\linewidth]{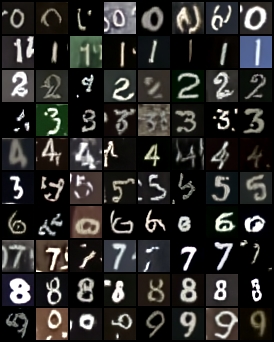}
        \captionsetup{font=tiny,justification=centering}
        \caption{LoRTA w/ p\\$\#$param 31\\IS 0.0871}
    \end{subfigure}
    \caption{Visualization of generated images under low-parameter level ($1{,}000$) and extremely-low level ($<100$).}
    \label{fig:visualization_low}
\end{figure}

When the number of parameters is as low as $1{,}000$, even though only few choices left like LoNKr and LoRTA, one can always stretch hyper-parameter settings from middle-parameter level coupled with fixed linear projection layer to compress the tensor size. 
In this way, the strength of middle-parameter level gets extended to low-parameter area. 
As shown in \Cref{fig:visualization_low}, compared with the visualization from middle-parameter results, more missing digits and more colorful backgrounds are presented. 

Finally, we also visualized a few images from extremely-low parameter level less than 100 in \Cref{fig:visualization_low}. 
Surprisingly, domain transfer in those images is somewhat realized from SVHN to MNIST even with such an extremely few parameter case such as $31$, which is more than four orders of magnitude smaller than dense FT.


\section{Conclusion}
We proposed a new unified framework called SuperLoRA, which generalizes and extends LoRA variants including LoKr and LoTR.
SuperLoRA provides some extended variants, which we refer to as LoNKr and LoRTA. 
It offers a rich and flexible set of hyper-parameters, including the rank of factorization, the choice of projection function, projection ratio, the number of groups, the order of tensor, and the number of Kronecker splits.
Through two types of transfer learning experiments, we demonstrated that SuperLoRA achieves promising results in parameter efficiency for fine-tuning at low-parameter and extremely-low-parameter regimes.
We could reduce the required number of parameters by 3 to 10 folds compared to LoRA. 
Future work includes to study the projection functions to further improve the efficiency in extremely-low-parameter regimes, and applications to various transfer learning tasks along with different large models such as LLMs.

\bibliographystyle{splncs04}
\bibliography{reference}
\newpage

\input{appendix}

\end{document}

%% file: appendix.tex
\appendix
\setcounter{figure}{10}

\section{Appendix}

\subsection{Illustration of ViT model in detail}
\label{appendix:vit}

\begin{figure}[b]
    \centering
    \includegraphics[width=0.65\linewidth]{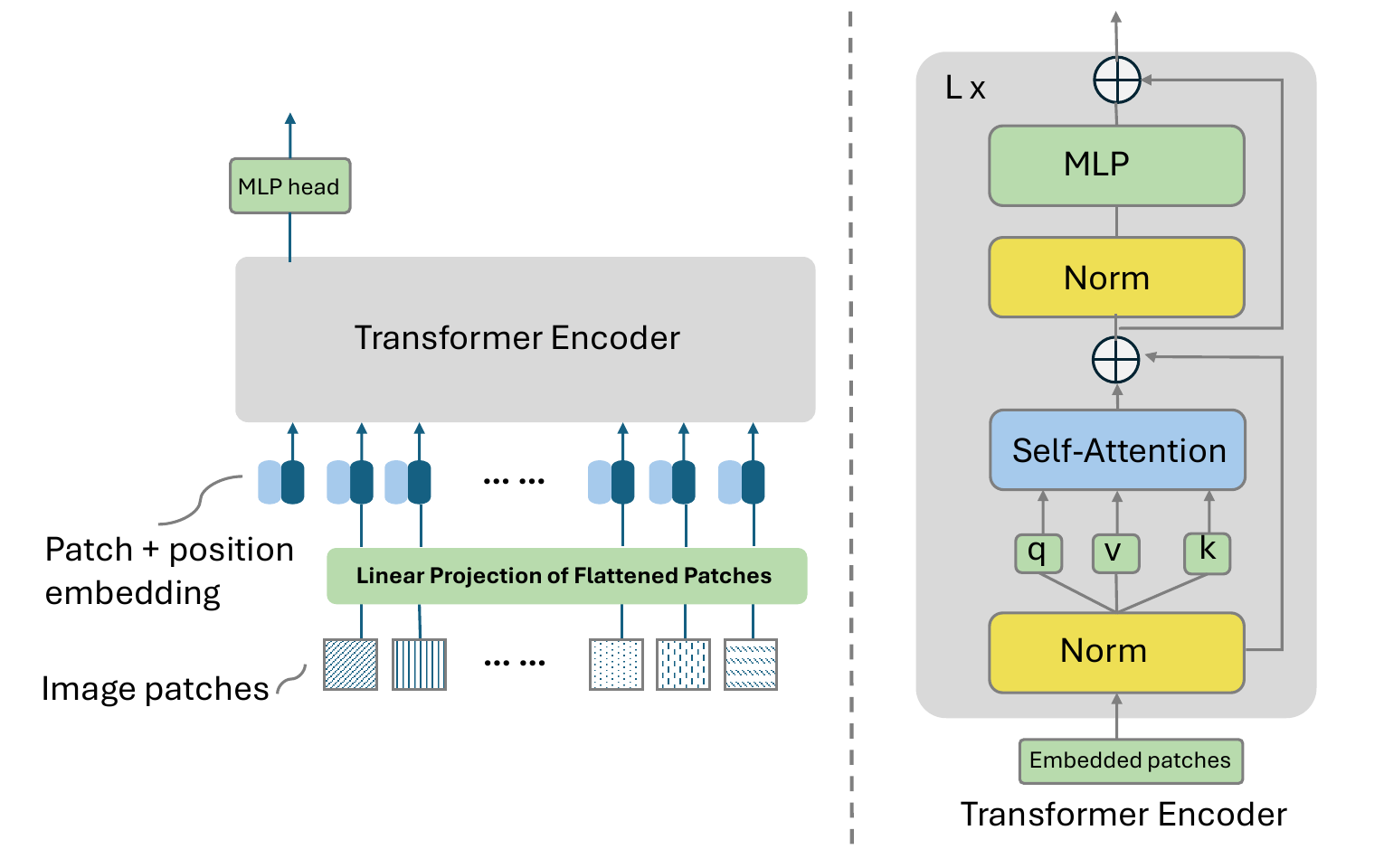}
    \caption{ViT model structure.}
    \label{fig:vit}       
\end{figure}

The ViT model that we used for the classification task is adapted from a public codebase\footnote{\url{https://github.com/bwconrad/vit-finetune}}.
The detailed structure of the ViT is depicted in \Cref{fig:vit}, where we only fine-tune the projection layers for query and value in the Self-Attention modules. %
In ViT-base, depth ($L$ in \Cref{fig:vit}) is set to be 12 and dimension is 768.
The total number of parameters of the ViT base model is $86.6$M.

\subsection{Illustration of diffusion model in detail}
\label{appendix:unet}

\begin{figure}[t]
    \centering
    \includegraphics[width=\linewidth]{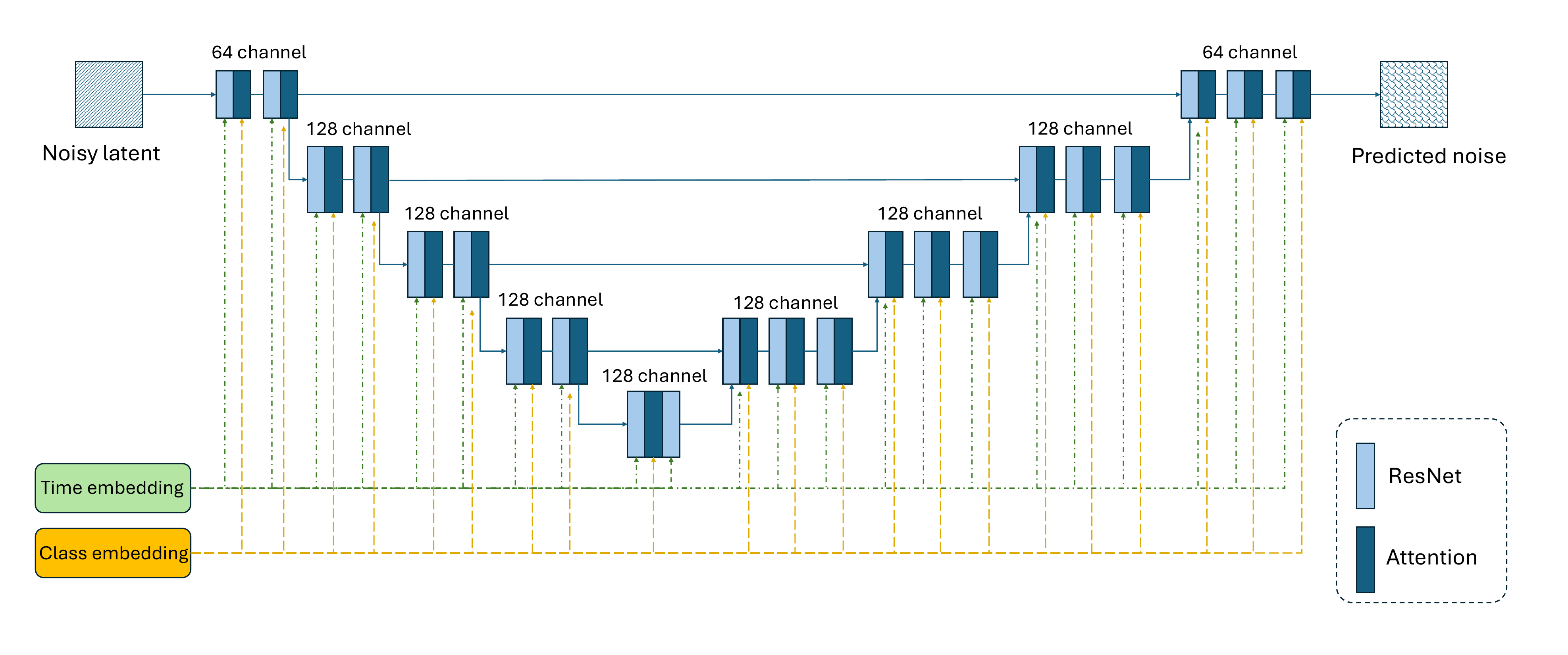}
    \caption{Classifier-free diffusion model structure.}
    \label{fig:unet}       
\end{figure}

The classifier-free diffusion model~\cite{ho2021classifier} that we used for image generation is adapted from a public codebase\footnote{\url{https://github.com/coderpiaobozhe/classifier-free-diffusion-guidance-Pytorch
}}. 
Its U-Net structure is illustrated in \Cref{fig:unet}, which contains 21 attention modules, where the number of input/output channels of the attention modules is either 64 or 128. 
We only fine-tune the query and value projection layers of those attention modules.
The total number of parameters of the U-Net base model is $10.42$M, including $300$ parameters for the class embedding.

\subsection{Illustration of grouping mechanism}
\label{appendix:grouping}

\begin{figure}[t]
    \centering
    \includegraphics[width=\linewidth, trim={50 50 50 10}, clip]{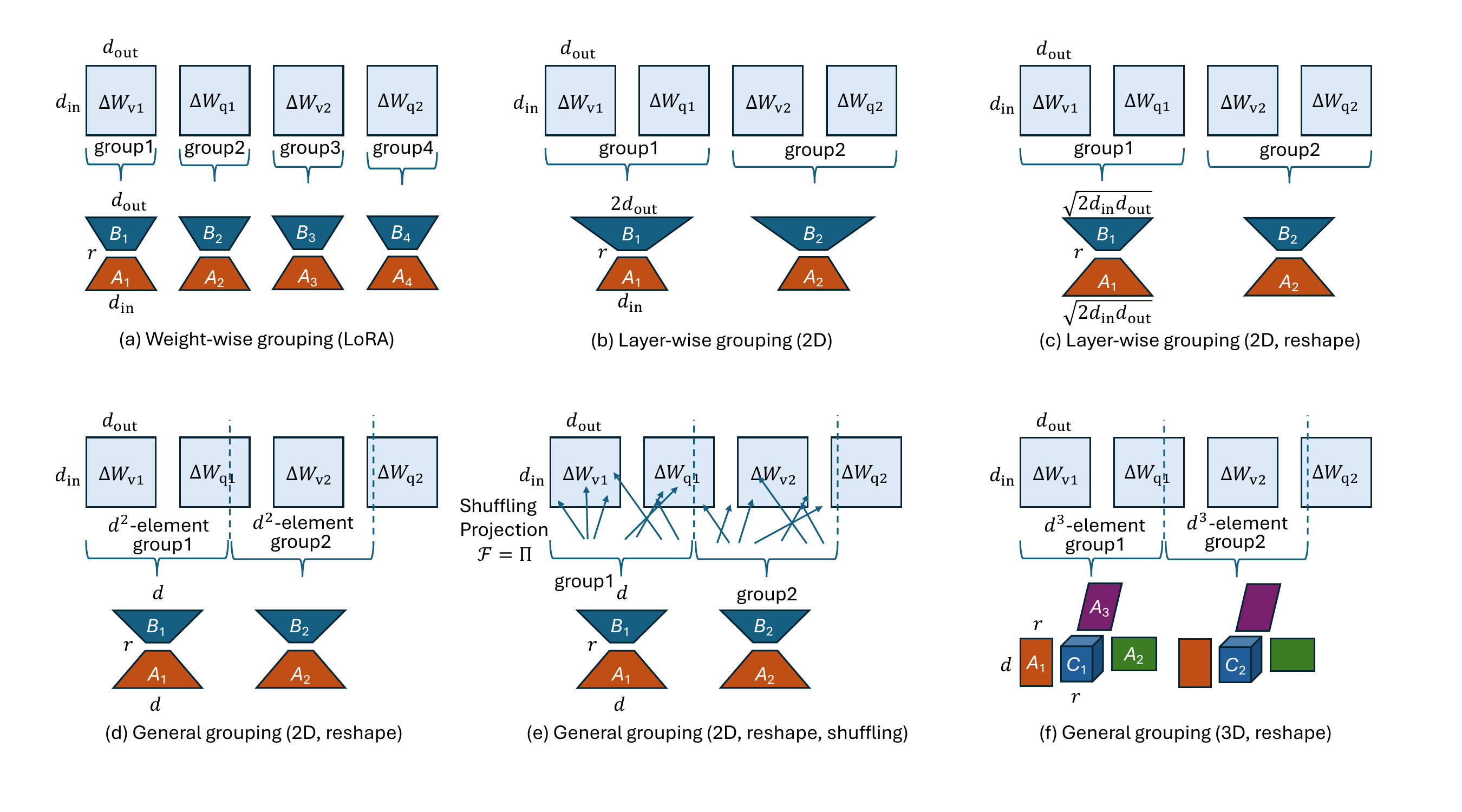}
    \caption{Examples of grouping mechanism.}
    \label{fig:grouping}       
\end{figure}

\Cref{fig:grouping} illustrates several different cases of the grouping mechanism.
\Cref{fig:grouping}(a) is the conventional weight-wise grouping, used for typical LoRA.
Each weight correction, \ie $\Delta W_{\mathrm{v}\ell}$ and $\Delta W_\mathrm{q\ell}$ for value and query projections at layer $\ell$, is individually represented by a rank-$r$ decomposition: $A_g B_g^\top$ for group $g$.
\Cref{fig:grouping}(b) shows layer-wise grouping, where the LoRA unit in each group jointly adapts both value and query projections in each layer.
When we stack multiple weight matrices in a na\"{i}ve way, the 2D array will have unbalanced fan-in/fan-out shape, leading to inefficient low-rank decomposition.
\Cref{fig:grouping}(c) can solve this issue by reshaping the 2D array into a regular square shape before low-rank decomposition.
As the reshaping is already breaking the geometric meaning of the original 2D weights, the grouping need not necessarily aligned with the weight boundary as shown in a general grouping case of \Cref{fig:grouping}(d).
Further, applying a projection function $\mathcal{F}(\cdot)$ as shown in \Cref{fig:grouping}(e), the element distribution can be shuffled and mixed-up to relax the geometric restriction of original LoRA.
LoRTA can further generalize the reshaping by folding the 2D array to any arbitrary $M$-dimensional tensor array by using the Tucker decomposition as shown in \Cref{fig:grouping}(f).
Relaxing the geometric constraint can improve the parameter efficiency as shown in this paper.
We further make a geometric analysis of our grouping methods below.

\subsection{Geometric analysis}
\label{appendix:geometric}
SuperLoRA adapts multiple attention modules at once, and relaxes the underlying geometric restrictions inherent to the 2D weights for each attention module, by employing grouping, reshaping, and projection (including shuffling).
To better understand how SuperLoRA works differently from LoRA, geometric analysis is conducted for the classification task. 
Specifically, we pick 4 different methods with a comparable number of parameters around $100{,}000$:
\begin{itemize}
    \item LoRA (2D): $\#$param $147{,}456$, accuracy $0.9113$;
    \item SuperLoRA (2D): $\#$param $115{,}200$, accuracy $0.9170$;
    \item SuperLoRA (2D, reshape): $\#$param $165{,}572$, accuracy $0.9218$;
    \item SuperLoRA (2D, reshape, w/ projection): $\#$param $138{,}372$, accuracy $0.9213$.
\end{itemize}
The weight correction term $\Delta W$ is compared to the full dense FT case, which involves $14$M parameters achieving an accuracy of $0.9290$.
We analyze three different geometric measures with respective to the FT weight $\Delta W_\mathrm{dense}$: 
i) left-singular similarity; ii) right-singular similarity; and iii) Euclidean distance. 
Letting $U$ and $V$ denote the left- and right-singular vectors of $\Delta W$ for each variant listed above, these metrics are defined as follows:
\begin{align}
    d_\mathrm{L} &= 
    \frac{1}{\sqrt{k}} \| U_\mathrm{dense}[:,:k]^\top U_\mathrm{variant}[:,:k]\|_2,\\
    d_\mathrm{R} &= 
    \frac{1}{\sqrt{k}} \|V_\mathrm{dense}[:k,:]V_\mathrm{variant}[:k,:]^\top\|_2, \\
    d_\mathrm{E} &= 
    \frac{\|\Delta W_\mathrm{dense}-\Delta W_\mathrm{variant}\|_2}{\|\Delta W_\mathrm{dense}\|_2}.
\end{align}
Note that $d_\mathrm{E}$ approaches to $0$ when $\Delta W$ converges to the dense FT case, while $d_\mathrm{L}$ and $d_\mathrm{R}$ converge to $1$.

\begin{figure}[t]
    \centering
    \includegraphics[width=\linewidth, trim={250 100 250 0},clip]{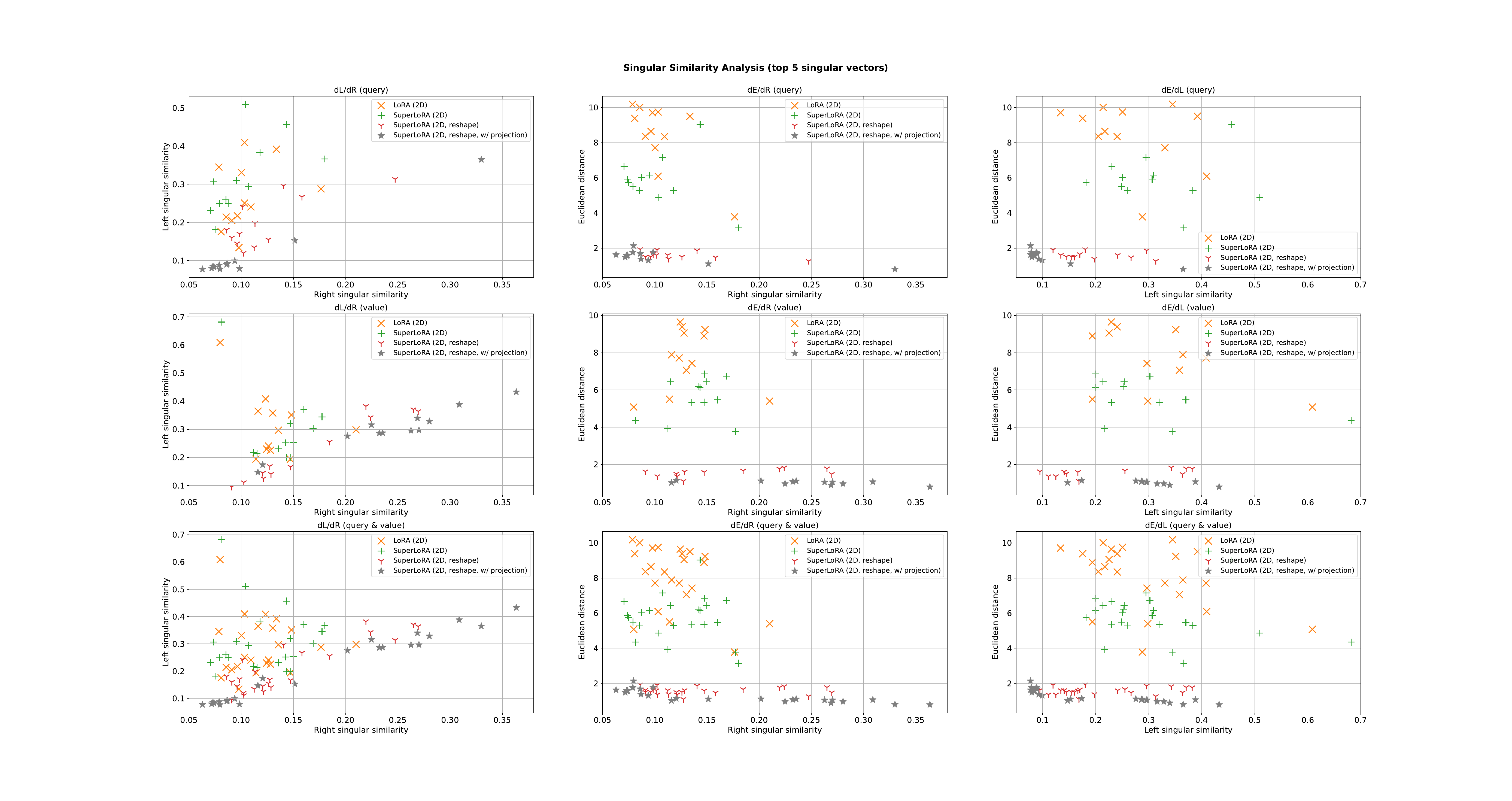}
    \caption{Geometric similarity analysis (top 5 principal singular vectors).}
    \label{fig:similarity_top5}       
\end{figure}

The top $k=5$ principal singular vectors are analyzed as shown in \Cref{fig:similarity_top5}.
The ViT model has $12$ attention modules, and we plot the total of $24$ points for the query and value projection weights.
The first row shows the query weights for $d_\mathrm{L}$ \vs $d_\mathrm{R}$, $d_\mathrm{E}$ \vs $d_\mathrm{R}$, and $d_\mathrm{E}$ \vs $d_\mathrm{L}$ from left to right across the columns.
The second and third rows are for the value weights, and both query and value weights, respectively.

We see that the Euclidean distance $d_\mathrm{E}$ is significantly decreased for SuperLoRA, especially with reshaping applied. 
It explains the improved accuracy with reshaping.
Although grouping, reshaping, and projection can break the geometric meaning of the original 2D weights, the subspace similarity is not completely lost.
Especially for query weights, SuperLoRA shows higher right-singular similarity than LoRA. 
As the embedding vector passes through right-hand side of the weight, principal right-singular vectors perform as a low-rank subspace mapping of the input vector while the left-singular vectors work as mapping the subspace towards the output vector.
While SuperLoRA with reshaping tends to preserve higher right-singular similarity, LoRA tends to preserve higher left-singular similarity.
Further, it is found that the corrections for query and value weights behave differently with reshaping, \ie, right-singular similarities for the value weights are much larger than for query weights.

\subsection{Grouping effect on SuperLoRA (1D, dense, with projection)}
\begin{figure}[t]
    \centering
    \includegraphics[width=\linewidth]{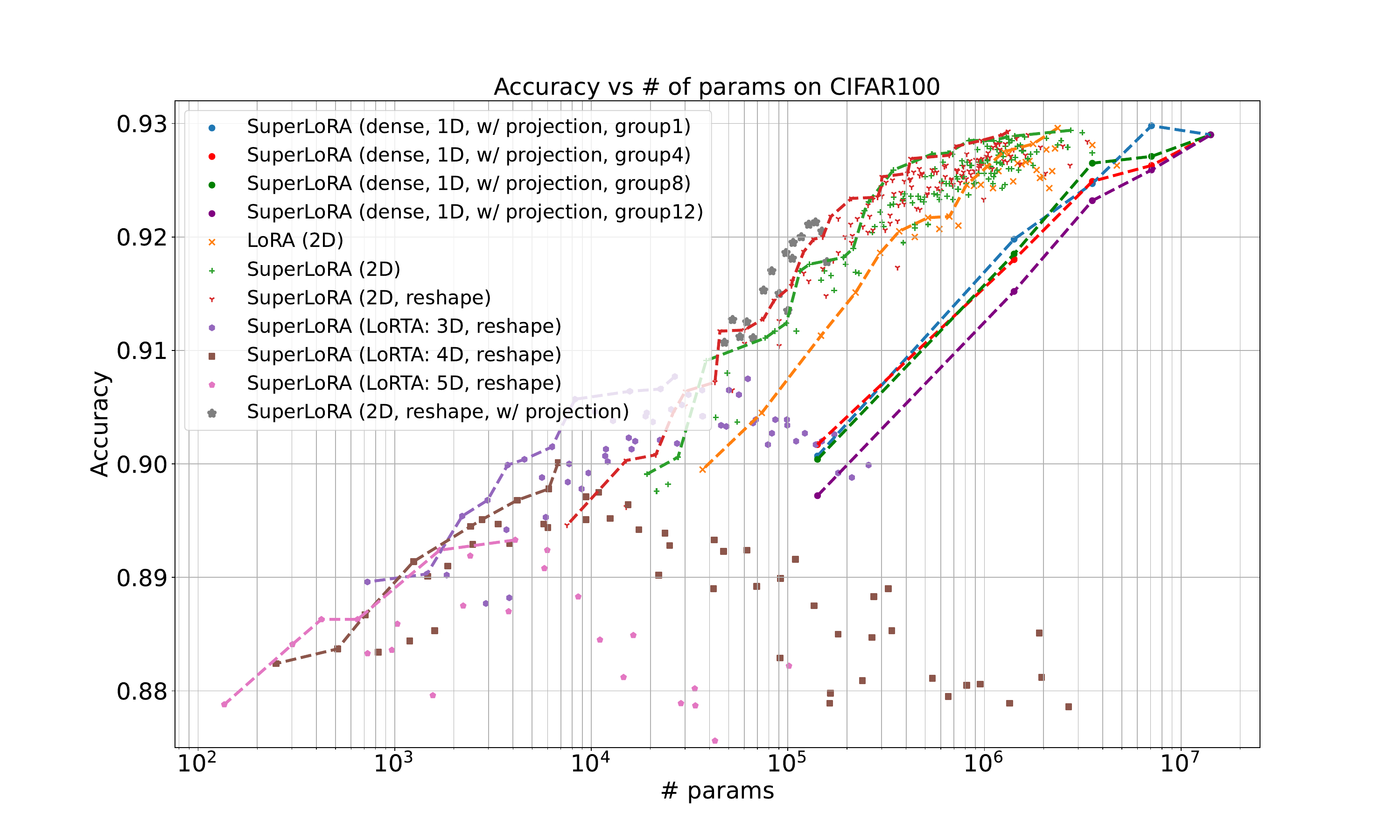}
    \caption{More groups (\ie less fixed projection parameters) on SuperLoRA (1D, dense, w/ projection).}
    \label{fig:dense_proj}       
\end{figure}

As the fixed projection matrix is shared across all groups, the number of groups will affect the size of the projection matrix directly. 
To explore this influence, dense FT with projection is tested for different splitting, from 1 to 12 groups. 
According to~\Cref{fig:dense_proj}, using 1 group achieves the best overall accuracy and using 4 or 8 groups are comparable to a smaller projection ratio. 
When the projection matrix is too small, \eg, with 12 groups, accuracy drops greatly.
This confirms that jointly updating multiple attention modules is beneficial.

\subsection{Linear \vs nonlinear projection}
\label{appendix:projection}

\begin{figure}[t]
    \centering
    \includegraphics[width=\linewidth]{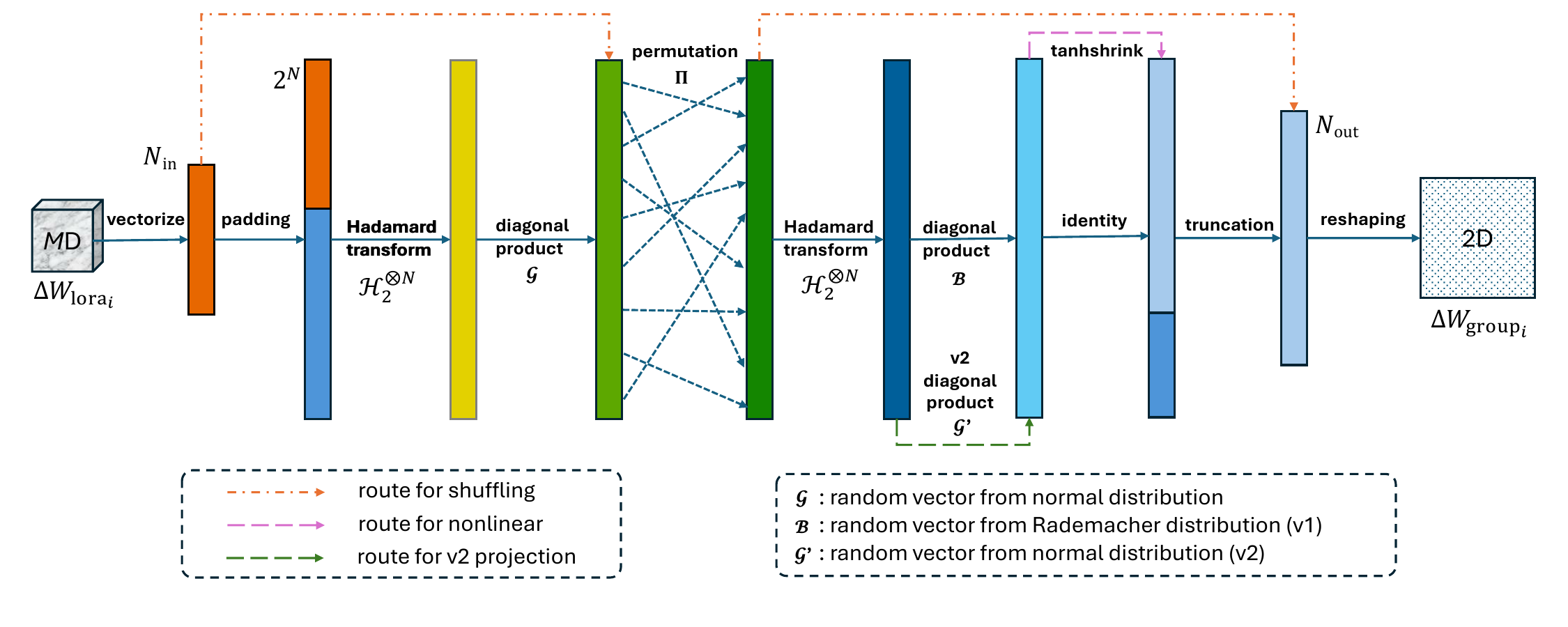}
    \caption{Illustration of fastfood projection and its variants.}
    \label{fig:fastfood}       
\end{figure}

Besides linear projection, nonlinear projection is also examined. 
We use tanhshrink after the fixed linear projection, resulting in `nonlinear' and `nonlinear\textsubscript{v2}'. 
Note that the `v2' projection uses a Gaussian random vector rather than a binary random vector $\mathcal{B}$ for the fastfood projection as shown in \Cref{fig:fastfood}.
More specifically, we consider six variants for the projection function $\mathcal{F}(\cdot)$ in this paper:
\begin{itemize}
\item identity (no projection): $\mathcal{F}(x) = x$;
\item shuffling: $\mathcal{F}(x) = x\,  \varPi$;
\item linear: $\mathcal{F}(x) = x\, \mathcal{H}'\, \mathsf{diag}[\mathcal{G}]\, \varPi\, \mathcal{H}\, \mathsf{diag}[\mathcal{B}]$;
\item linear\textsubscript{v2}: $\mathcal{F}(x) = x\, \mathcal{H}'\, \mathsf{diag}[\mathcal{G}]\, \varPi\, \mathcal{H}\, \mathsf{diag}[\mathcal{G}']$;
\item nonlinear: $\mathcal{F}(x) = \mathsf{tanhshrink}\big[ x\, \mathcal{H}'\, \mathsf{diag}[\mathcal{G}]\, \varPi\, \mathcal{H}\, \mathsf{diag}[\mathcal{B}] \big]$;
\item nonlinear\textsubscript{v2}: $\mathcal{F}(x) = \mathsf{tanhshrink}\big[ x\, \mathcal{H}'\, \mathsf{diag}[\mathcal{G}]\, \varPi\, \mathcal{H}\, \mathsf{diag}[\mathcal{G}'] \big]$.
\end{itemize}

Here, $\varPi$ performs a random permutation of a vector.
The Walsh--Hadamard matrices $\mathcal{H}'\in \mathbb{R}^{N_\mathrm{in} \times 2^N}$ and $\mathcal{H} \in \mathbb{R}^{2^N \times N_\mathrm{out}}$ are left- and right-truncated versions of a regular Walsh--Hadamard matrix $\mathcal{H}_2^{\otimes N} \in \mathbb{R}^{2^N\times 2^N}$, where $[\cdot]^{\otimes N}$ denotes $N$-fold Kronecker power and $\mathcal{H}_2 = \tfrac{1}{\sqrt{2}}\big[\begin{smallmatrix} 1 & 1\\ 1 & -1 \end{smallmatrix}\big]$.
Letting $N_\mathrm{in}$ and $N_\mathrm{out}$ be the number of elements for the input and output of the projection function $\mathcal{F}(\cdot)$ with a compression ratio of $\rho=N_\mathrm{in}/N_\mathrm{out}$, the exponent $N$ is chosen as $N=\mathrm{ceil}[\log_2(\max(N_\mathrm{in}, N_\mathrm{out}))]$.
In practice, the left-truncated Walsh--Hadamard matrix is realized by input zero-padding before fast Walsh--Hadamard transform.
The random vector $\mathcal{G}$ is hence of size $2^N$, and drawn from the normal distribution.
Here, $\mathcal{G}'\in \mathbb{R}^{N_\mathrm{out}}$ is another random vector drawn from the normal distribution while $\mathcal{B}\in \{\pm 1\}^{N_\mathrm{out}}$ is a random vector drawn from the Rademacher distribution.

\begin{figure}[t]
    \centering
    \begin{subfigure}{0.49\linewidth}
        \includegraphics[width=\linewidth]{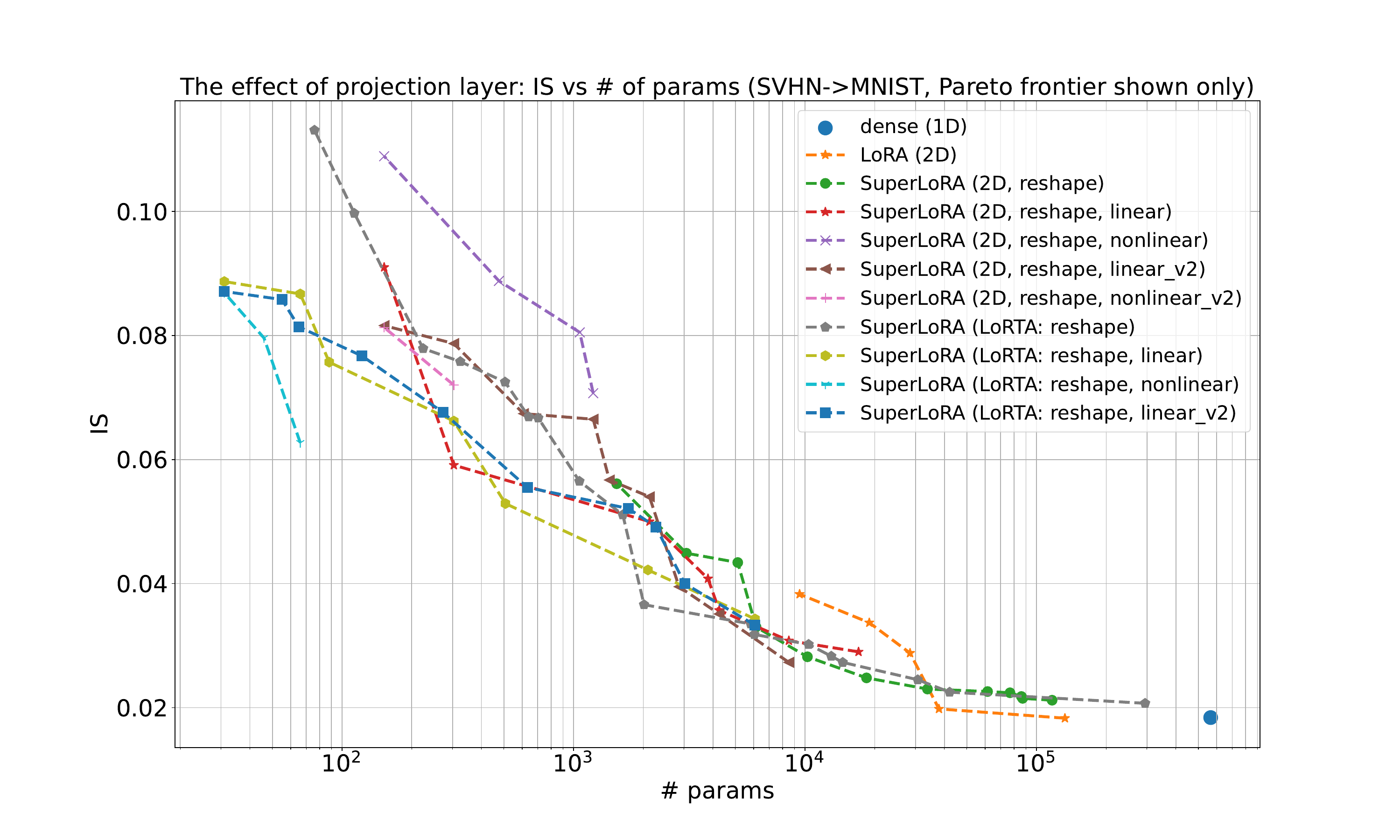}
        \caption{linear \vs nonlinear (IS, Pareto only)}
        \label{fig:fig_group_nonlinear_pareto}       
    \end{subfigure}
    \begin{subfigure}{0.49\linewidth}
        \includegraphics[width=\linewidth]{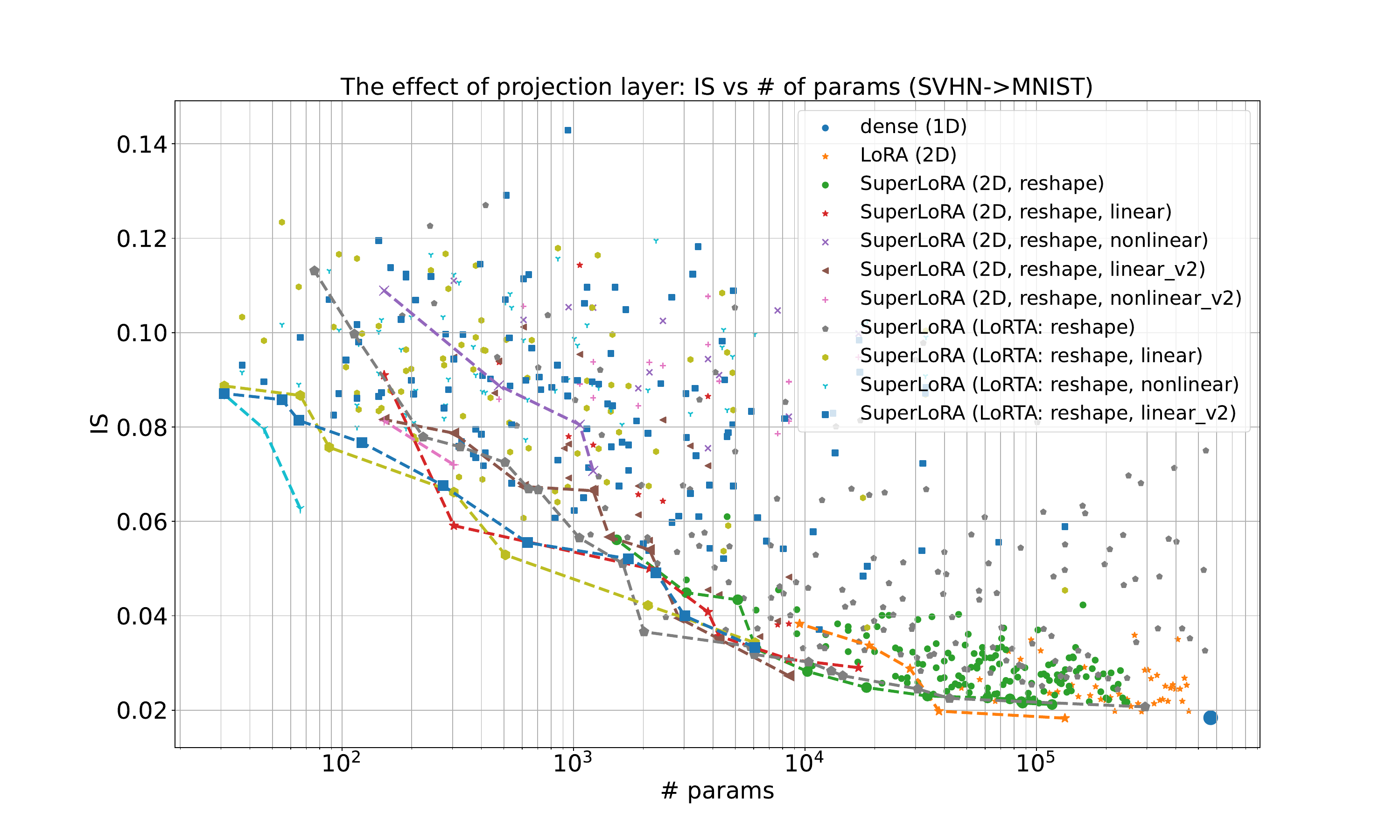}
        \caption{linear \vs nonlinear (IS, all points)}
        \label{fig:fig_nonlinear_is_all}       
    \end{subfigure}
    \caption{Comparison between Linear/Linear\textsubscript{v2}/Nonlinear/Nonlinear\textsubscript{v2} projections.}
    \label{fig:fig_nonlinear_is}       
\end{figure}

\Cref{fig:fig_nonlinear_is} shows the comparison of several projection variants.
Surprisingly, with the same number of parameters, the linear version outperforms the nonlinear versions in most cases.

\subsection{Transfer learning from ImageNet1k to CIFAR10}
\label{appendix:cifar10}

\begin{figure}[t]
    \centering
    \includegraphics[width=\linewidth]{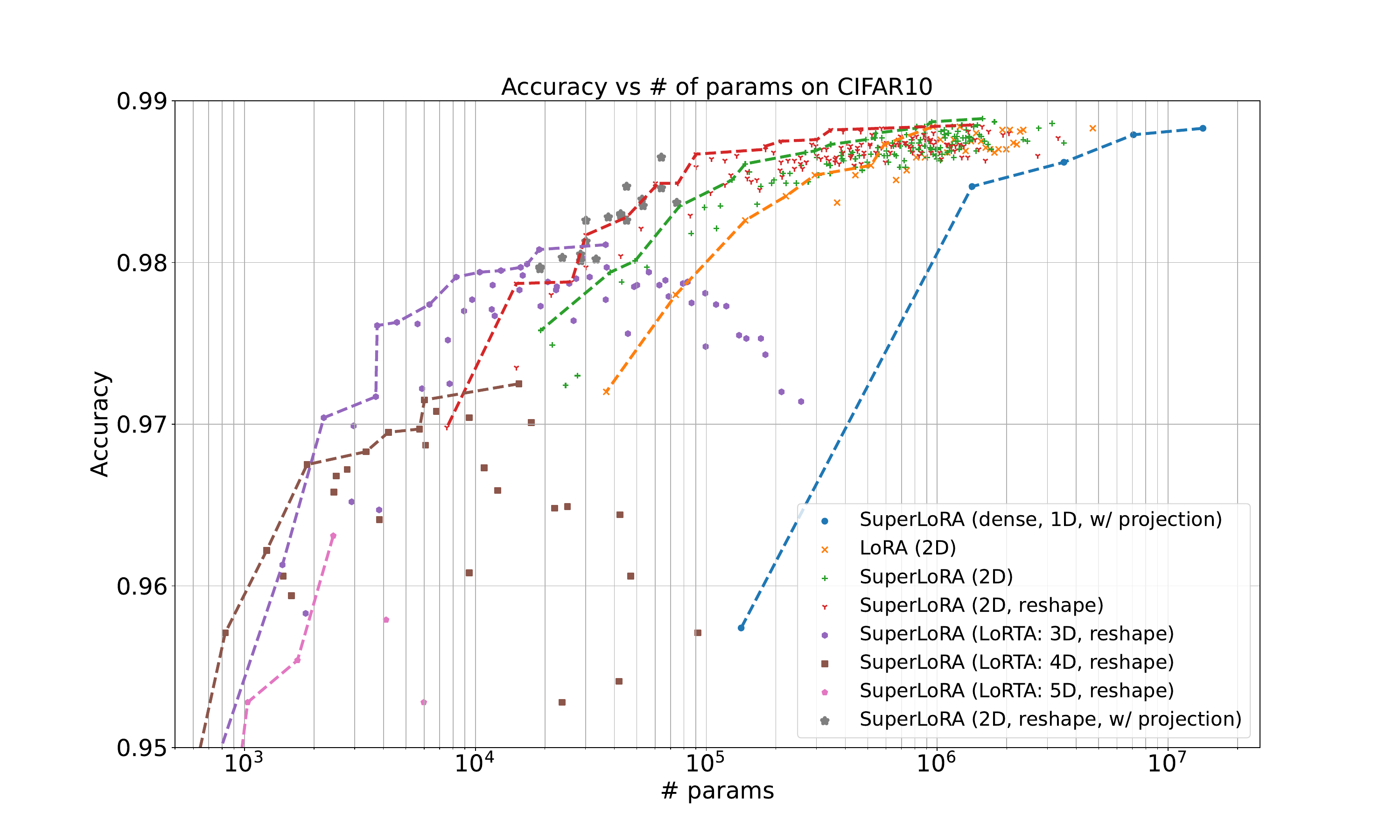}
    \caption{Classification accuracy for transfer learning from ImageNet1K to  CIFAR10 with SuperLoRA.}
    \label{fig:classification_cifar10}       
\end{figure}

For the classification task, transfer learning from ImageNet1k to CIFAR10 is also examined. 
Most settings are same as \Cref{fig:dense_proj} for transfer learning from ImageNet21k to CIFAR100. 
The ViT base model\footnote{\url{https://huggingface.co/google/vit-base-patch16-224}} is pretrained for ImageNet1k.
It fine-tunes with 3000 steps at most and the best accuracy is reported. 
Detailed ranks tested are as follows:
\begin{itemize}
    \item LoRA (2D): ranks: 1, 2, 4, 6, 8, \ldots, 64, 128
    \item SuperLoRA (2D): groups: 1, 4, 8, 12; ranks: 1, 2, 4, 6, 8, \ldots, 64, 128
    \item SuperLoRA (2D, reshape): 
        \begin{itemize}
            \item groups: 1, 4, 12, ranks: 1, 2, 4, 6, 8, \ldots, 64, 128;
            \item group 8, ranks: 1, 2, 4, 6, 8, \ldots, 24, 28, 32, 36, \ldots, 64
        \end{itemize} 
    \item SuperLoRA (LoRTA: 3D, reshape): groups: 1, 4, 8, 12; ranks: 1--6, 8, 10, 12, \ldots, 24
     \item SuperLoRA (LoRTA: 4D, reshape): 
     \begin{itemize}
         \item group 1; ranks: 1--6, 8, 10, 12, \ldots, 22
         \item group 4; ranks: 1--6, 8, 10, 12, \ldots, 16
        \item group 8; ranks: 1--6, 8, 10, 12, \ldots, 18
         \item group 12; ranks: 1--6, 8, 10, 12
     \end{itemize}
    \item SuperLoRA (LoRTA: 5D, reshape): 
     \begin{itemize}
         \item groups 1, 4, 8; ranks: 1--6, 8
         \item group 12; ranks: 1--6
     \end{itemize}
\end{itemize}
The classifier head is frozen after selecting most relevant labels in ImageNet1k, \ie [404, 436, 94, 284, 345, 32, 340, 510, 867], corresponding to [airliner, humming bird, siamese cat, ox, golden retriever, tailed frog, zebra, container ship, trailer truck]. 
Classification results can be found in \Cref{fig:classification_cifar10}. 
Even though only attention modules are adapted, overall transfer learning is excellent, reaching an accuracy close to $0.99$. 
Besides, SuperLoRA significantly outperforms original LoRA in terms of both classification accuracy and the parameter range it covers as the transfer learning.
SuperLoRA (2D, reshape) shows at least 3-fold reduction in the required number of parameters compared to LoRA.
Noticeably, when comparing the lowest-rank LoRA with around $0.97$ accuracy, SuperLoRA (2D, reshape, w/ projection) improves the accuracy by about $1\%$, and moreover the required number of parameters can be greatly reduced by $10$ folds with SuperLoRA (LoRTA: 3D, reshape) to maintain the comparable accuracy.

\subsection{Transfer learning from SVHN to MNIST}\label{appendix:2mnist}
\subsubsection{Grouping effect (complete results)}
\begin{figure}[t]
    \centering
    \begin{subfigure}{0.49\linewidth}
        \includegraphics[width=\linewidth]{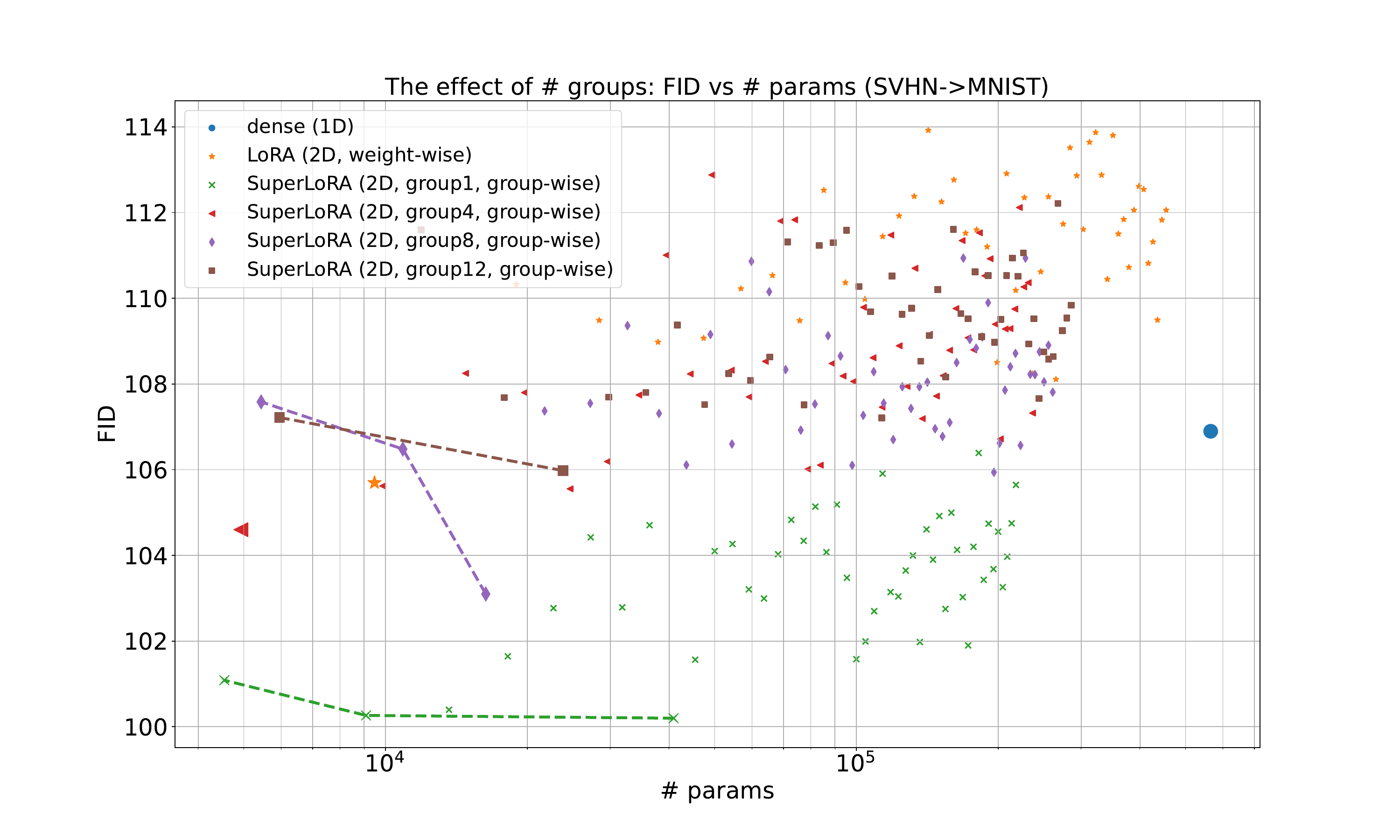}
        \caption{weight-wise \vs group-wise (FID)}
    \end{subfigure}
    \begin{subfigure}{0.49\linewidth}
        \includegraphics[width=\linewidth]{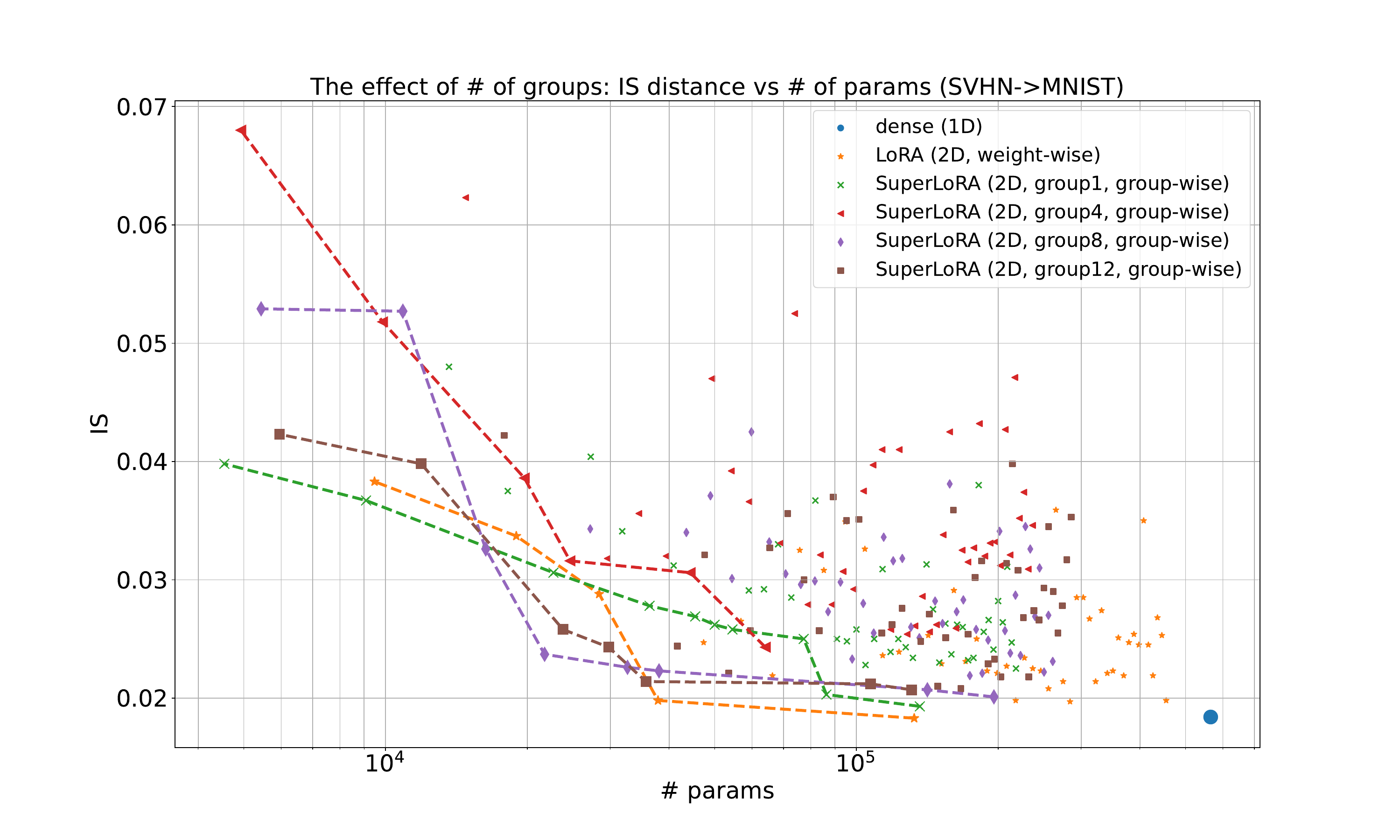}
        \caption{weight-wise \vs group-wise (IS)}
    \end{subfigure}
   \begin{subfigure}{0.49\linewidth}
        \includegraphics[width=\linewidth]{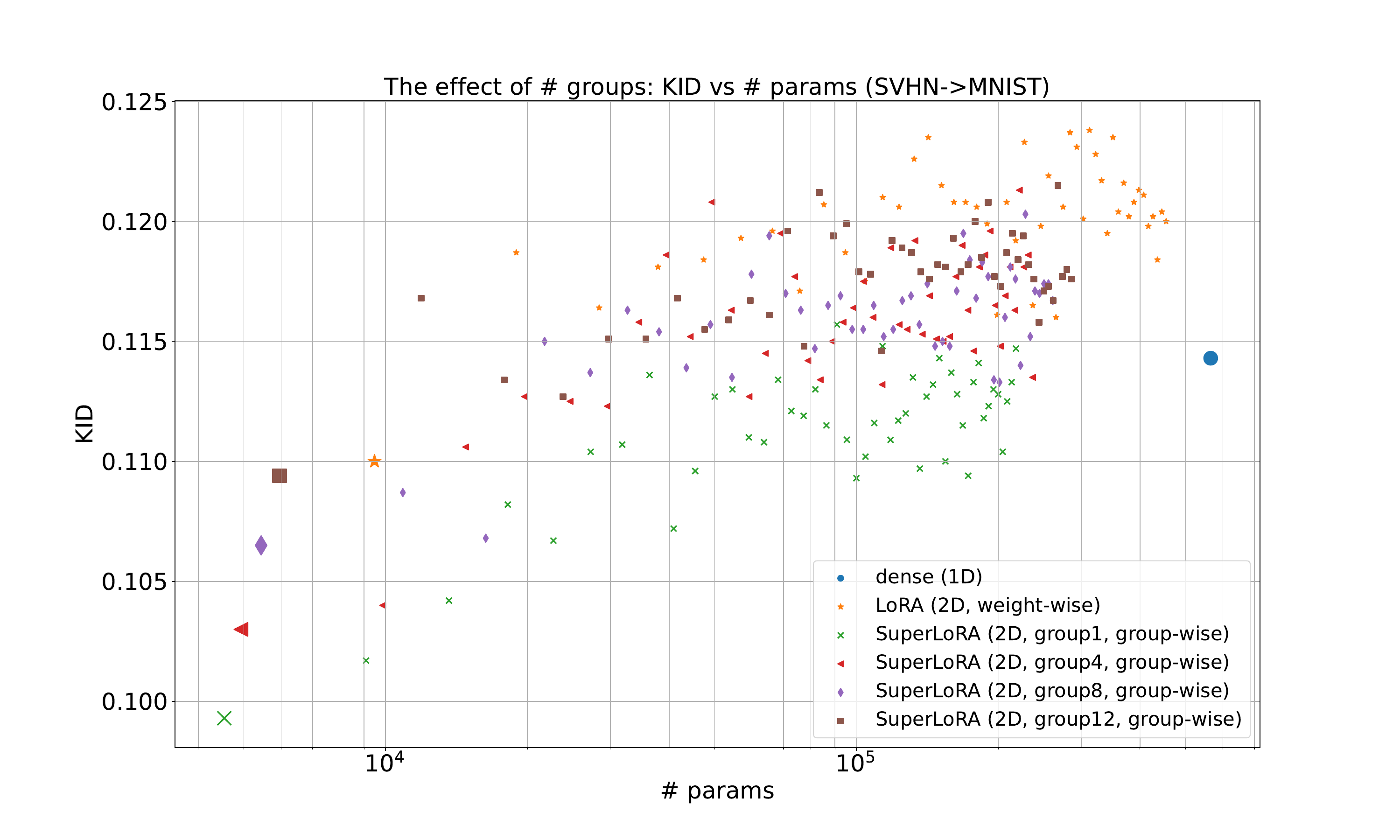}
        \caption{weight-wise \vs group-wise (KID)}
    \end{subfigure}
    \begin{subfigure}{0.49\linewidth}
        \includegraphics[width=\linewidth]{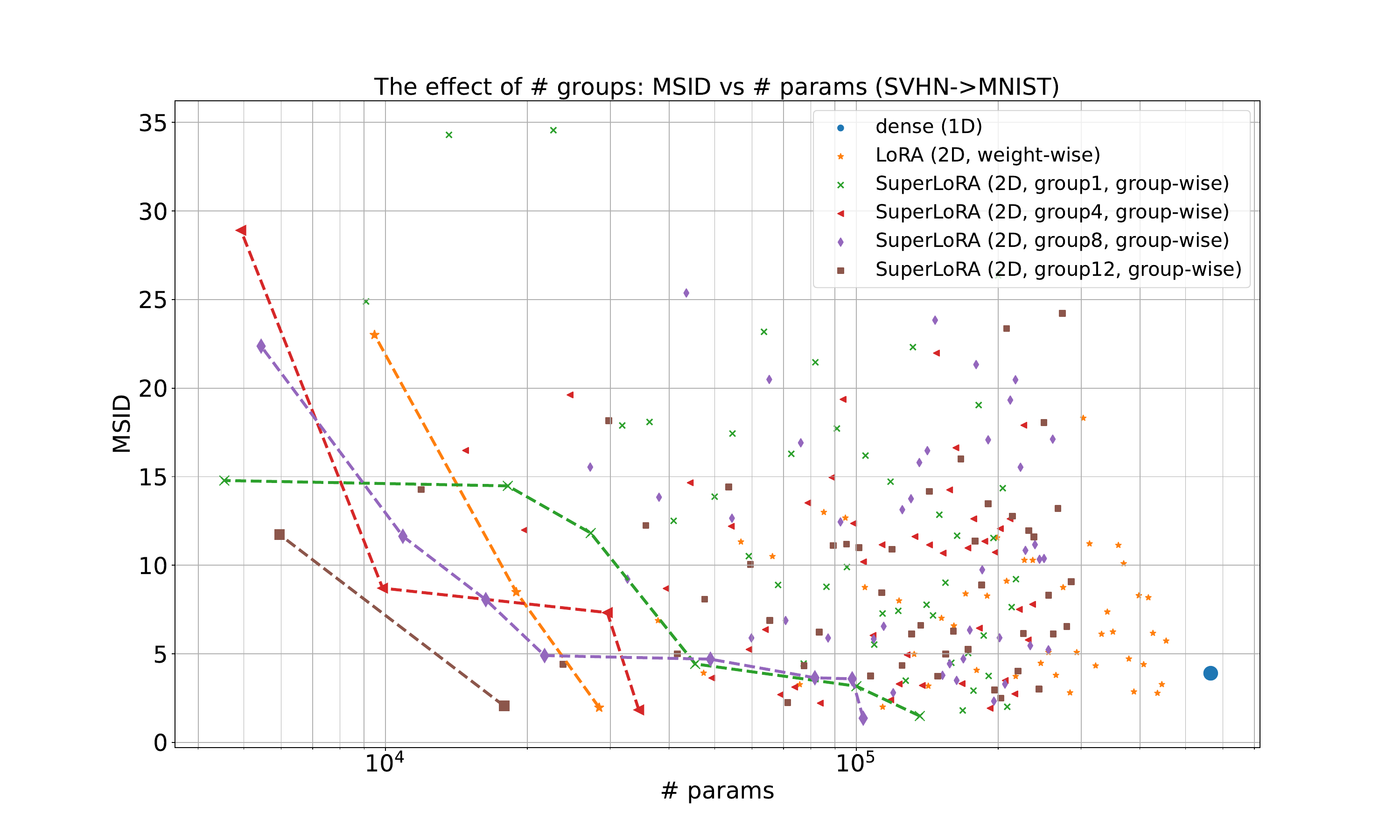}
        \caption{weight-wise \vs group-wise (MSID)}
    \end{subfigure}    
       \begin{subfigure}{0.49\linewidth}
        \includegraphics[width=\linewidth]{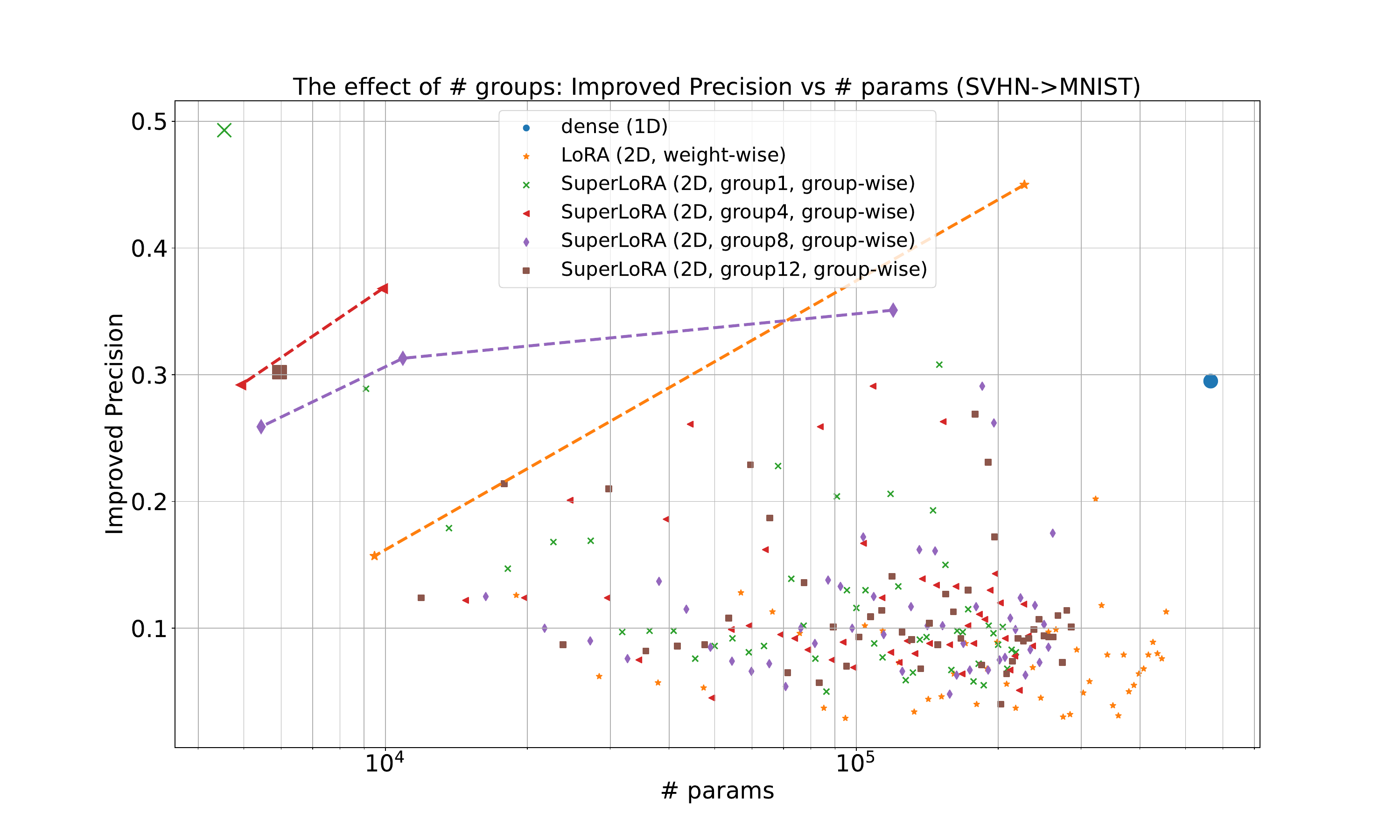}
        \caption{weight-wise \vs group-wise (Improved Precision)}
    \end{subfigure}
    \begin{subfigure}{0.49\linewidth}
        \includegraphics[width=\linewidth]{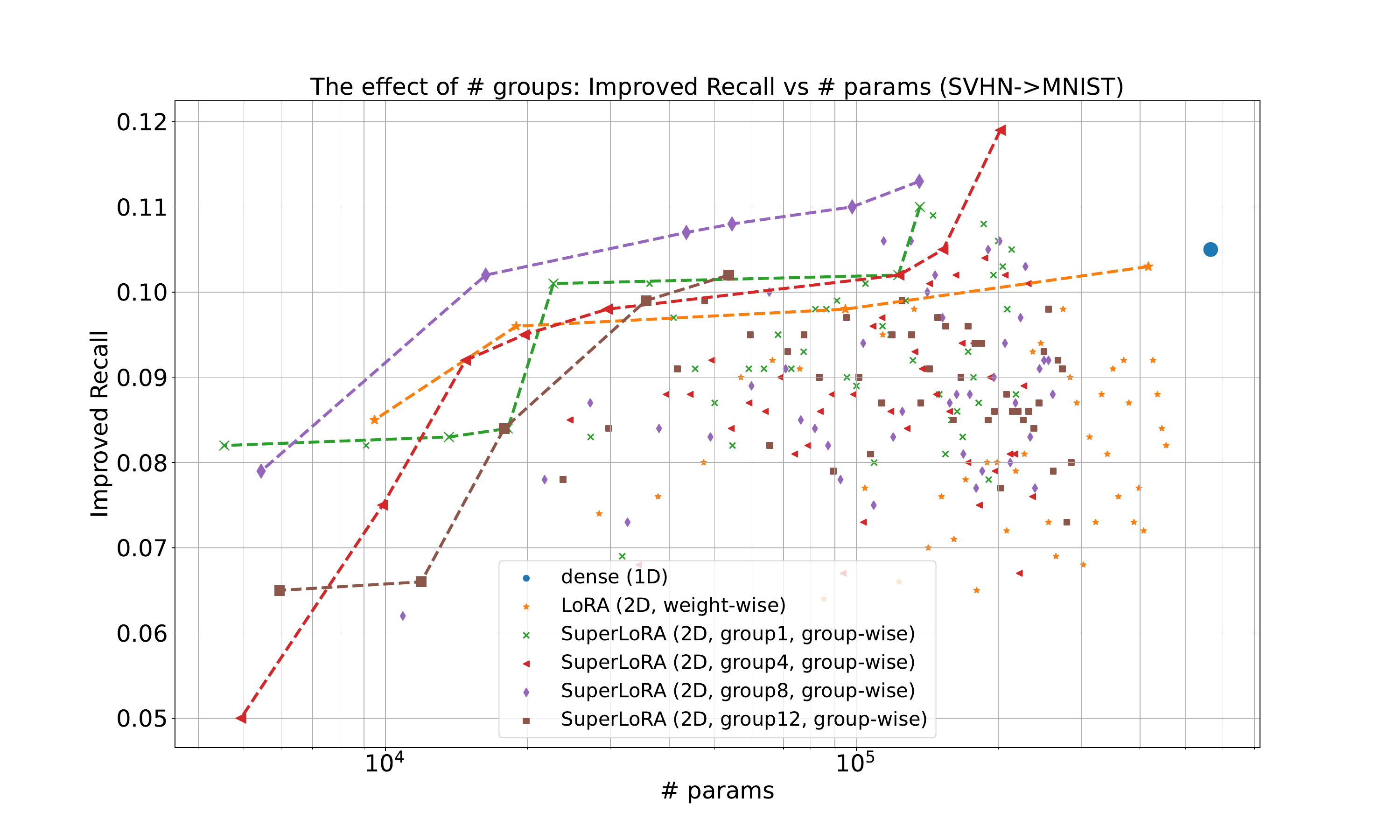}
        \caption{weight-wise \vs group-wise (Improved Recall)}
    \end{subfigure}    
    \caption{Complete comparison between weight-wise LoRA and group-wise SuperLoRA.}
    \label{fig:2mnist_group}
\end{figure}

Scatter plots of all metrics (FID, IS, KID, MSID, Improved Precision and Improved Recall) are given in~\Cref{fig:2mnist_group}. Except IS, all metrics show many examples performing better than dense FT, while worse according to the visualization results, indicating IS is a more reasonable quantitative metric in this case.

\subsubsection{Reshaping effect (complete results)}
\begin{figure}[t]
    \centering
    \begin{subfigure}{0.49\linewidth}
        \includegraphics[width=\linewidth]{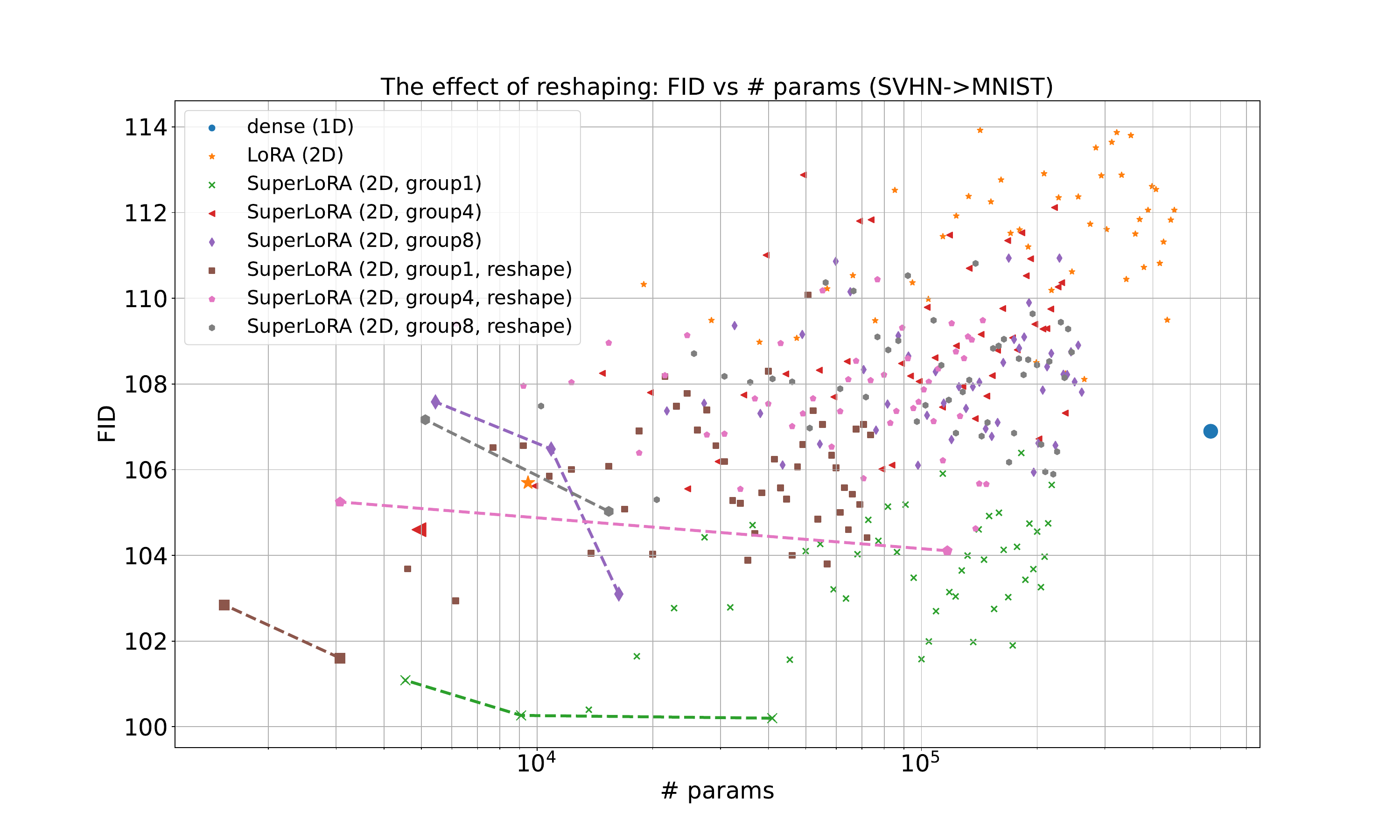}
        \caption{reshaping \vs non-reshaping (FID)}
    \end{subfigure}
    \begin{subfigure}{0.49\linewidth}
        \includegraphics[width=\linewidth]{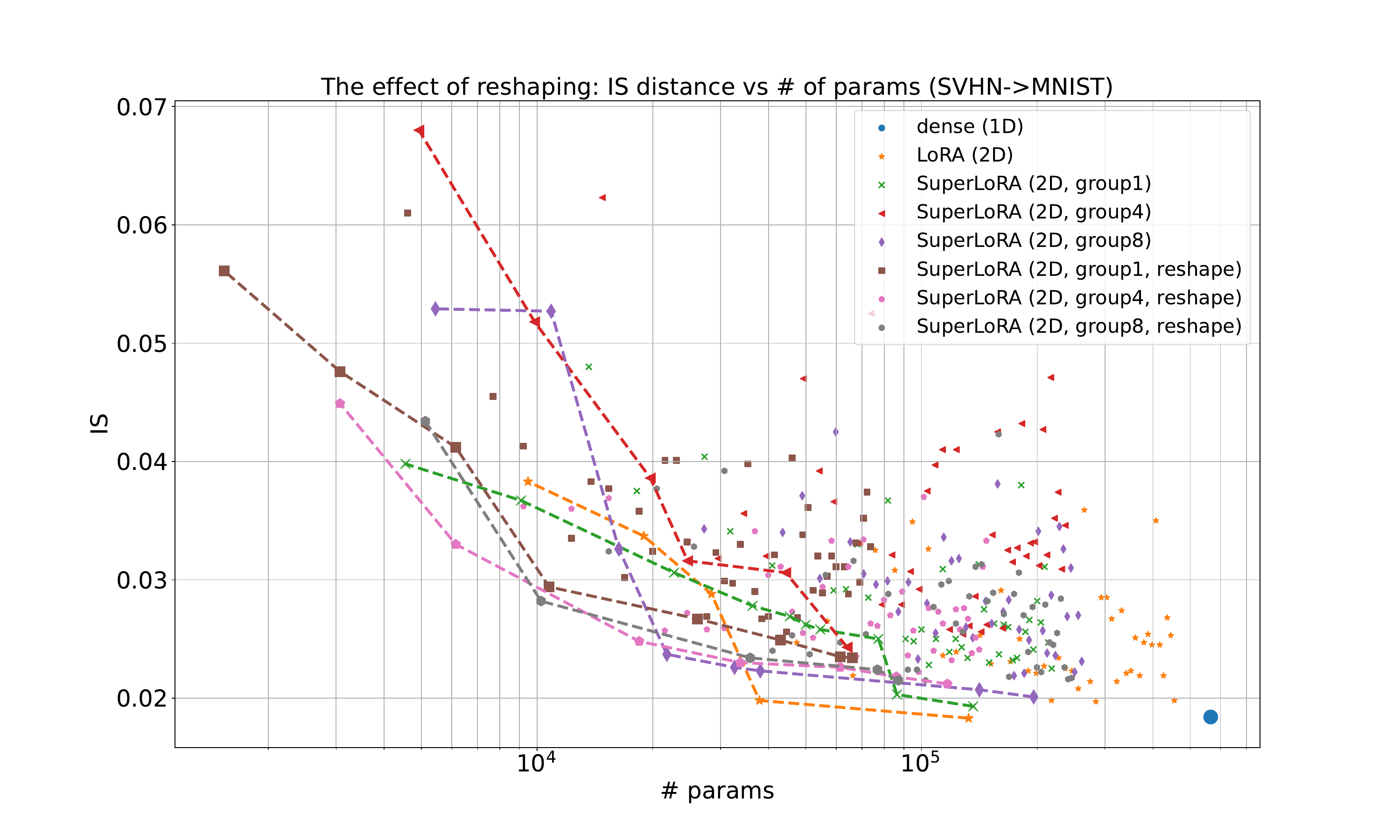}
        \caption{reshaping \vs non-reshaping (IS)}
    \end{subfigure}
   \begin{subfigure}{0.49\linewidth}
        \includegraphics[width=\linewidth]{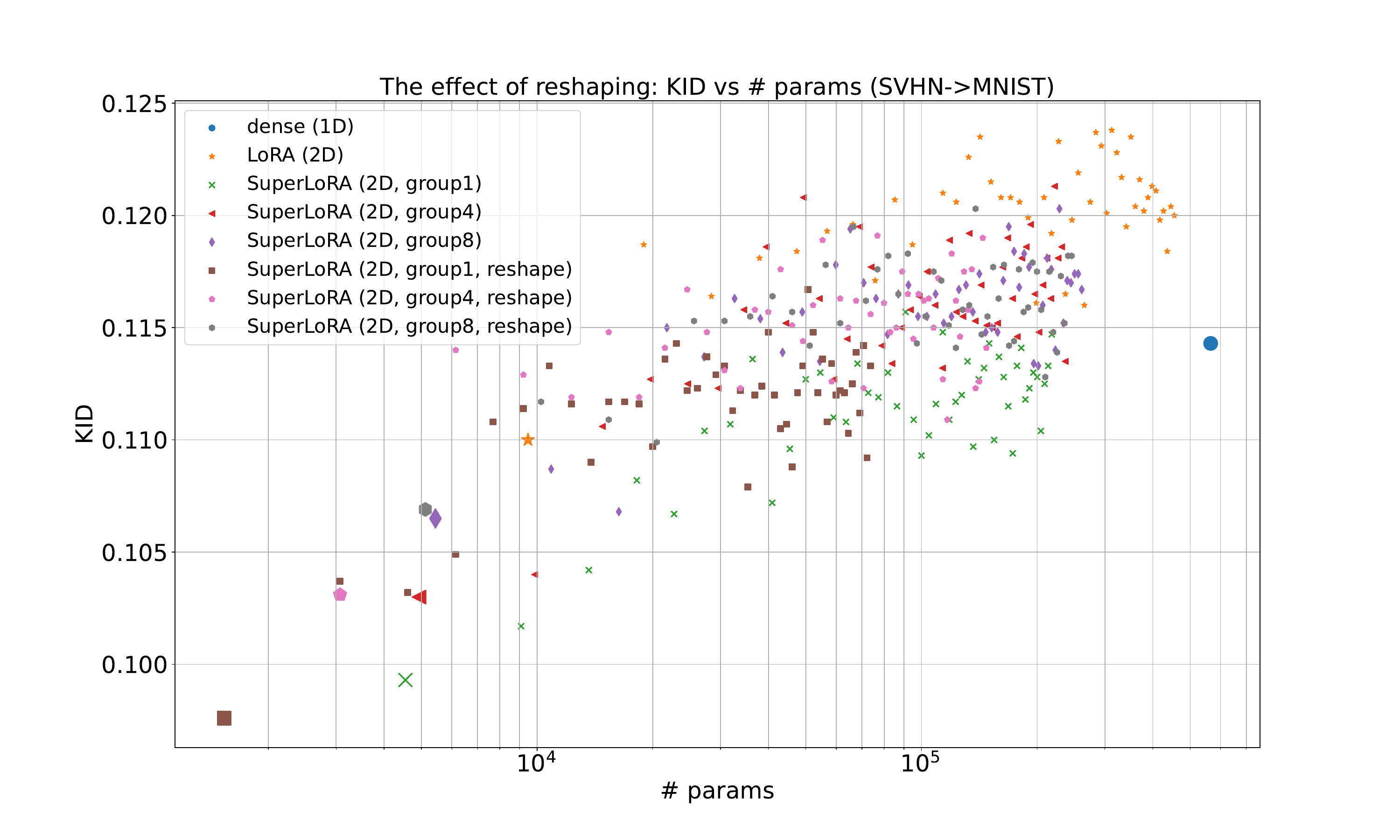}
        \caption{reshaping \vs non-reshaping (KID)}
    \end{subfigure}
    \begin{subfigure}{0.49\linewidth}
        \includegraphics[width=\linewidth]{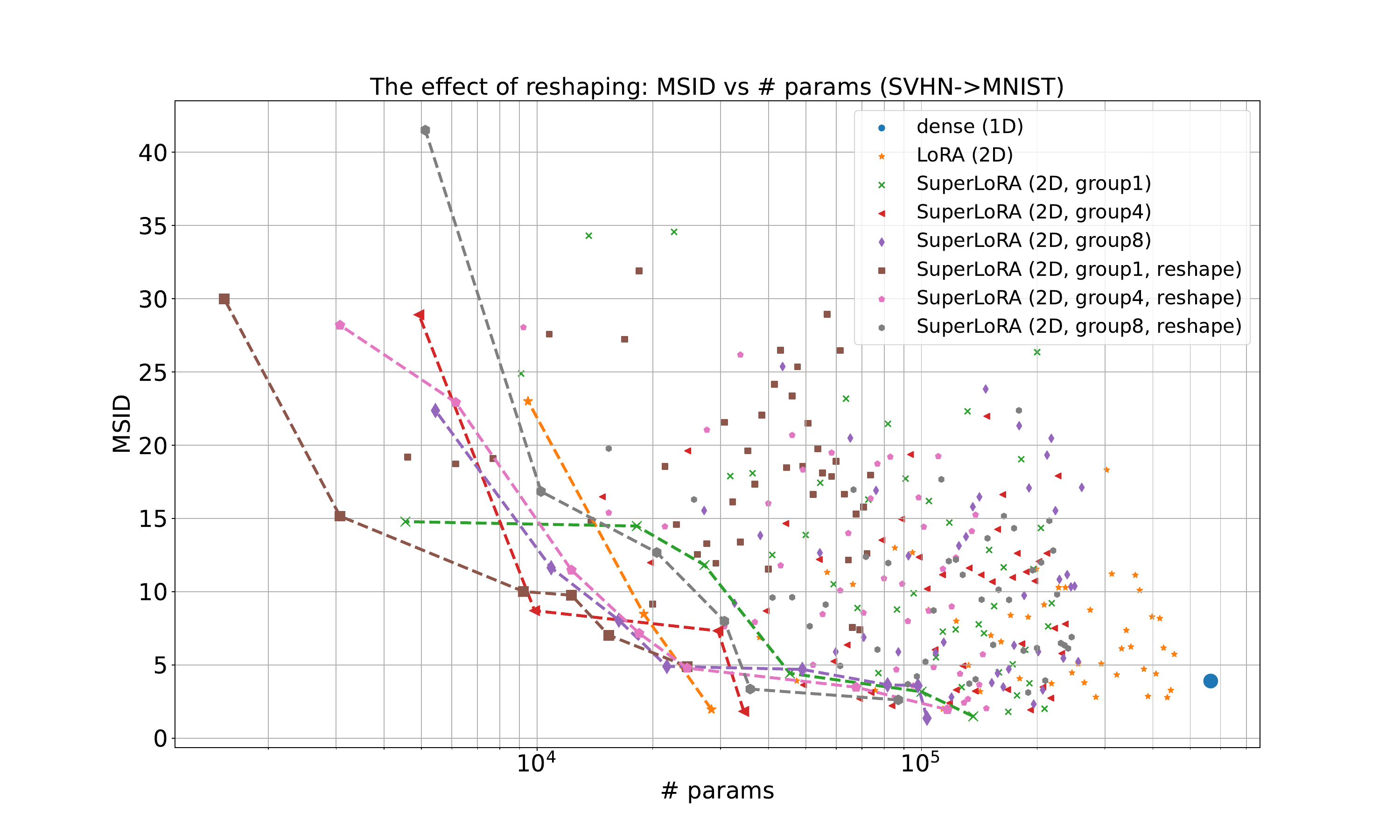}
        \caption{reshaping \vs non-reshaping (MSID)}
    \end{subfigure}    
       \begin{subfigure}{0.49\linewidth}
        \includegraphics[width=\linewidth]{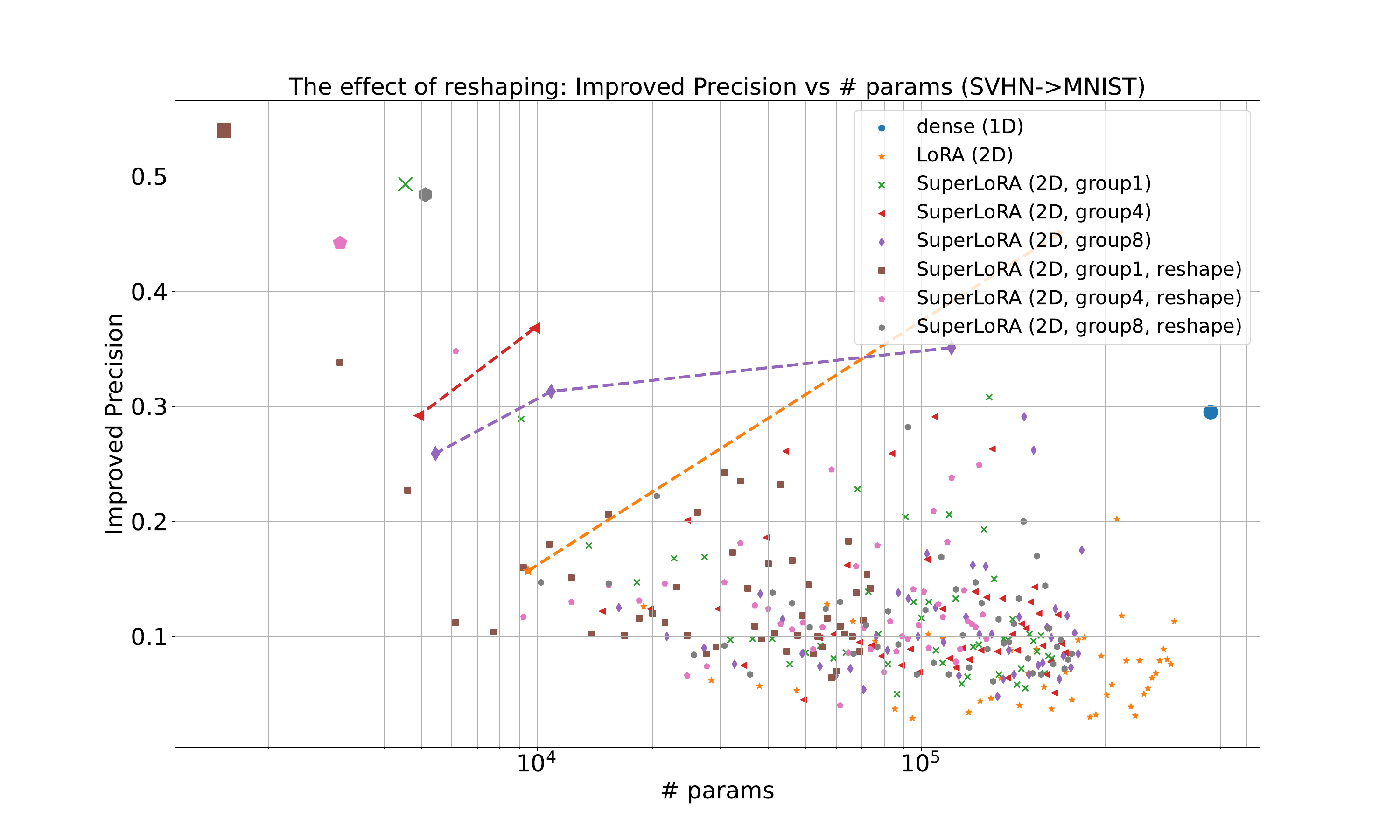}
        \caption{reshaping \vs non-reshaping (Improved Precision)}
    \end{subfigure}
    \begin{subfigure}{0.49\linewidth}
        \includegraphics[width=\linewidth]{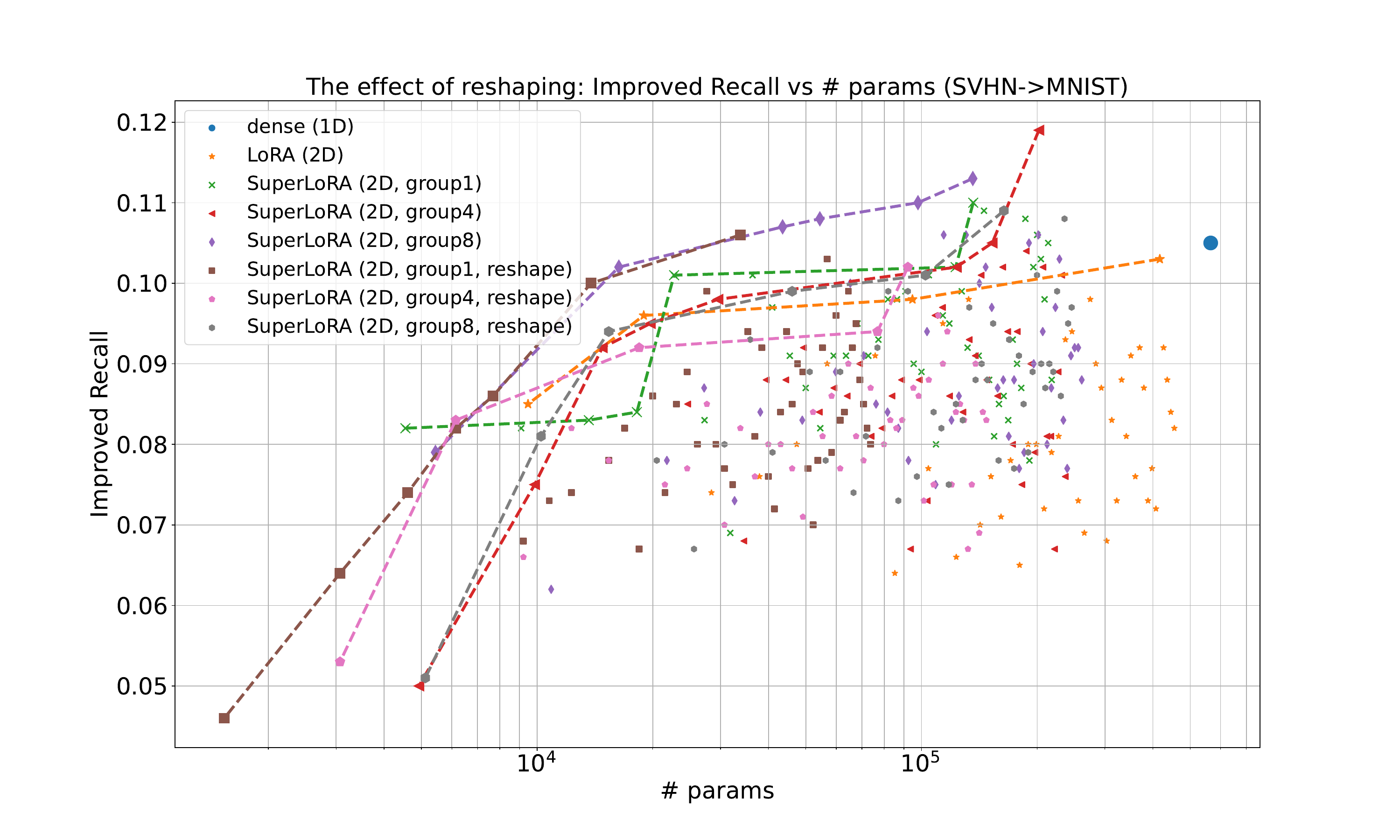}
        \caption{reshaping \vs non-reshaping (Improved Recall)}
    \end{subfigure}    
    \caption{Complete comparison between reshaping and non-reshaping SuperLoRA.}
    \label{fig:2mnist_reshape}
\end{figure}

Complete results for reshaping, with scatter plots for all metrics, are shown in \Cref{fig:2mnist_reshape}.
\subsubsection{SuperLoRA (LoNKr, complete results)}
\begin{figure}[t]
    \centering
    \begin{subfigure}{0.49\linewidth}
        \includegraphics[width=\linewidth]{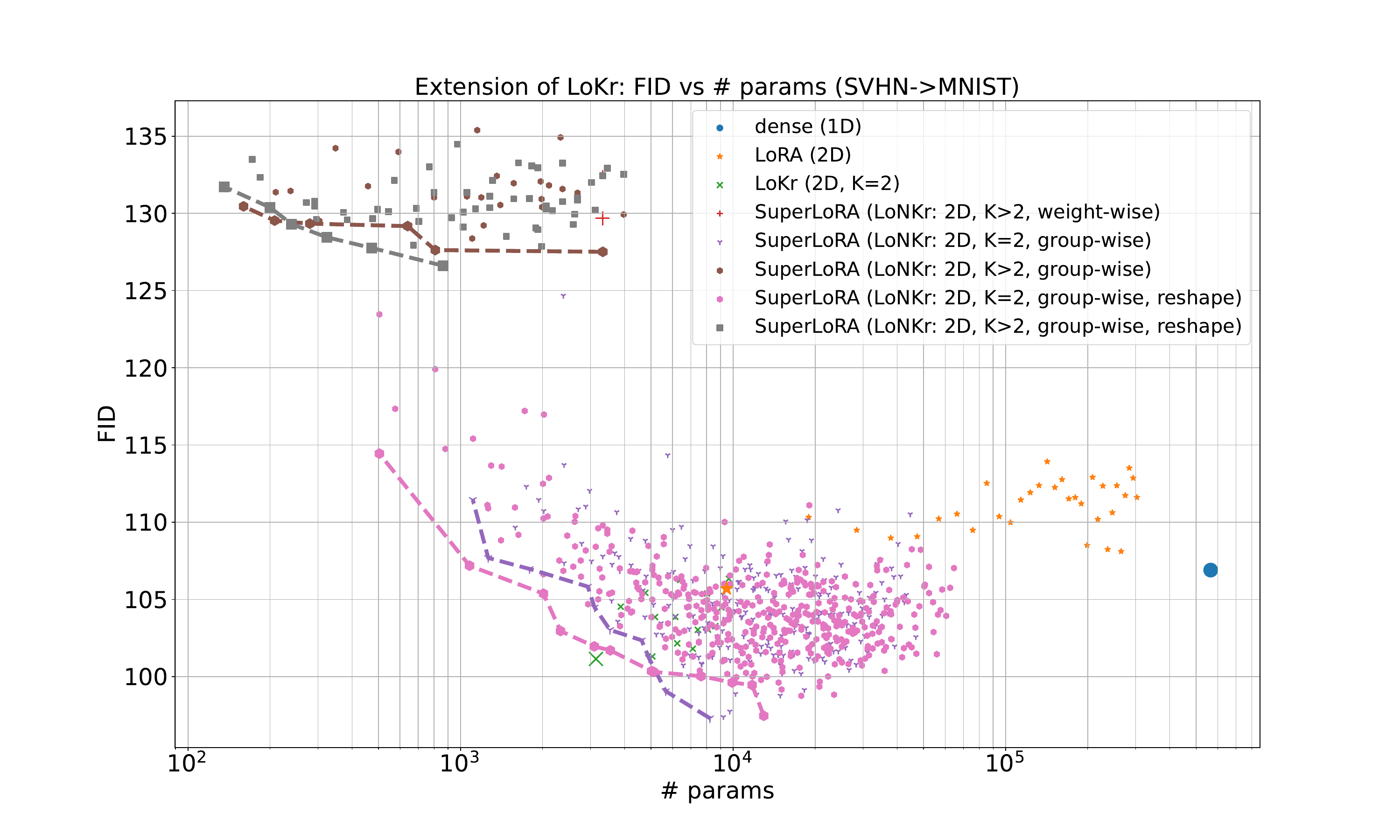}
        \caption{SuperLoRA (LoNKr, FID)}
    \end{subfigure}
    \begin{subfigure}{0.49\linewidth}
        \includegraphics[width=\linewidth]{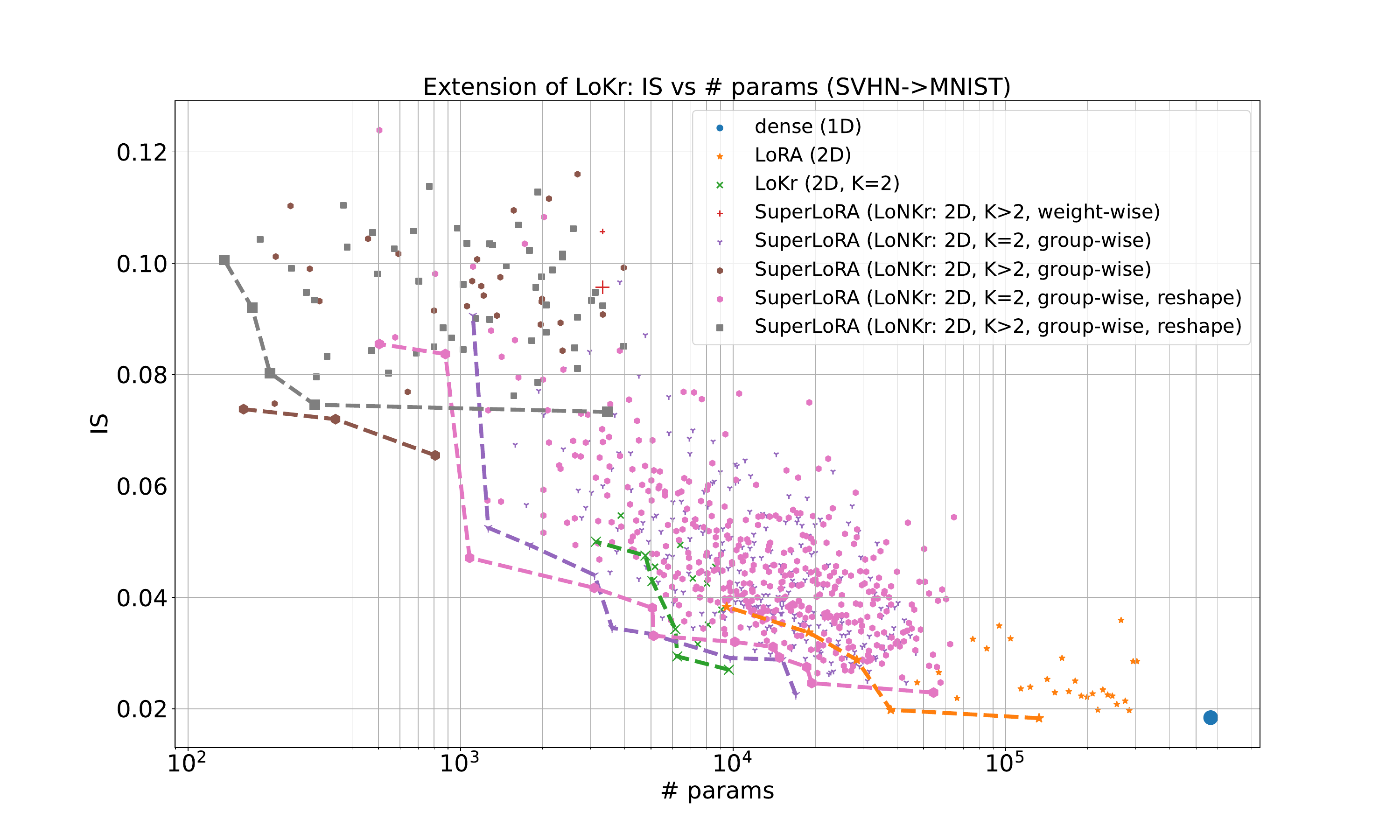}
        \caption{SuperLoRA (LoNKr, IS)}
    \end{subfigure}
   \begin{subfigure}{0.49\linewidth}
        \includegraphics[width=\linewidth]{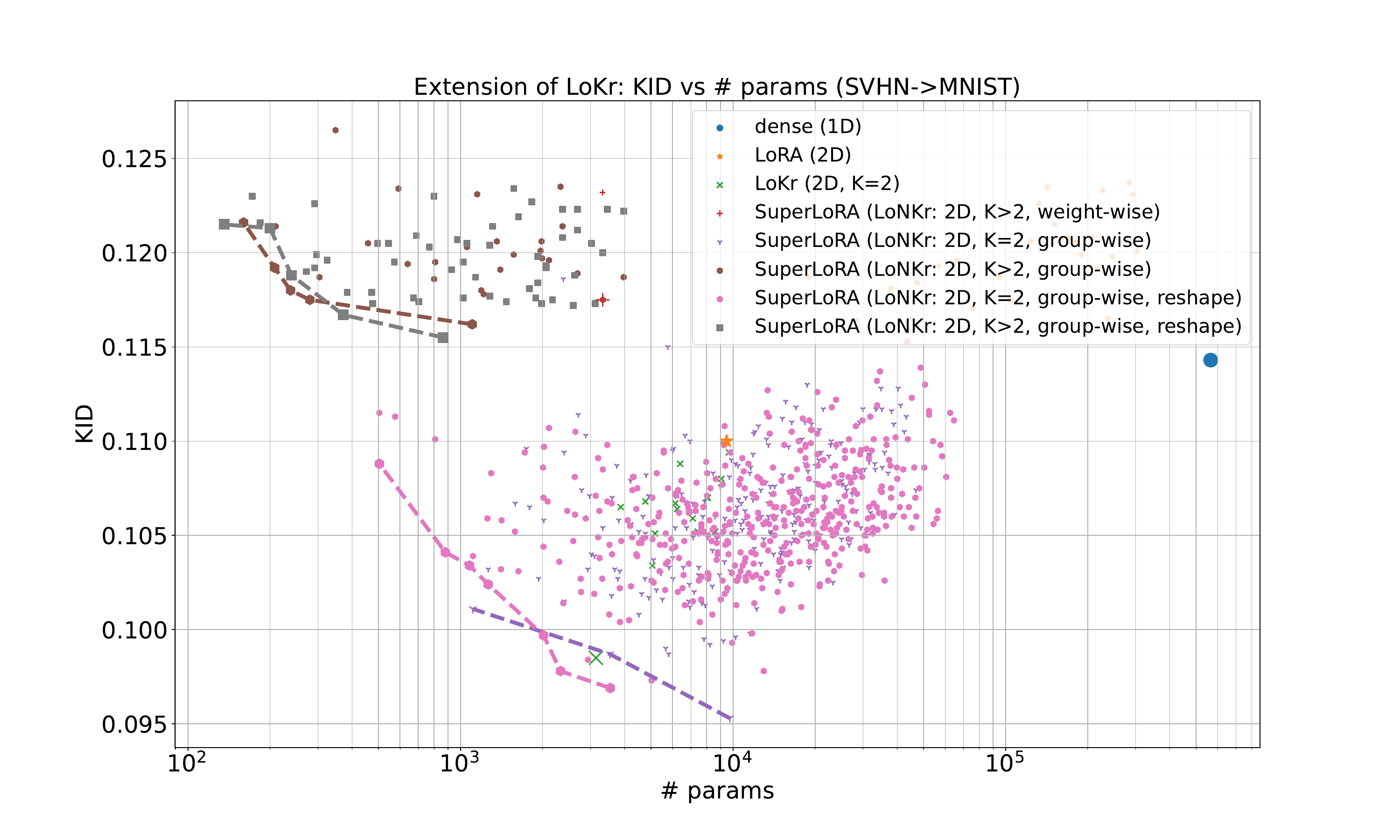}
        \caption{SuperLoRA (LoNKr, KID)}
    \end{subfigure}
    \begin{subfigure}{0.49\linewidth}
        \includegraphics[width=\linewidth]{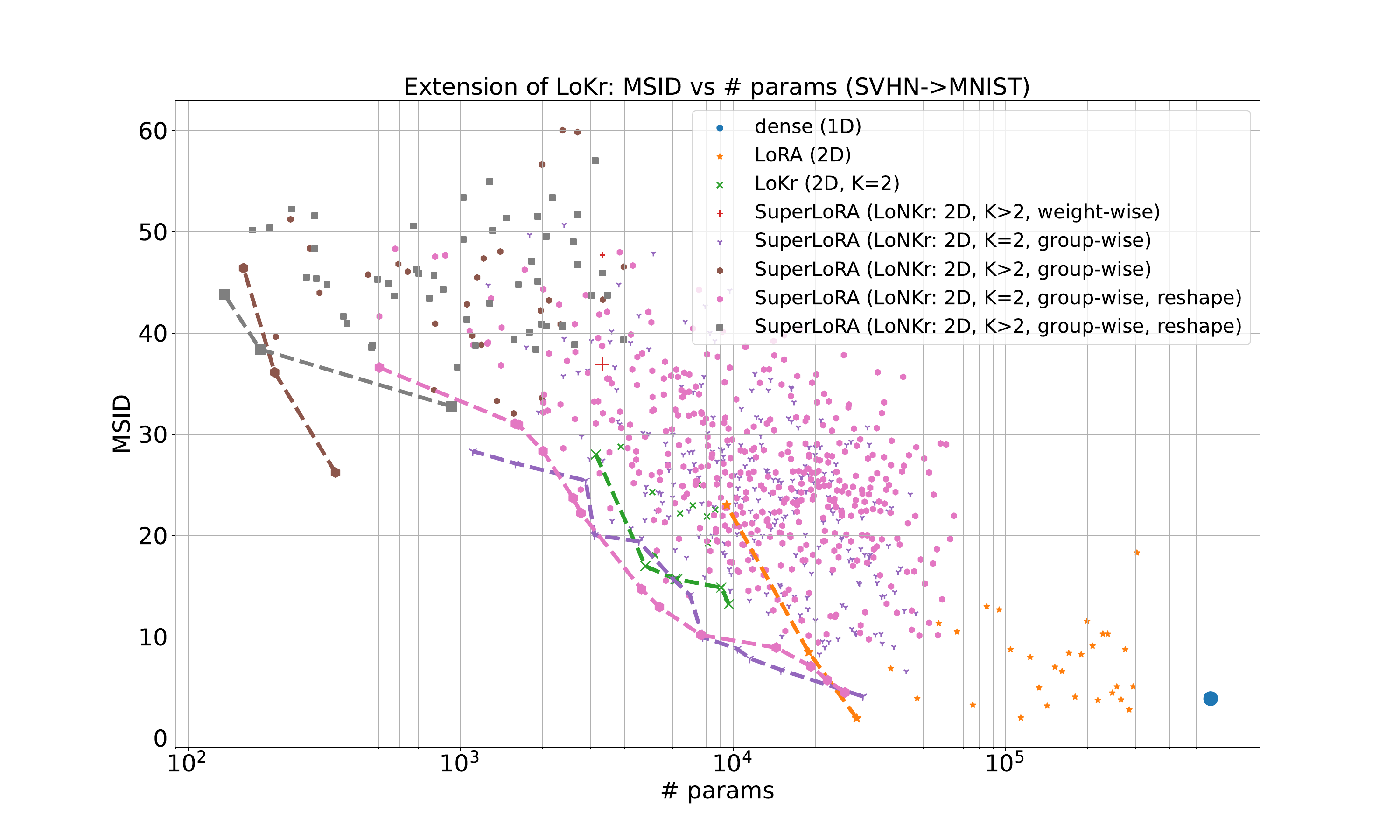}
        \caption{SuperLoRA (LoNKr, MSID)}
    \end{subfigure}    
       \begin{subfigure}{0.49\linewidth}
        \includegraphics[width=\linewidth]{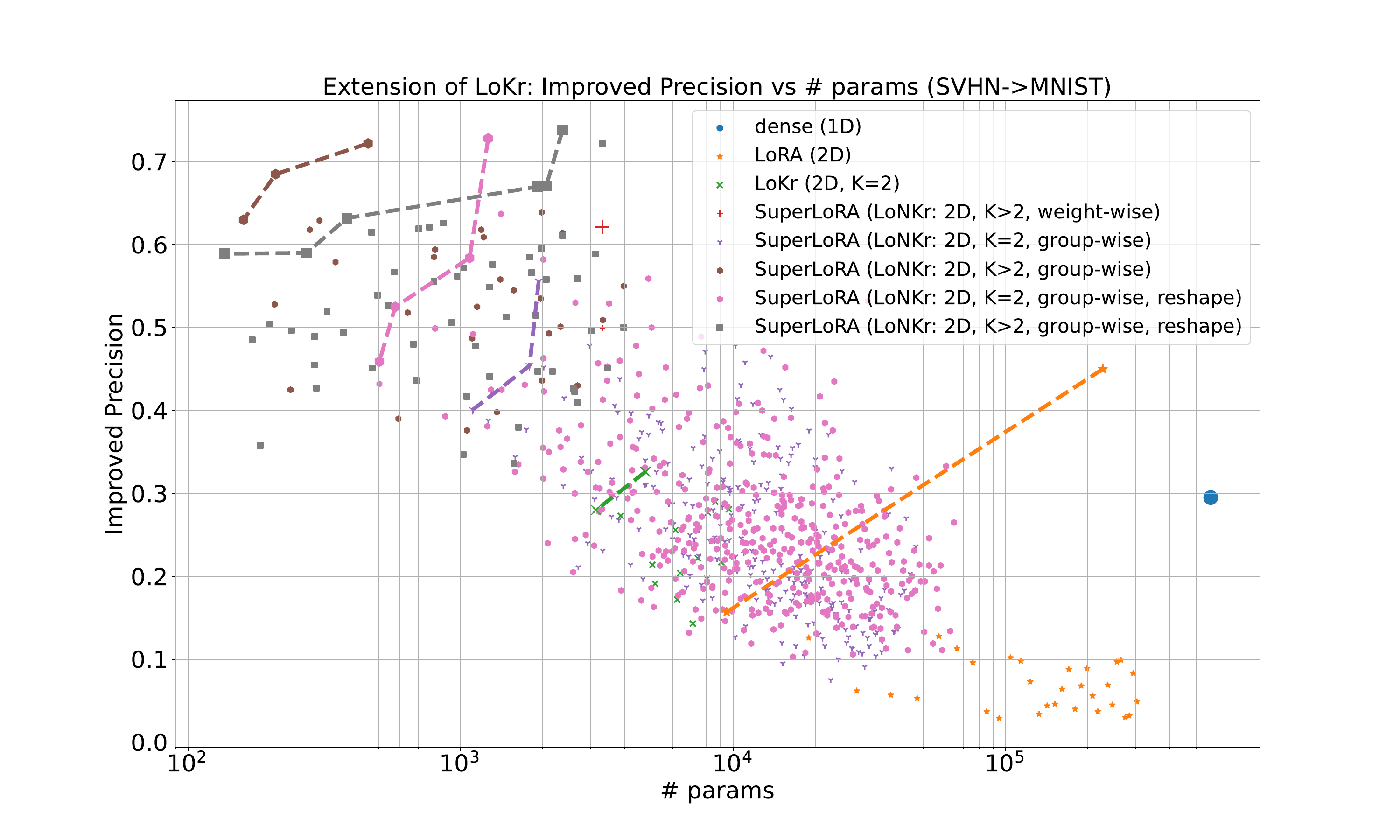}
        \caption{SuperLoRA (LoNKr, Improved Precision)}
    \end{subfigure}
    \begin{subfigure}{0.49\linewidth}
        \includegraphics[width=\linewidth]{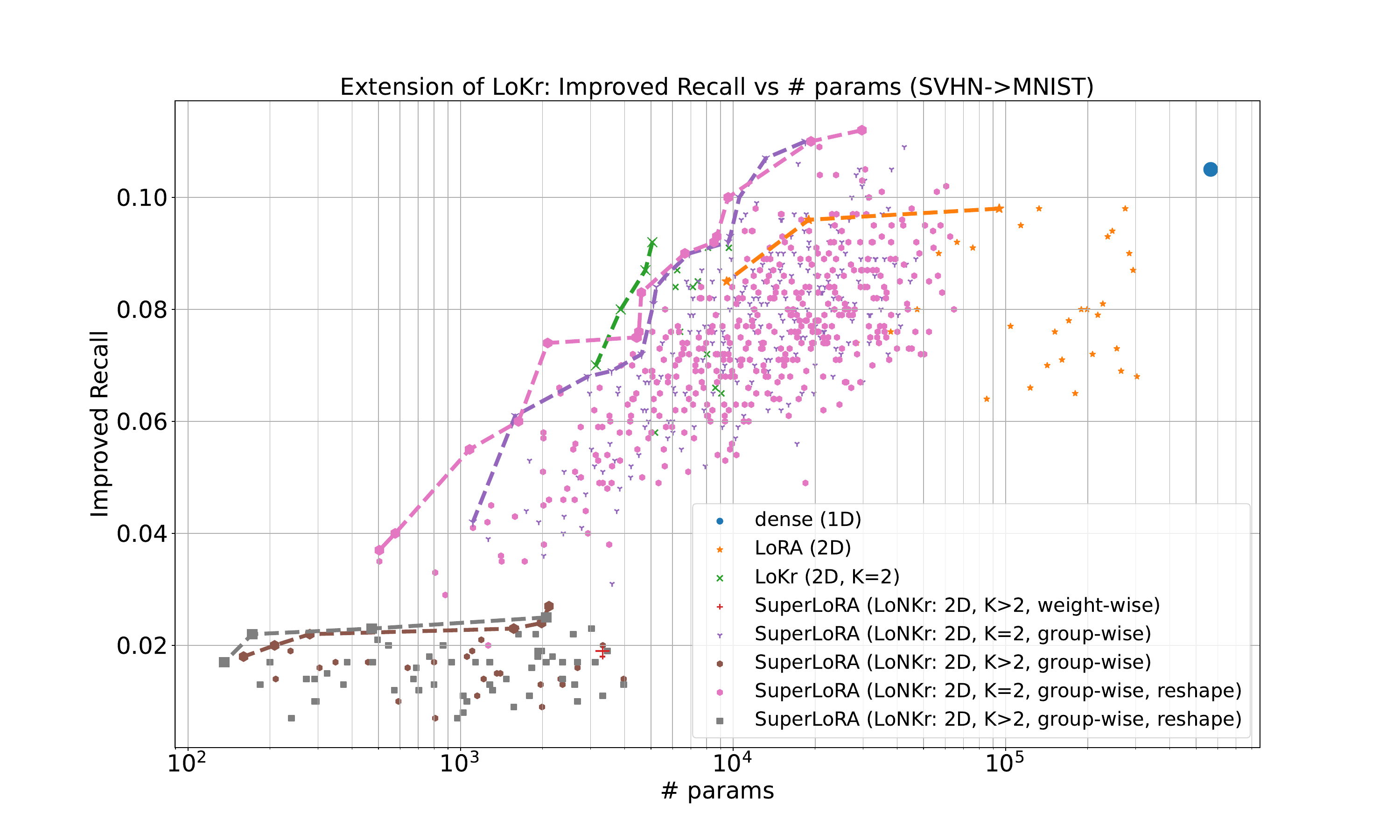}
        \caption{SuperLoRA (LoNKr, Improved Recall)}
    \end{subfigure}    
    \caption{Complete results for LoNKr.}
    \label{fig:2mnist_lonkr}
\end{figure}

Complete results for SuperLoRA (LoNKr), with scatter plots for all metrics, are shown in \Cref{fig:2mnist_lonkr}.
\subsubsection{SuperLoRA (LoRTA, complete results)}
\begin{figure}[t]
    \centering
    \begin{subfigure}{0.49\linewidth}
        \includegraphics[width=\linewidth]{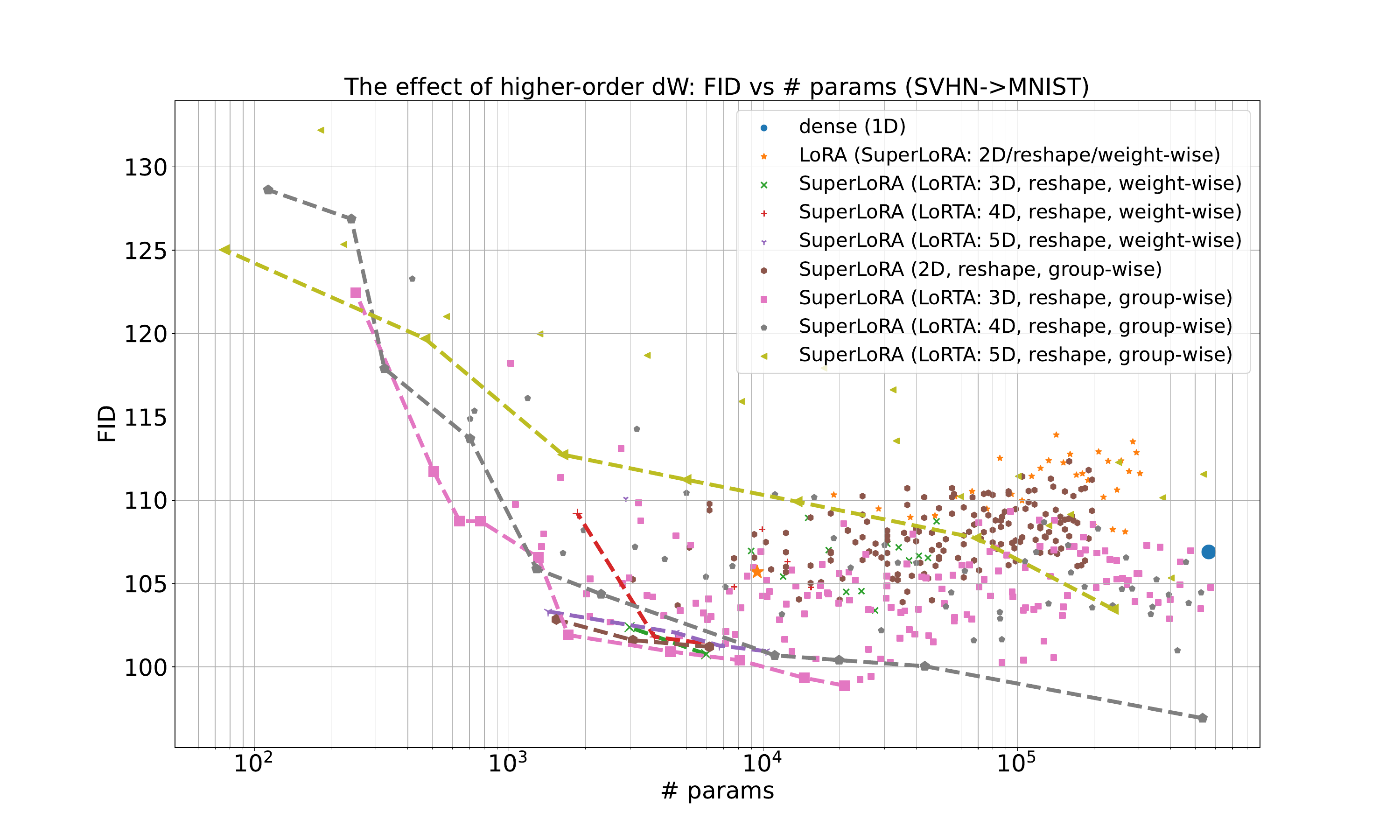}
        \caption{SuperLoRA (LoRTA, FID)}
    \end{subfigure}
    \begin{subfigure}{0.49\linewidth}
        \includegraphics[width=\linewidth]{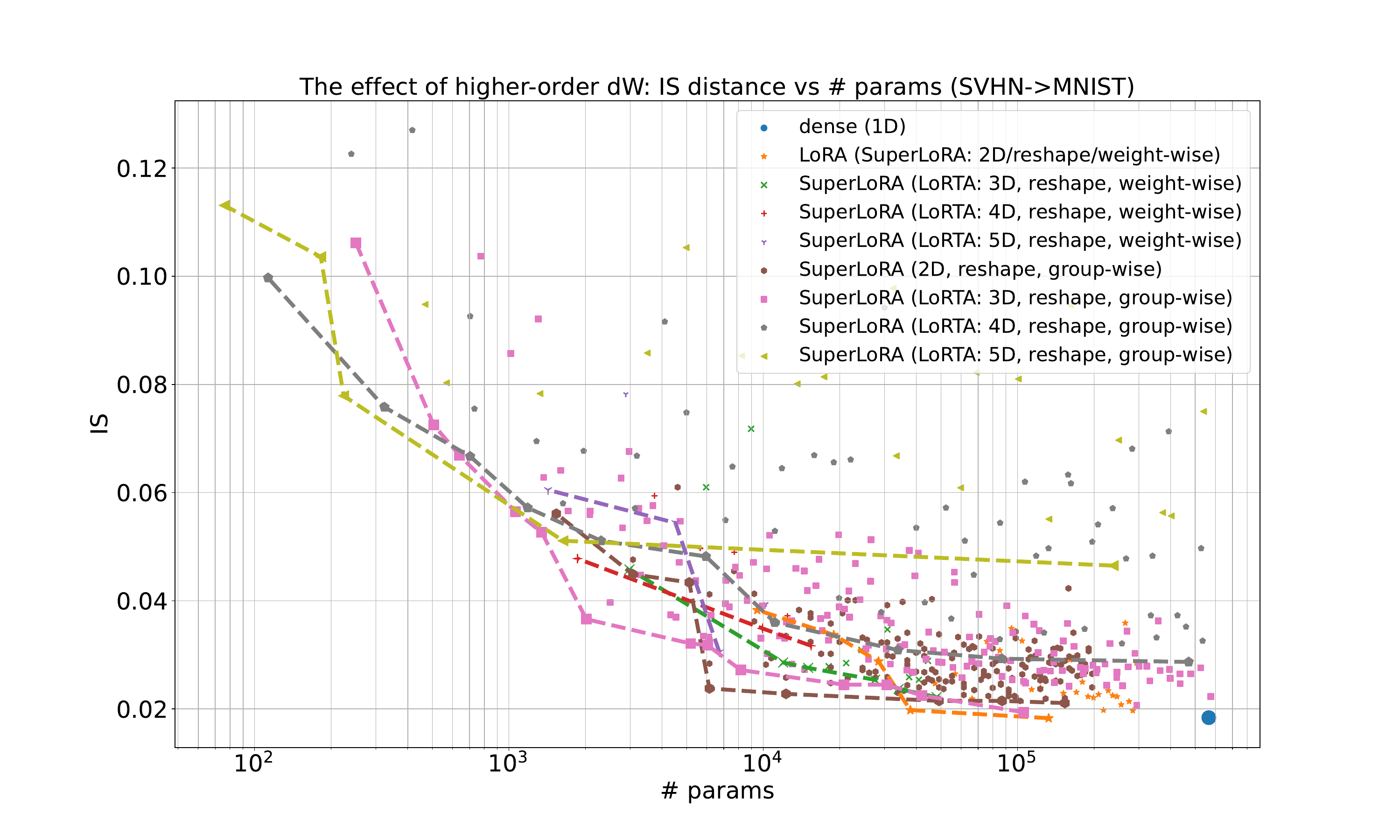}
        \caption{SuperLoRA (LoRTA, IS)}
    \end{subfigure}
   \begin{subfigure}{0.49\linewidth}
        \includegraphics[width=\linewidth]{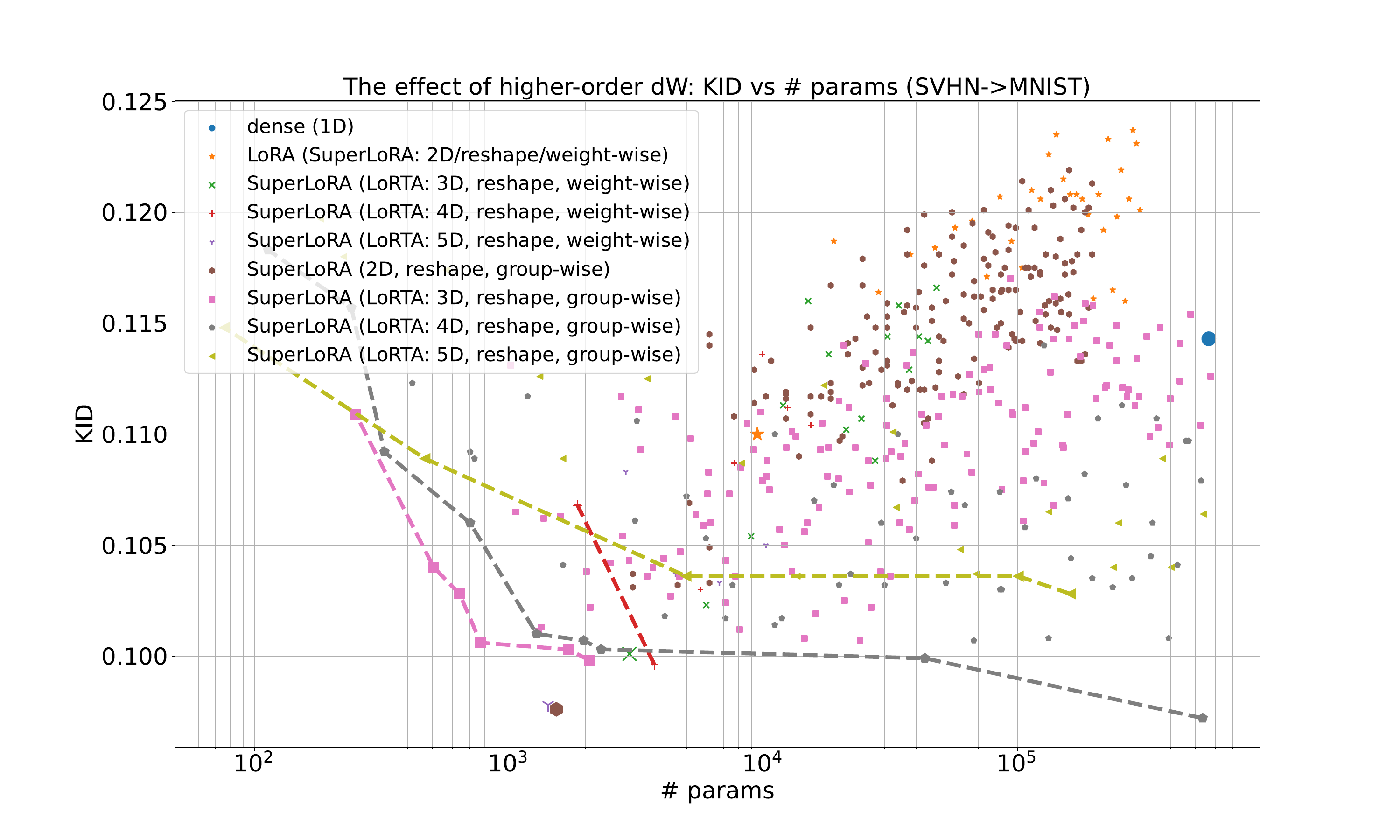}
        \caption{SuperLoRA (LoRTA, KID)}
    \end{subfigure}
    \begin{subfigure}{0.49\linewidth}
        \includegraphics[width=\linewidth]{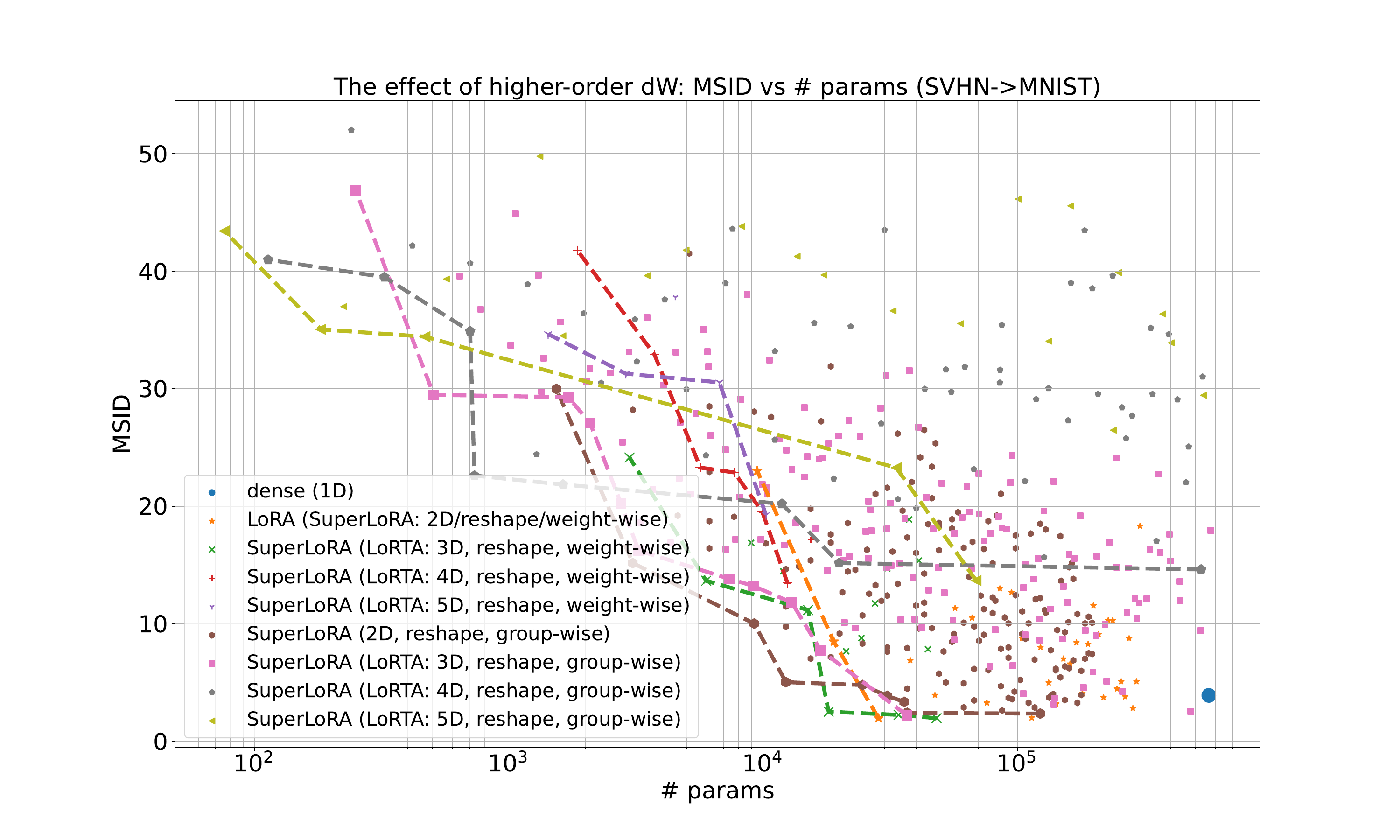}
        \caption{SuperLoRA (LoRTA, MSID)}
    \end{subfigure}    
       \begin{subfigure}{0.49\linewidth}
        \includegraphics[width=\linewidth]{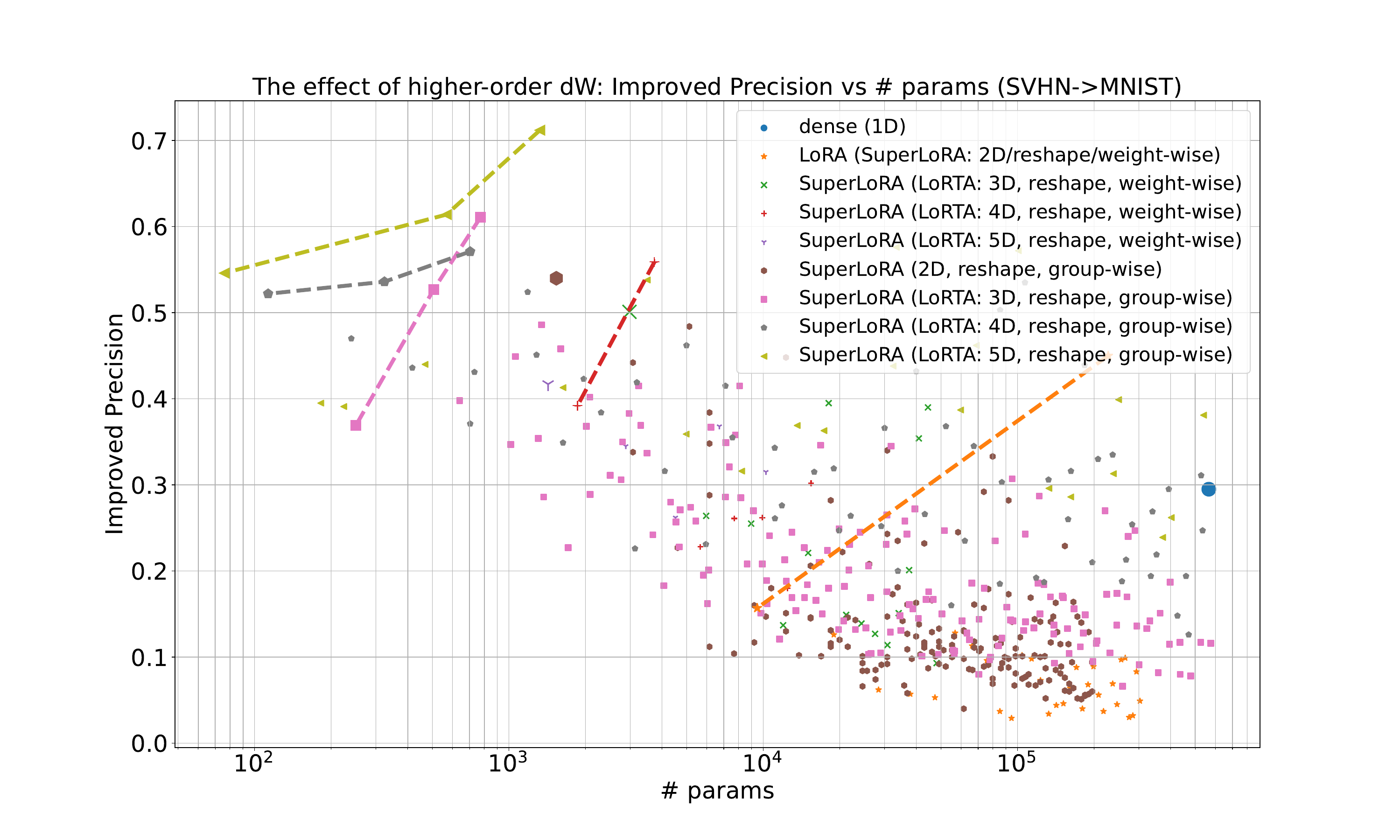}
        \caption{SuperLoRA (LoRTA, Improved Precision)}
    \end{subfigure}
    \begin{subfigure}{0.49\linewidth}
        \includegraphics[width=\linewidth]{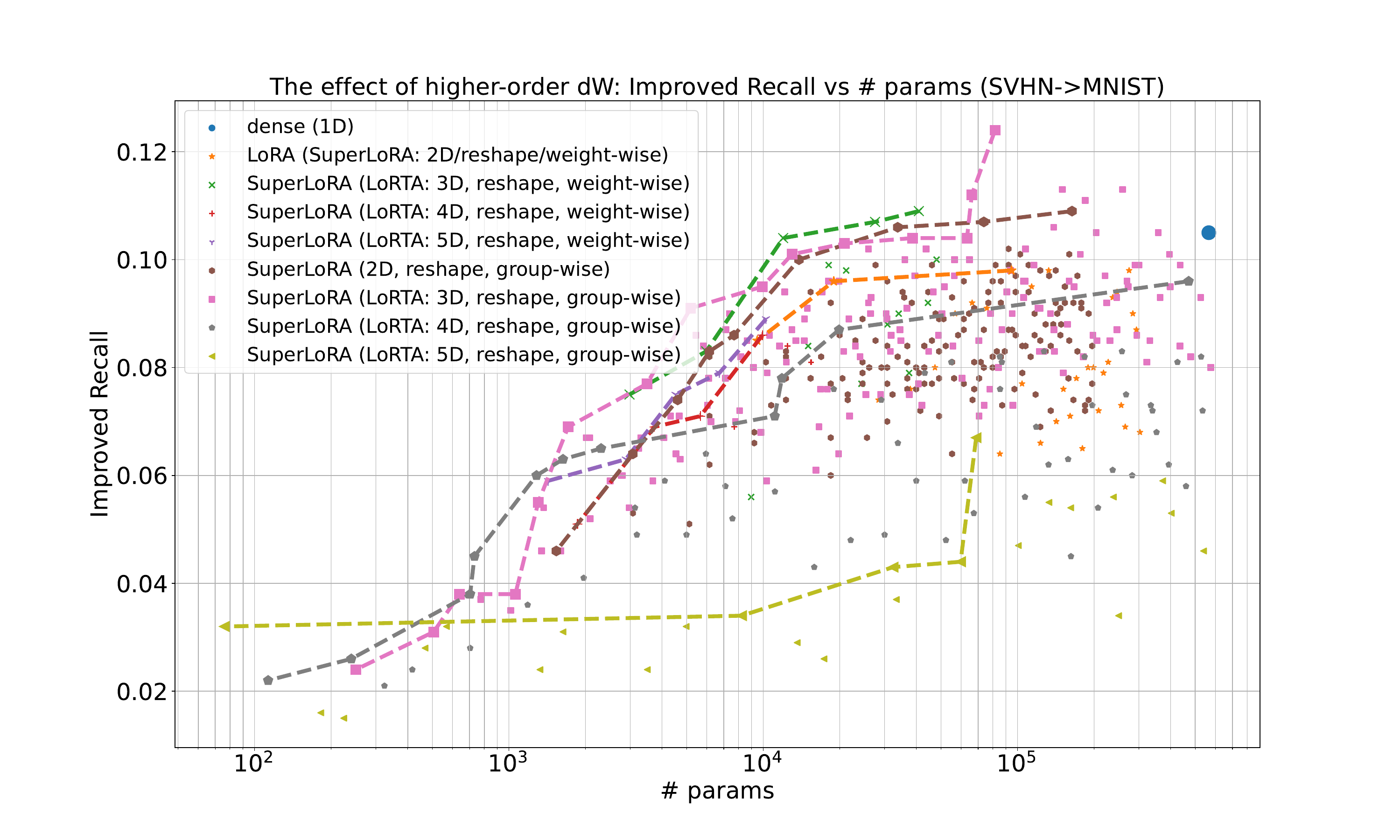}
        \caption{SuperLoRA (LoRTA, Improved Recall)}
    \end{subfigure}    
    \caption{Complete results for LoRTA.}
    \label{fig:2mnist_lorta}
\end{figure}

Complete results for SuperLoRA (LoRTA), with scatter plots for all metrics, are shown in \Cref{fig:2mnist_lorta}.

\subsection{Transfer learning from MNIST to SVHN}
\label{appendix:2svhn}

\subsubsection{Grouping effect}
\begin{figure}[t]
    \centering
    \begin{subfigure}{0.49\linewidth}
        \includegraphics[width=\linewidth]{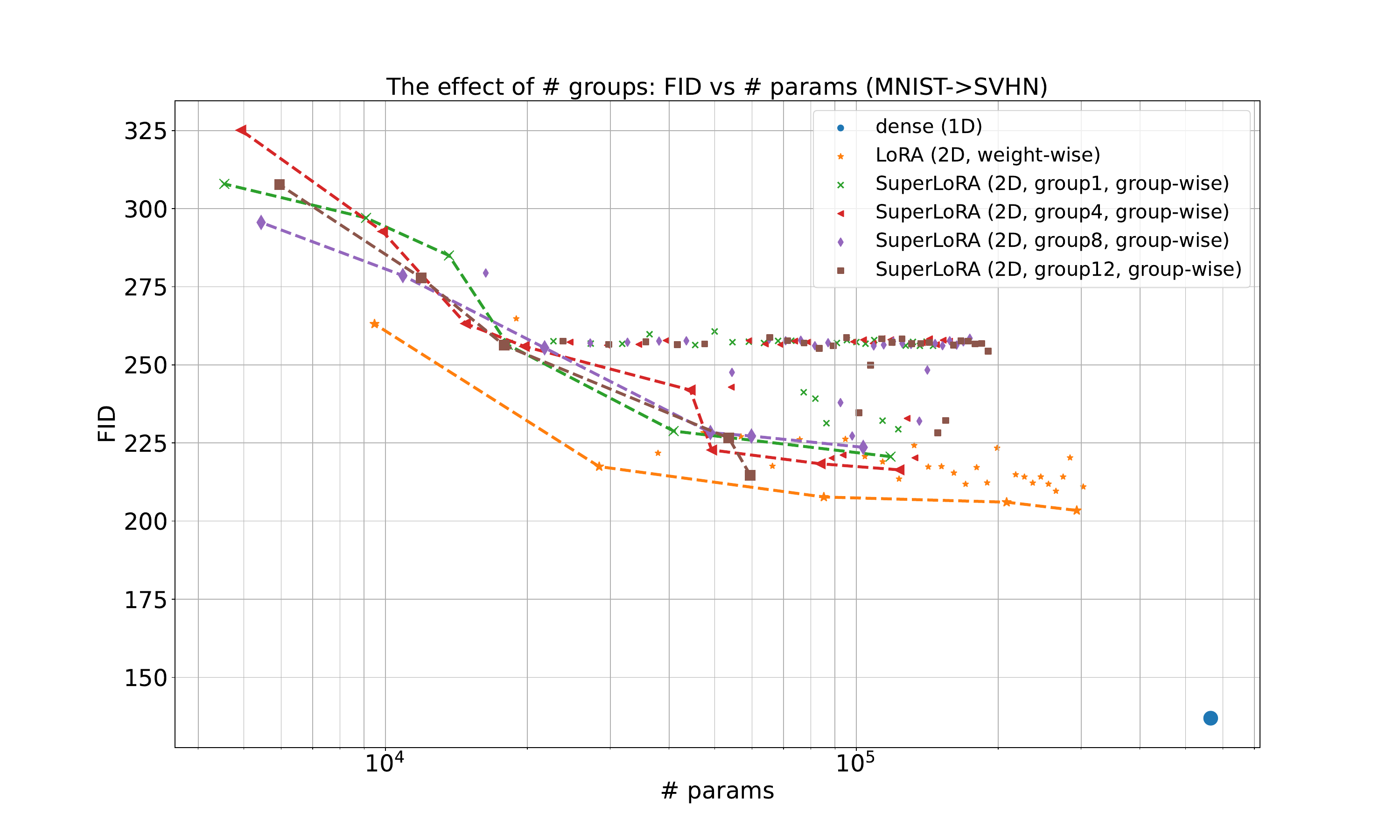}
        \caption{weight-wise \vs group-wise (FID)}
    \end{subfigure}
    \begin{subfigure}{0.49\linewidth}
        \includegraphics[width=\linewidth]{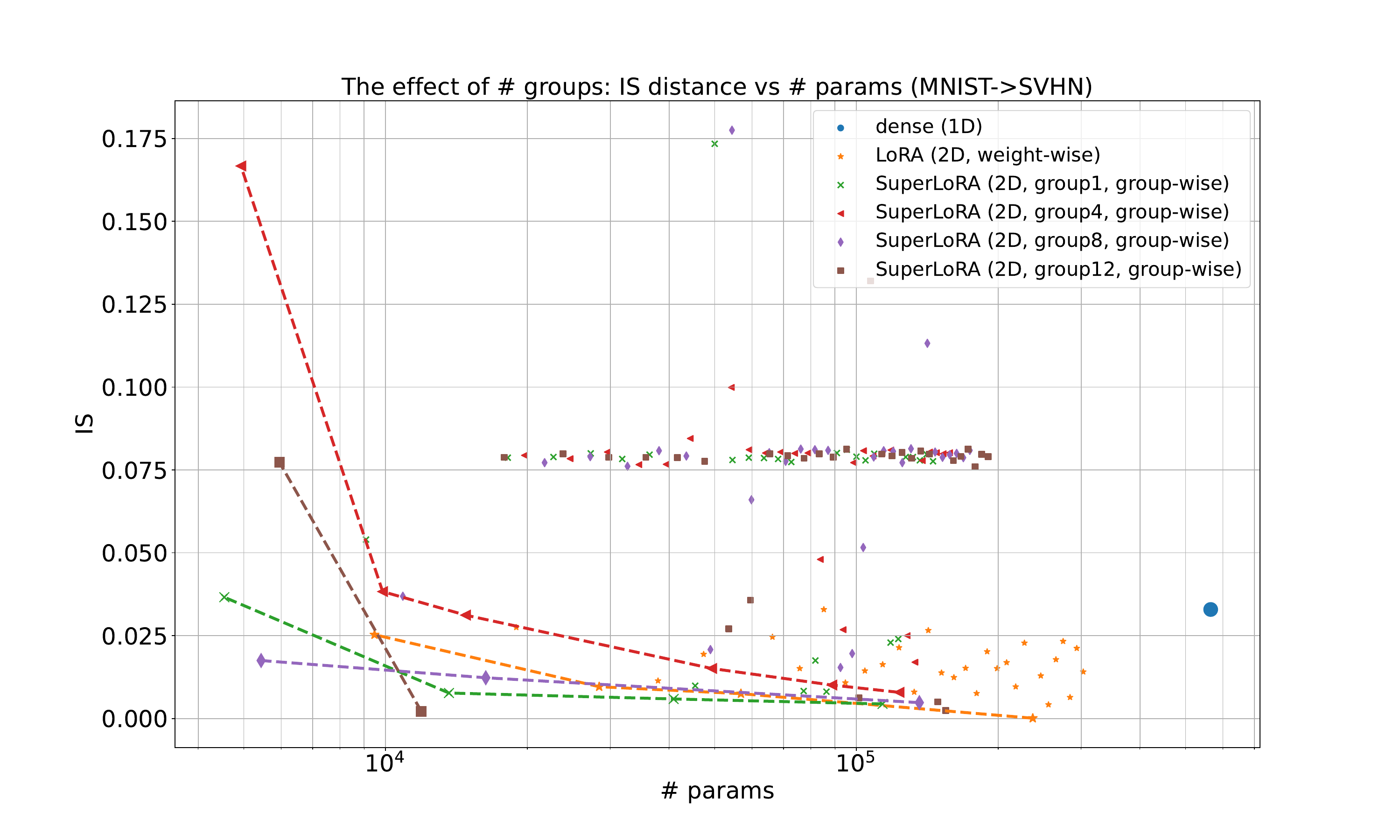}
        \caption{weight-wise \vs group-wise (IS)}
    \end{subfigure}
   \begin{subfigure}{0.49\linewidth}
        \includegraphics[width=\linewidth]{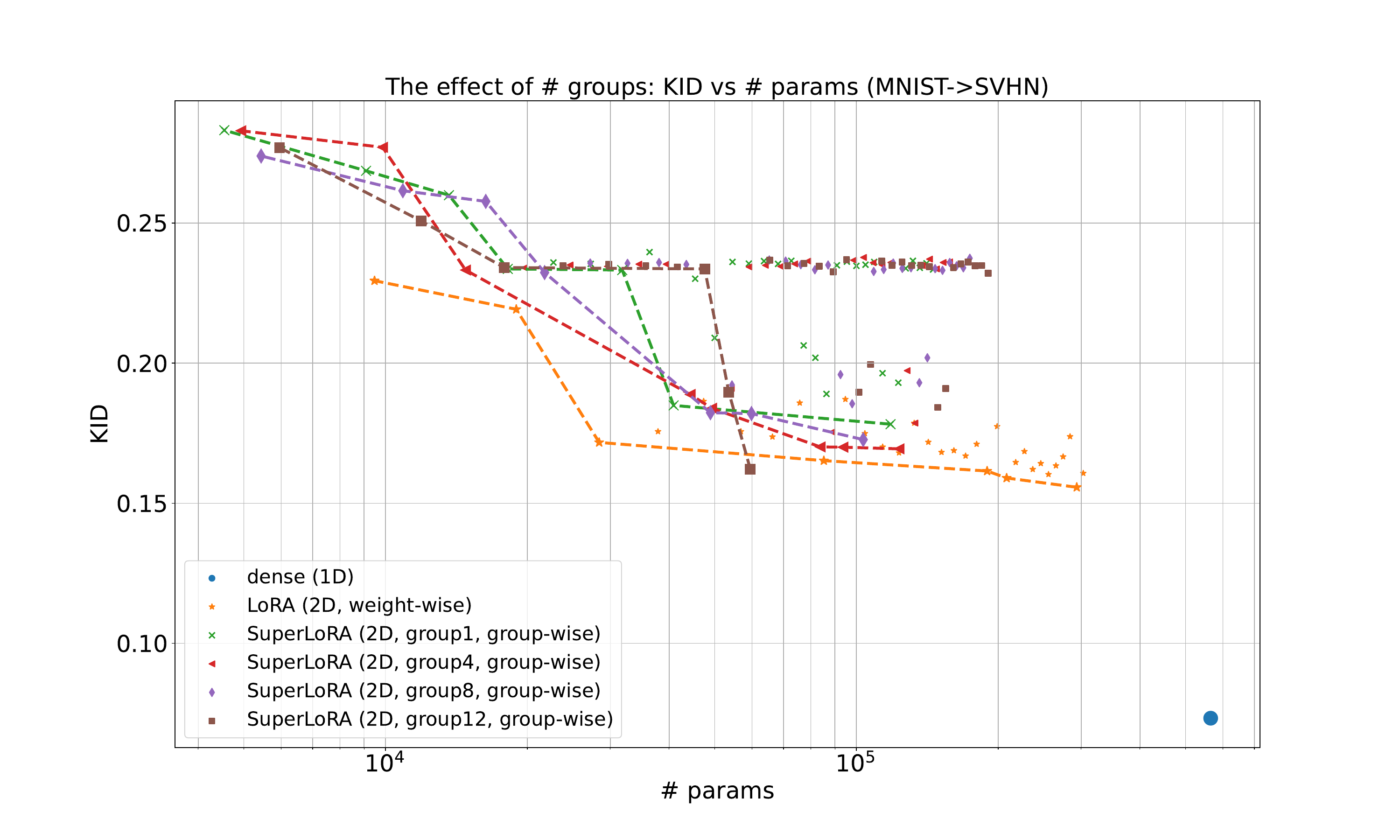}
        \caption{weight-wise \vs group-wise (KID)}
    \end{subfigure}
    \begin{subfigure}{0.49\linewidth}
        \includegraphics[width=\linewidth]{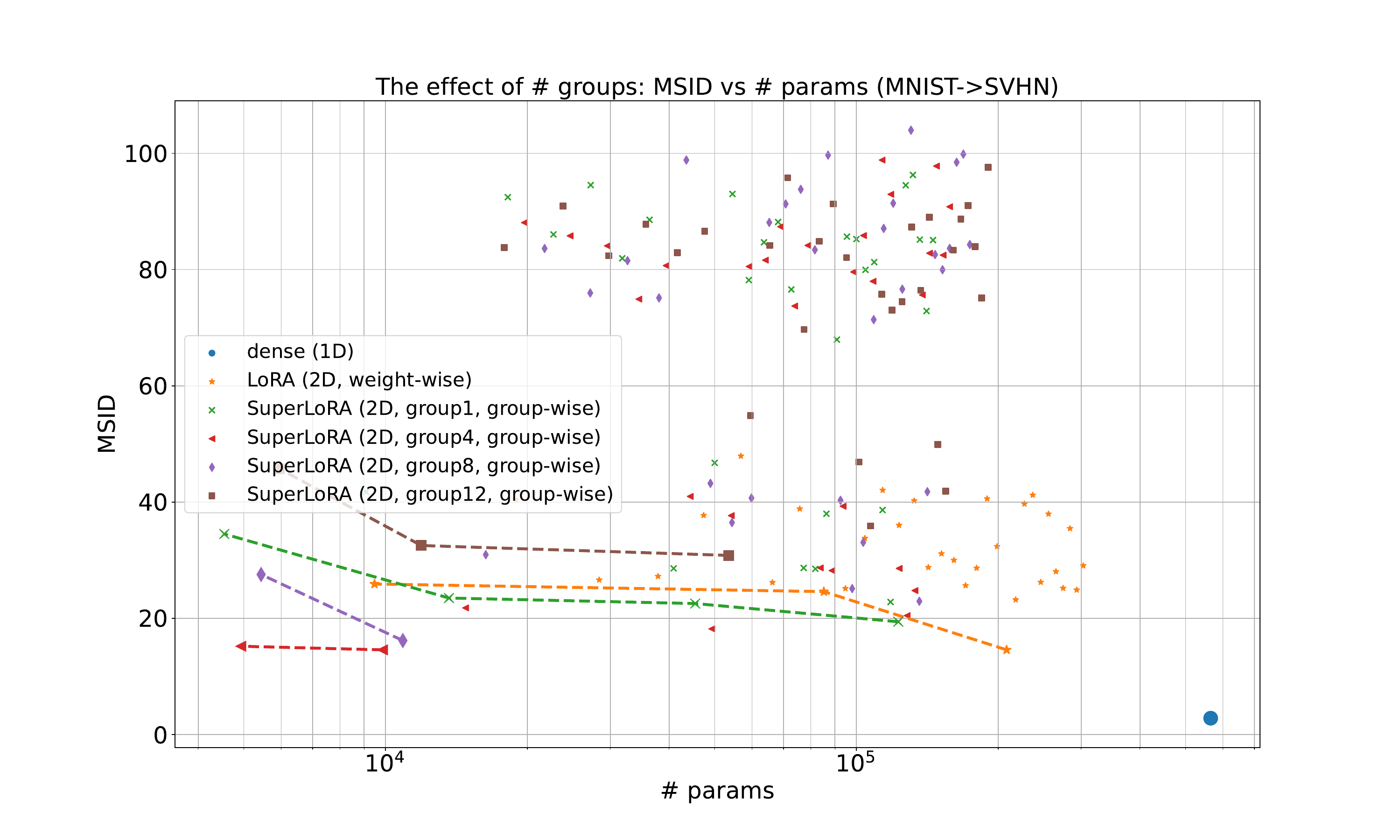}
        \caption{weight-wise \vs group-wise (MSID)}
    \end{subfigure}    
       \begin{subfigure}{0.49\linewidth}
        \includegraphics[width=\linewidth]{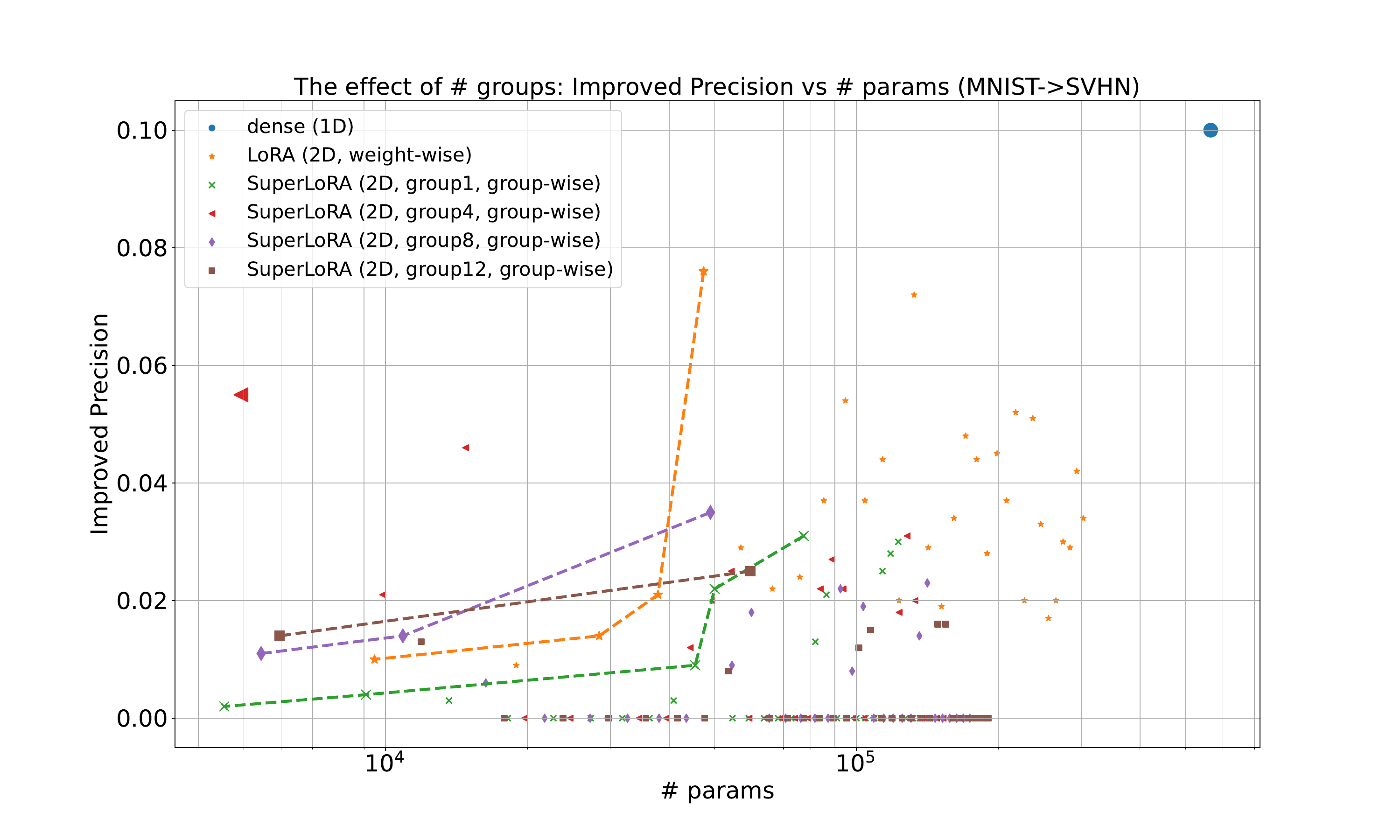}
        \caption{weight-wise \vs group-wise (Improved Precision)}
    \end{subfigure}
    \begin{subfigure}{0.49\linewidth}
        \includegraphics[width=\linewidth]{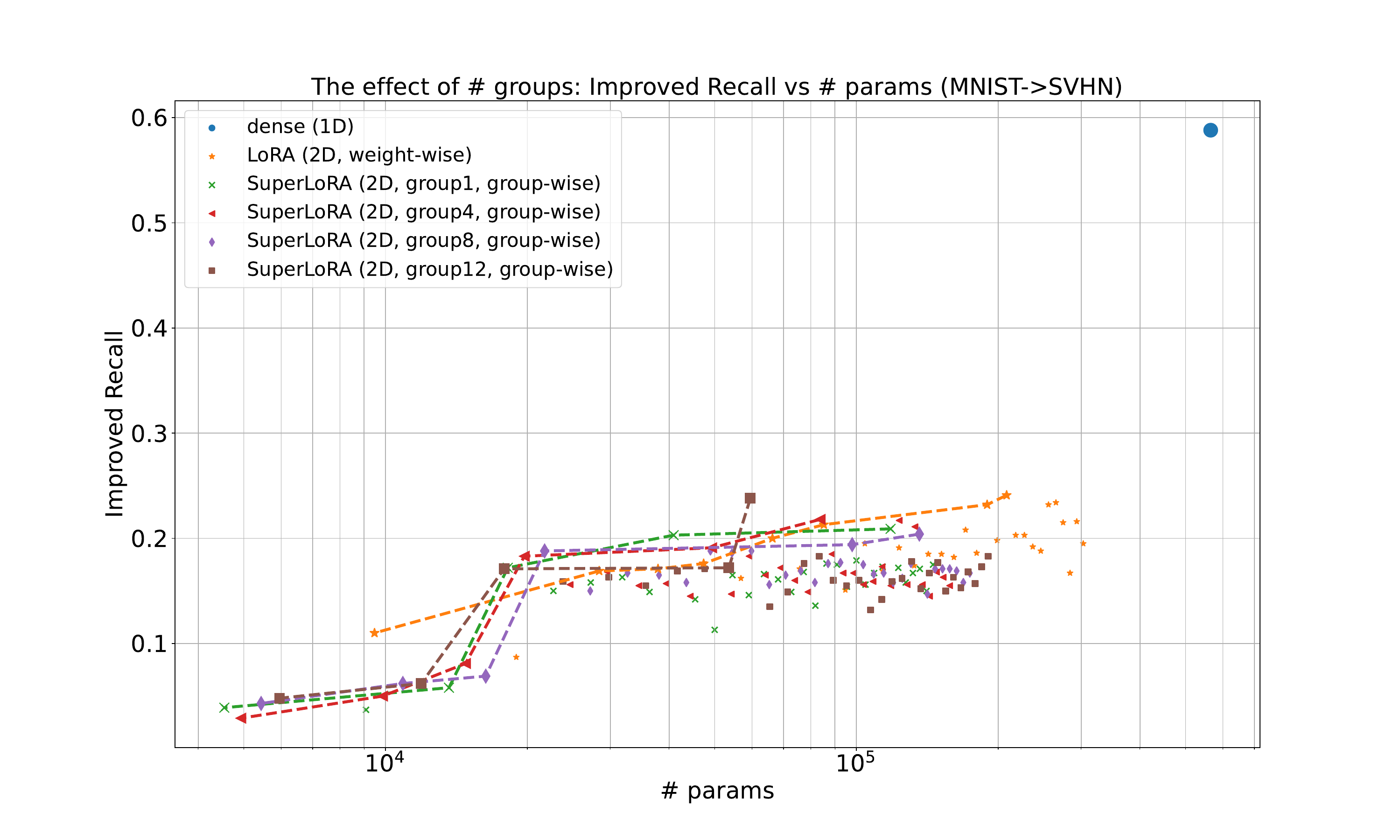}
        \caption{weight-wise \vs group-wise (Improved Recall)}
    \end{subfigure}    
    \caption{Complete comparison between weight-wise LoRA and group-wise SuperLoRA for transfer learning from MNIST to SVHN.}
    \label{fig:2svhn_group}
\end{figure}

Transfer learning from MNIST to SVHN is also tested. \Cref{fig:2svhn_group} shows that some metrics cannot function when transferred from a simpler dataset to a more complicated one, \eg FID, IS, KID and Improved Precision, where some ill-posed cases appear. Besides this, we can still find from the Pareto frontiers that SuperLoRA extends LoRA to low-parameter regime and works better occasionally in terms of IS, MSID, Improved Precision and Improved Recall.

\subsubsection{Reshaping effect}
\begin{figure}[t]
    \centering
    \begin{subfigure}{0.49\linewidth}
        \includegraphics[width=\linewidth]{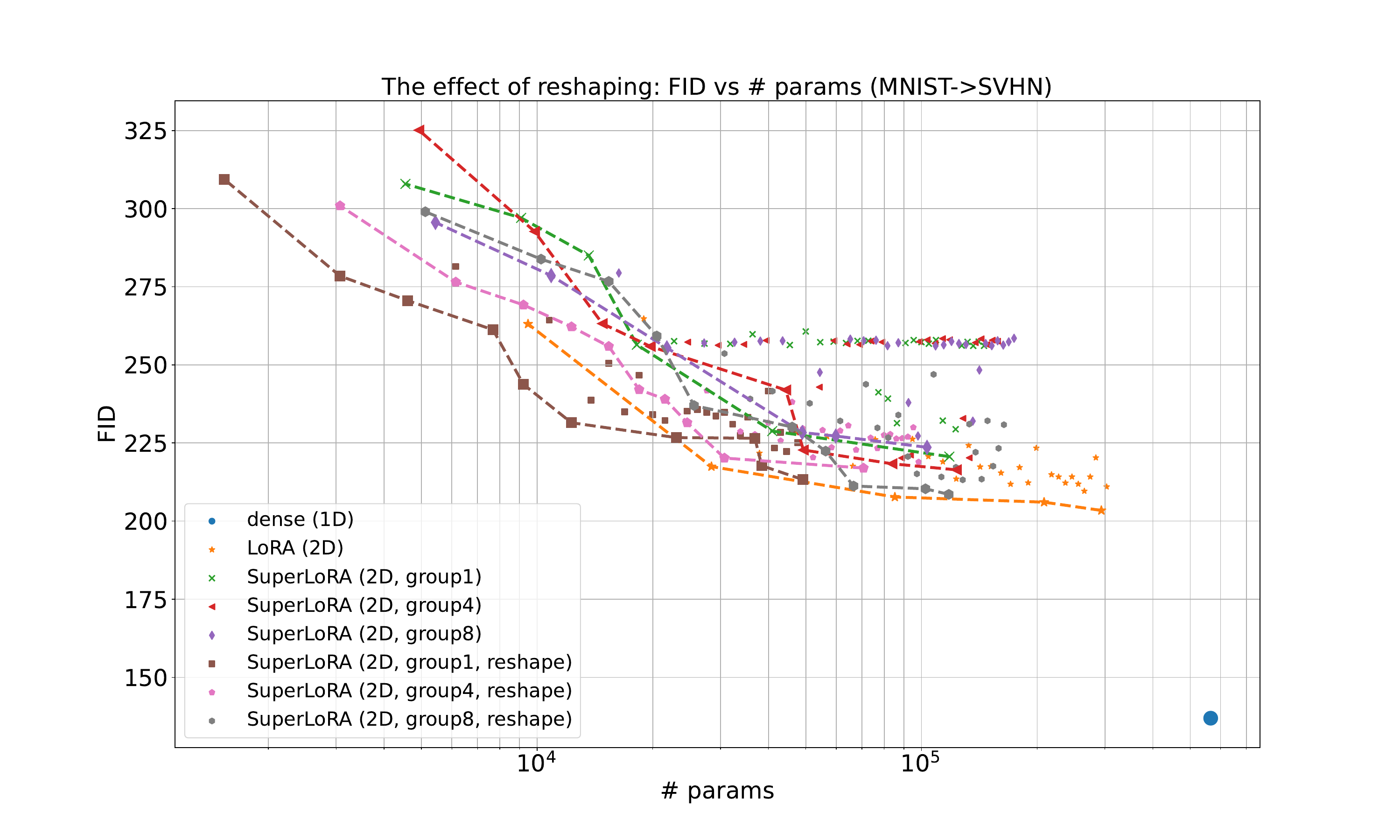}
        \caption{reshaping \vs non-reshaping (FID)}
    \end{subfigure}
    \begin{subfigure}{0.49\linewidth}
        \includegraphics[width=\linewidth]{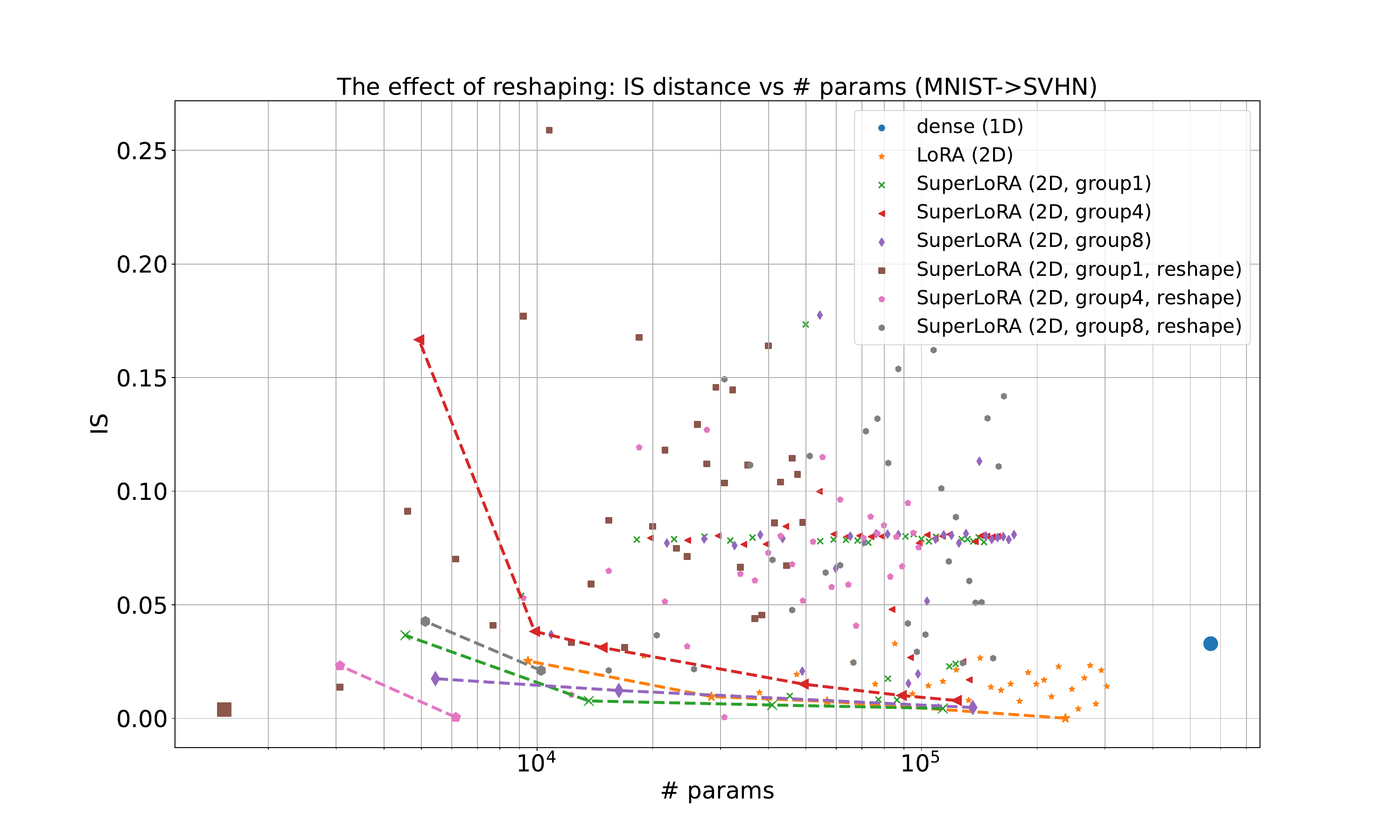}
        \caption{reshaping \vs non-reshaping (IS)}
    \end{subfigure}
   \begin{subfigure}{0.49\linewidth}
        \includegraphics[width=\linewidth]{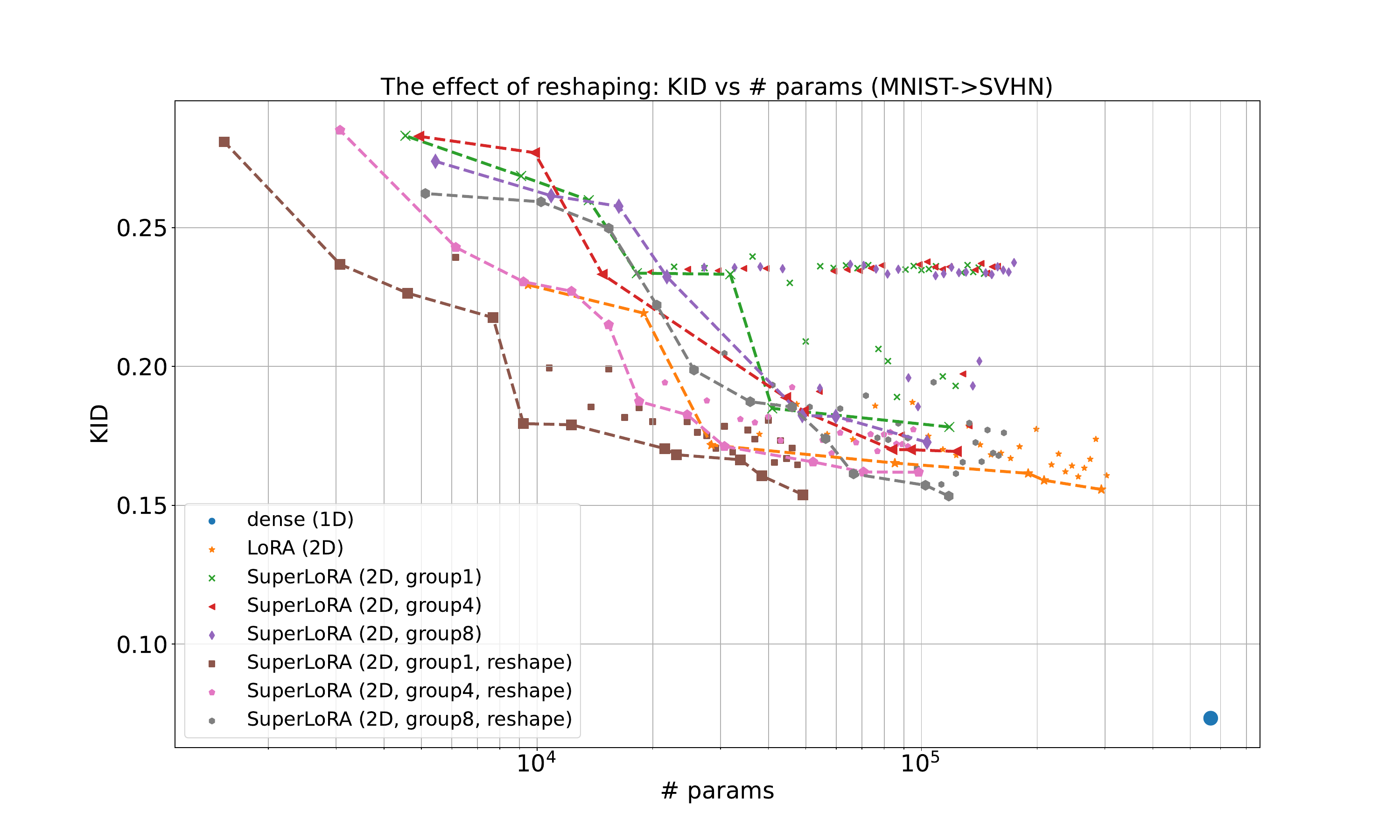}
        \caption{reshaping \vs non-reshaping (KID)}
    \end{subfigure}
    \begin{subfigure}{0.49\linewidth}
        \includegraphics[width=\linewidth]{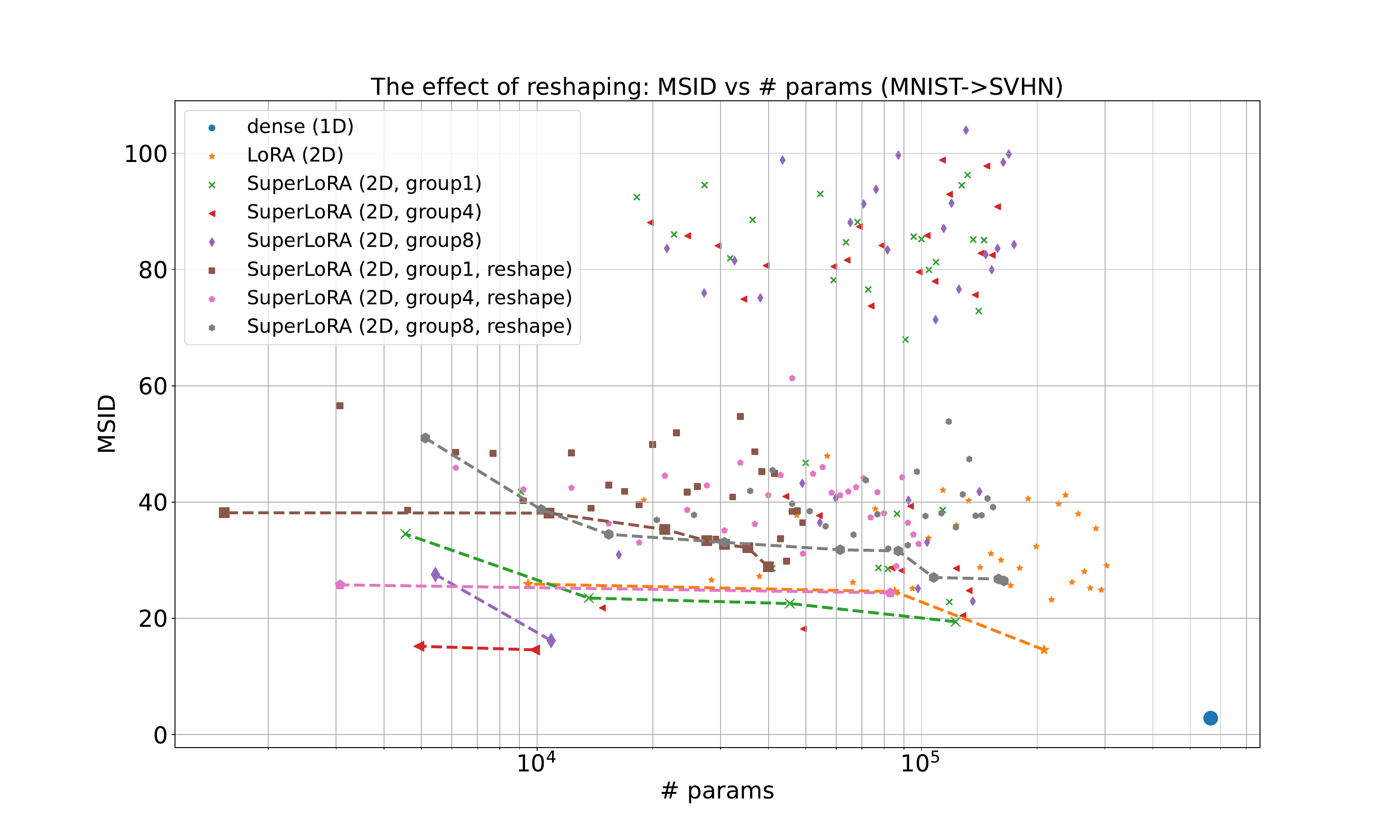}
        \caption{reshaping \vs non-reshaping (MSID)}
    \end{subfigure}    
       \begin{subfigure}{0.49\linewidth}
        \includegraphics[width=\linewidth]{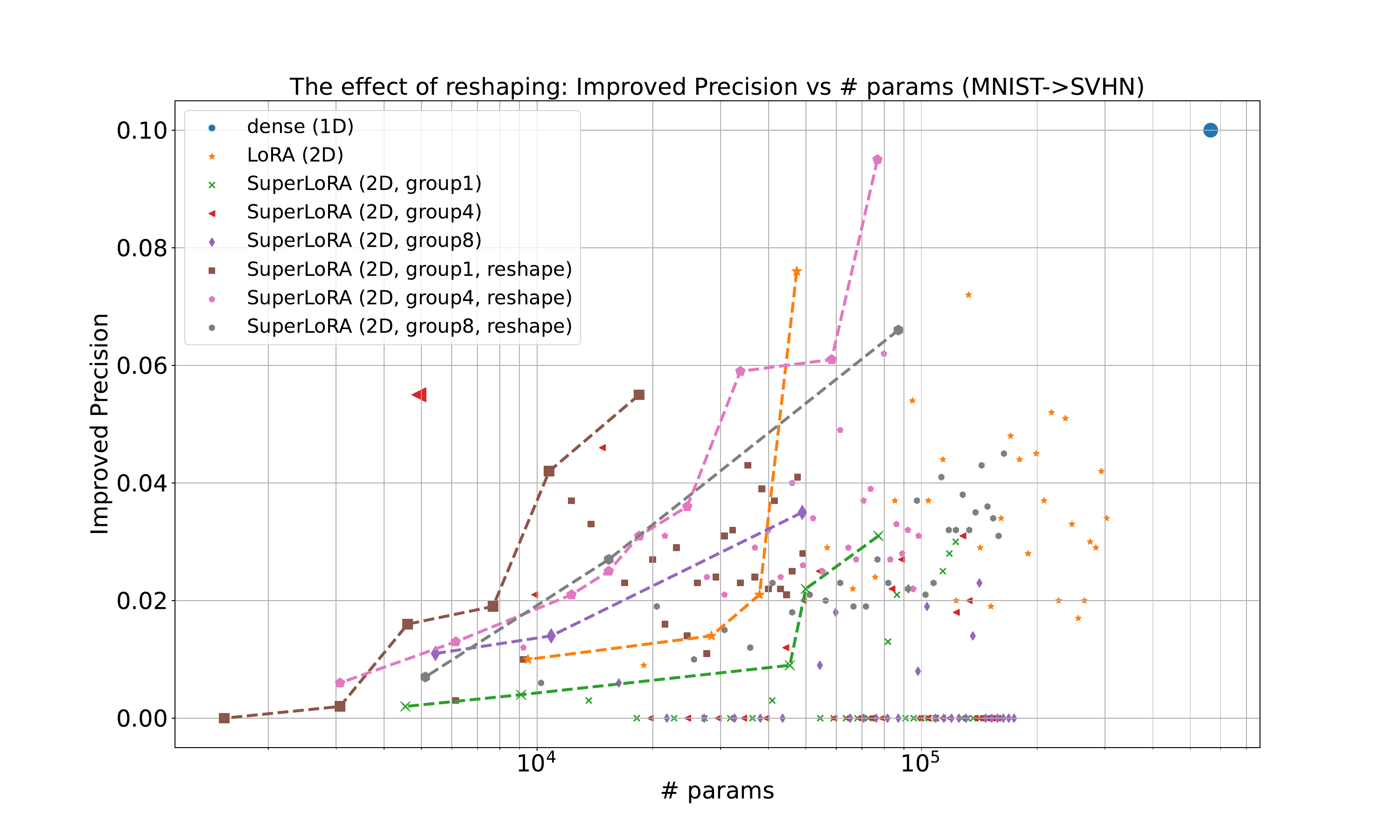}
        \caption{reshaping \vs non-reshaping (Improved Precision)}
    \end{subfigure}
    \begin{subfigure}{0.49\linewidth}
        \includegraphics[width=\linewidth]{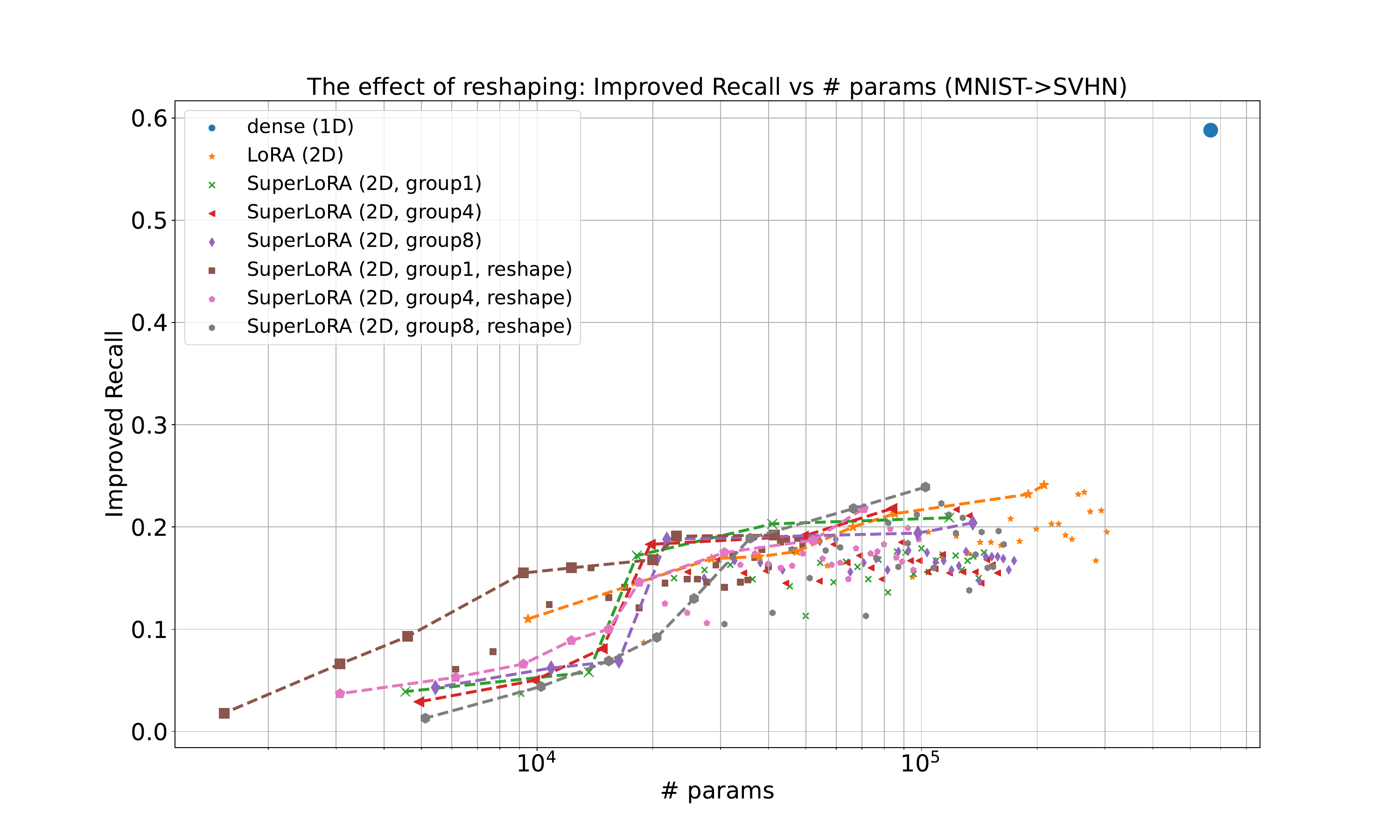}
        \caption{reshaping \vs non-reshaping (Improved Recall)}
    \end{subfigure}    
    \caption{Complete comparison between reshaping and non-reshaping SuperLoRA for transfer learning from MNIST to SVHN.}
    \label{fig:2svhn_reshape}
\end{figure}

\Cref{fig:2svhn_reshape} shows that SuperLoRA with reshaping works better than non-reshaping in most cases in transfer learning from MNIST to SVHN, consistent with the results in transfer learning from SVHN to MNIST.

\subsubsection{SuperLoRA (LoNKr)}
\begin{figure}[t]
    \centering
    \begin{subfigure}{0.49\linewidth}
        \includegraphics[width=\linewidth]{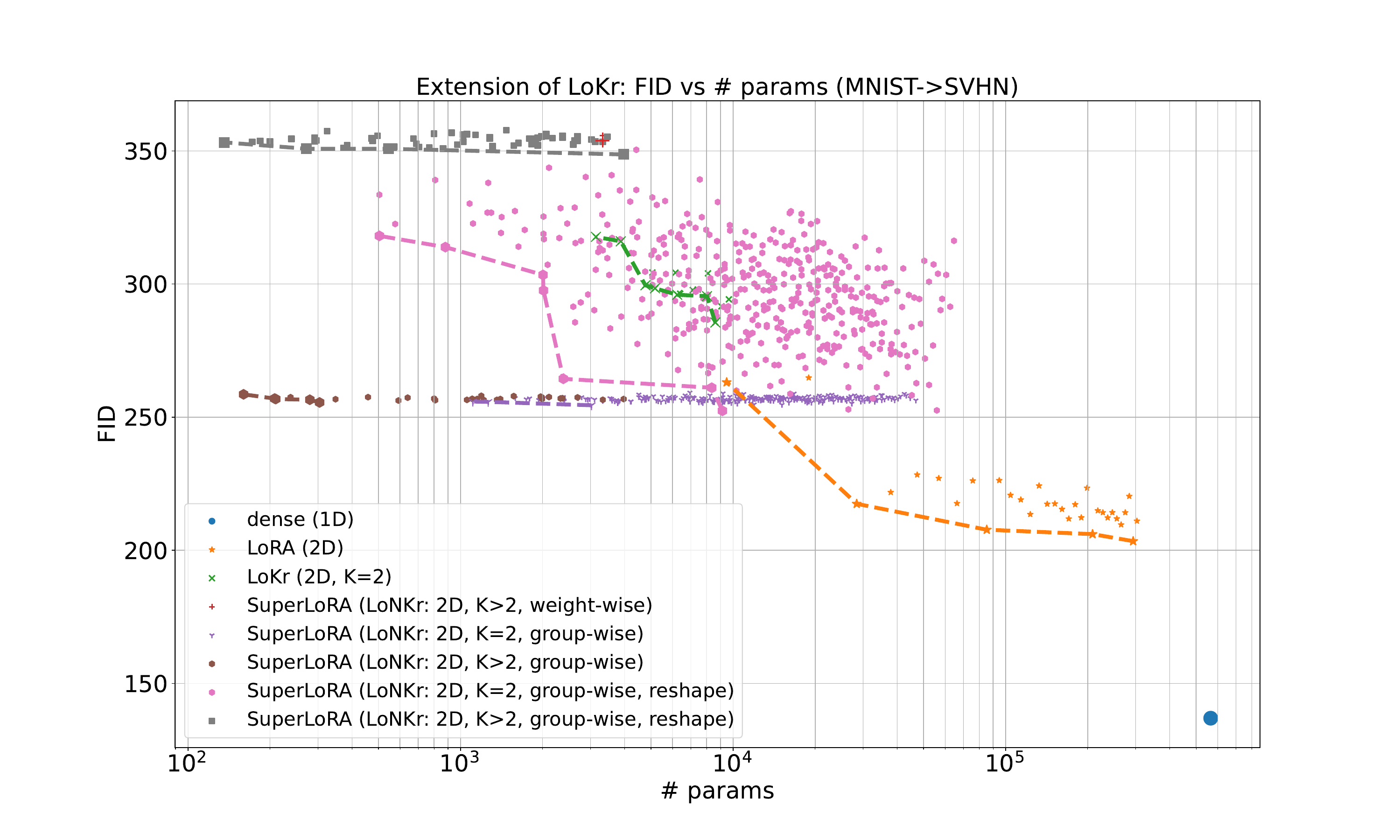}
        \caption{SuperLoRA (LoNKr, FID)}
    \end{subfigure}
    \begin{subfigure}{0.49\linewidth}
        \includegraphics[width=\linewidth]{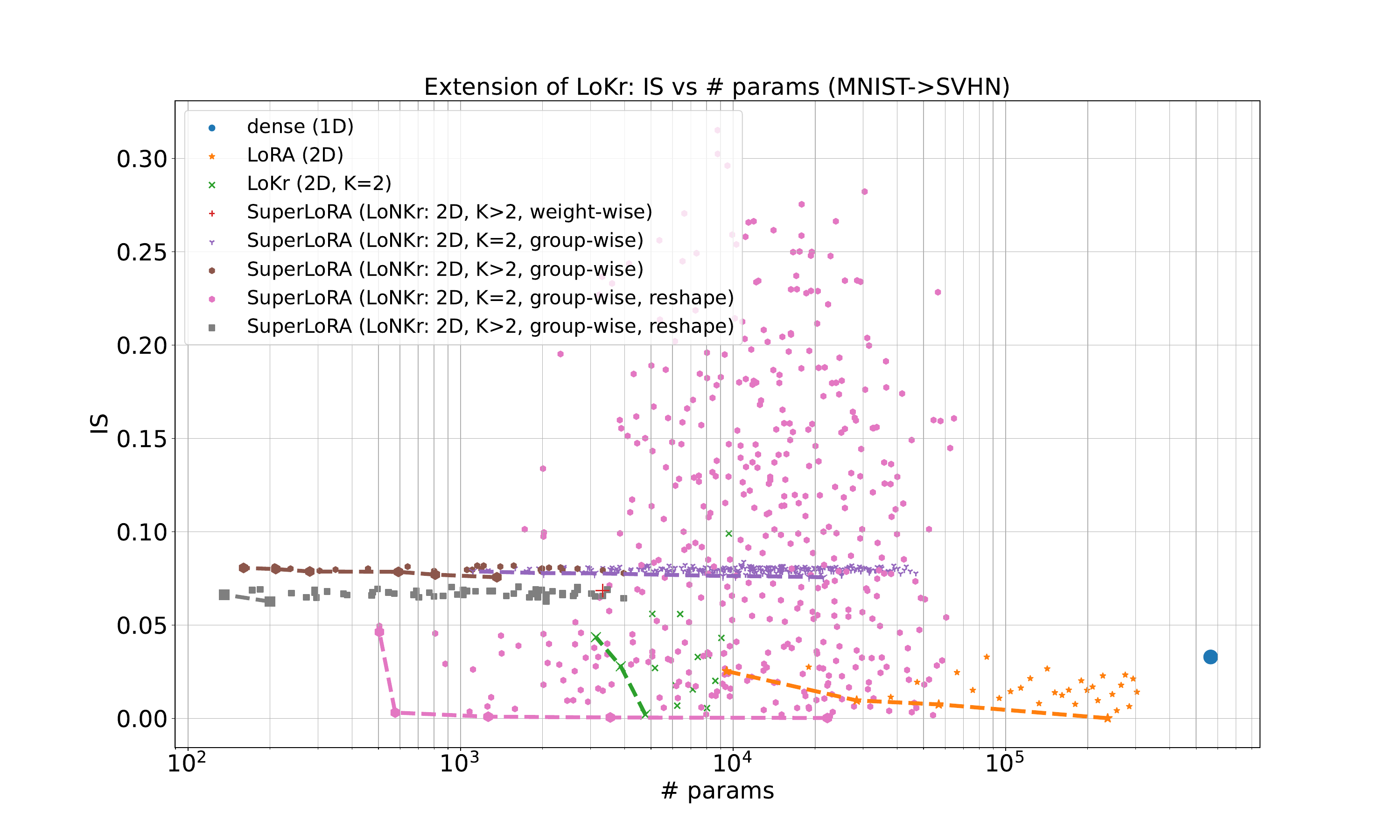}
        \caption{SuperLoRA (LoNKr, IS)}
    \end{subfigure}
   \begin{subfigure}{0.49\linewidth}
        \includegraphics[width=\linewidth]{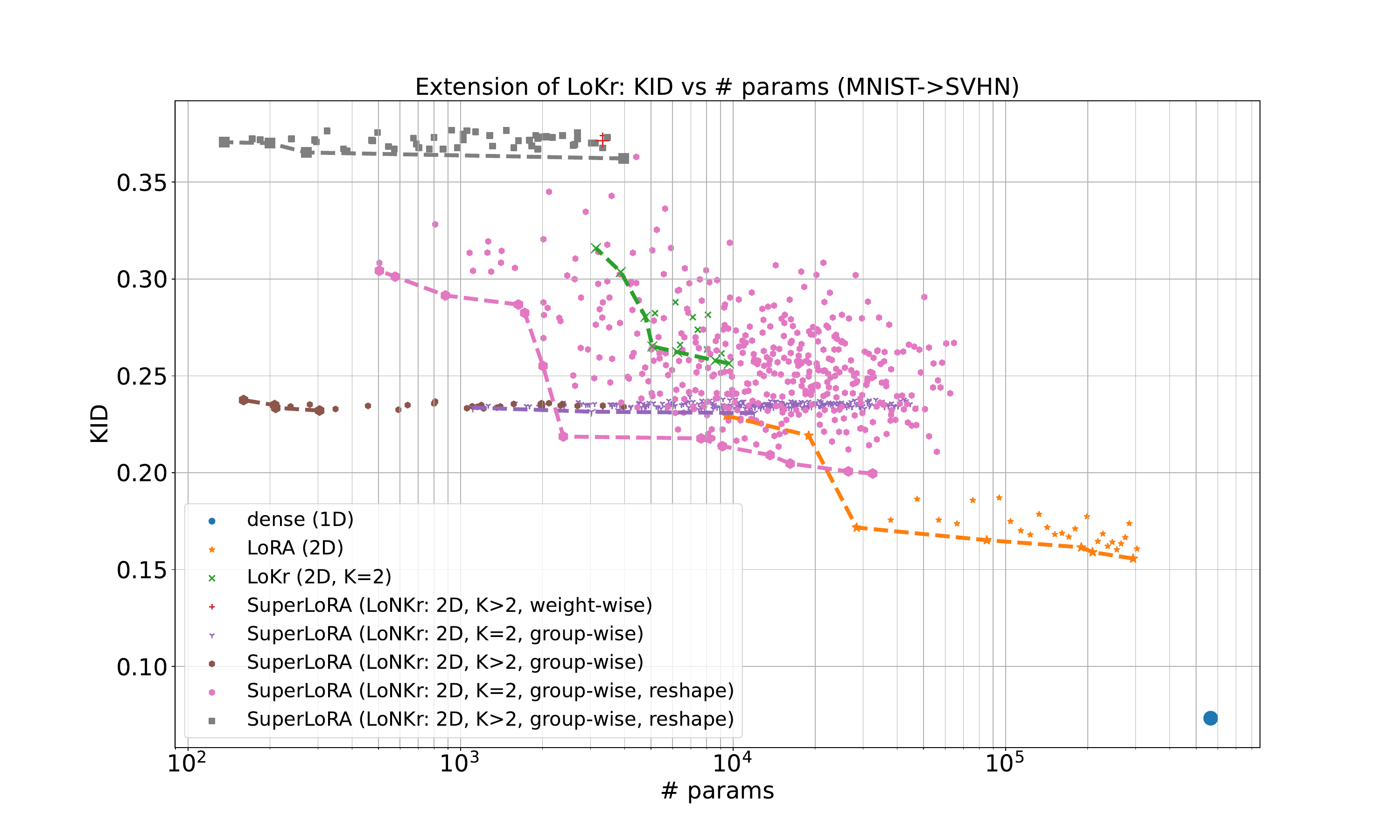}
        \caption{SuperLoRA (LoNKr, KID)}
    \end{subfigure}
    \begin{subfigure}{0.49\linewidth}
        \includegraphics[width=\linewidth]{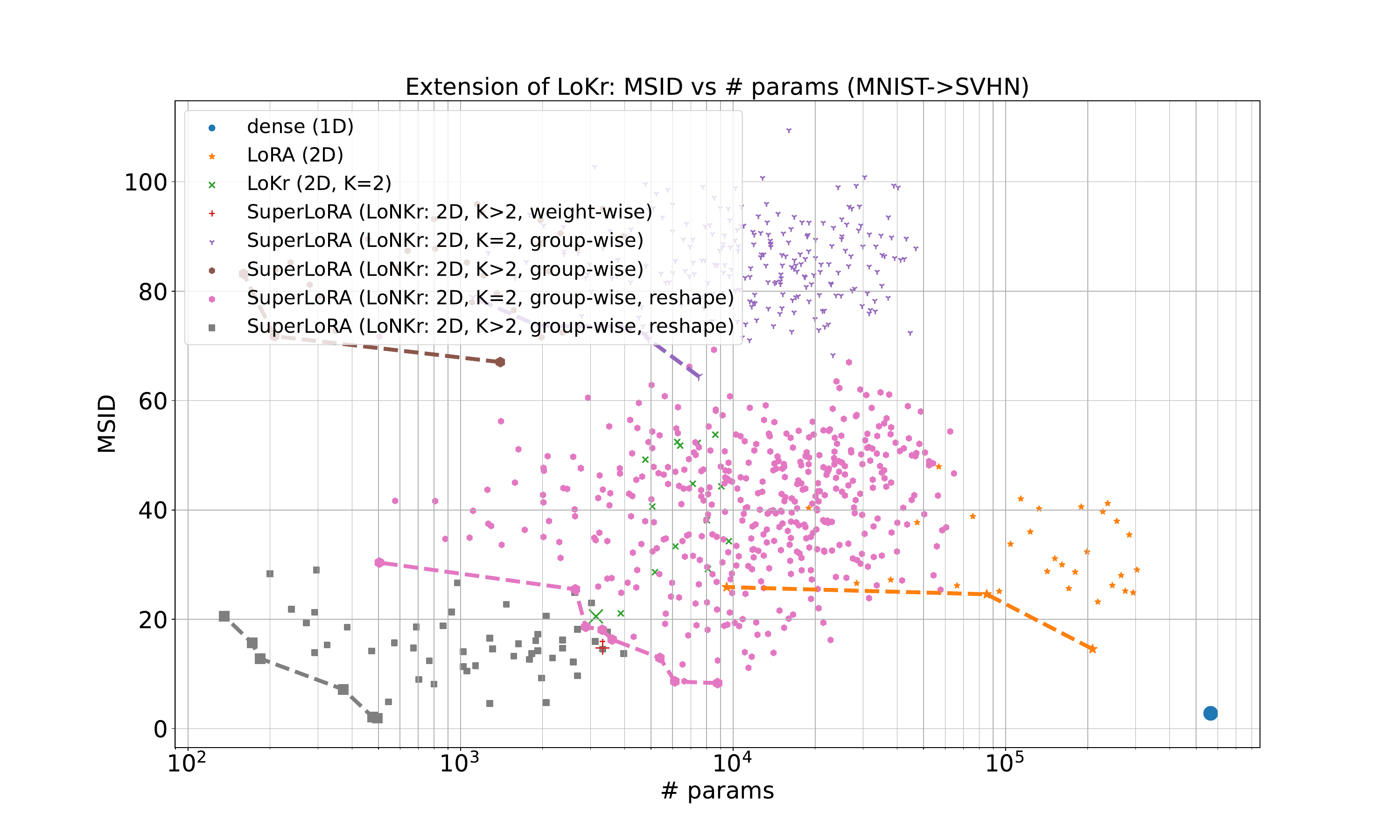}
        \caption{SuperLoRA (LoNKr, MSID)}
    \end{subfigure}    
       \begin{subfigure}{0.49\linewidth}
        \includegraphics[width=\linewidth]{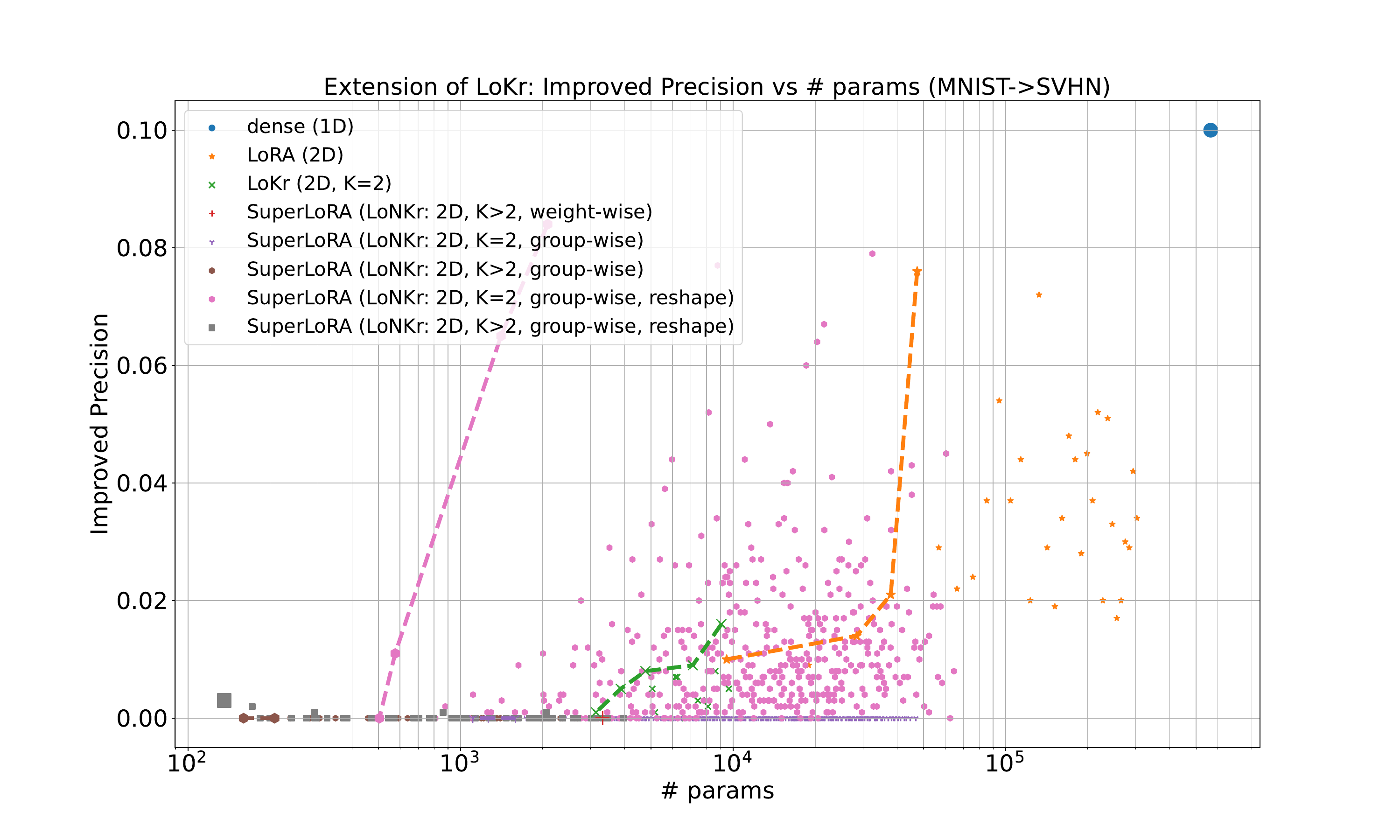}
        \caption{SuperLoRA (LoNKr, Improved Precision)}
    \end{subfigure}
    \begin{subfigure}{0.49\linewidth}
        \includegraphics[width=\linewidth]{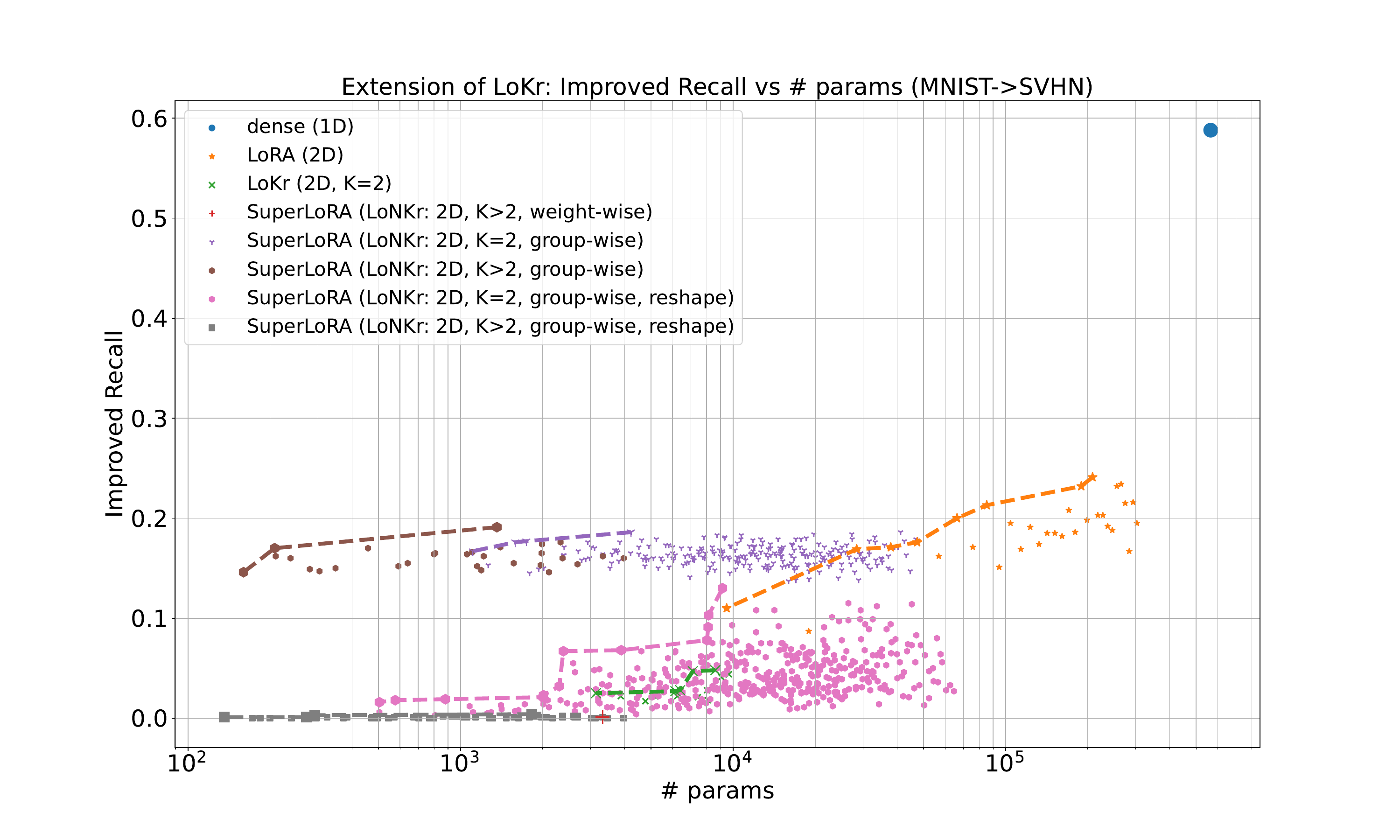}
        \caption{SuperLoRA (LoNKr, Improved Recall)}
    \end{subfigure}    
    \caption{Complete results of SuperLoRA (LoNKr) for transfer learning from MNIST to SVHN.}
    \label{fig:2svhn_lonkr}
\end{figure}

\Cref{fig:2svhn_lonkr} demonstrates the results of SuperLoRA (LoNKr). From MSID figure, we can see that, LoNKr extends LoKr to low-parameter regime, and achieves a better MSID. 
\subsubsection{SuperLoRA (LoRTA)}
\begin{figure}[t]
    \centering
    \begin{subfigure}{0.49\linewidth}
        \includegraphics[width=\linewidth]{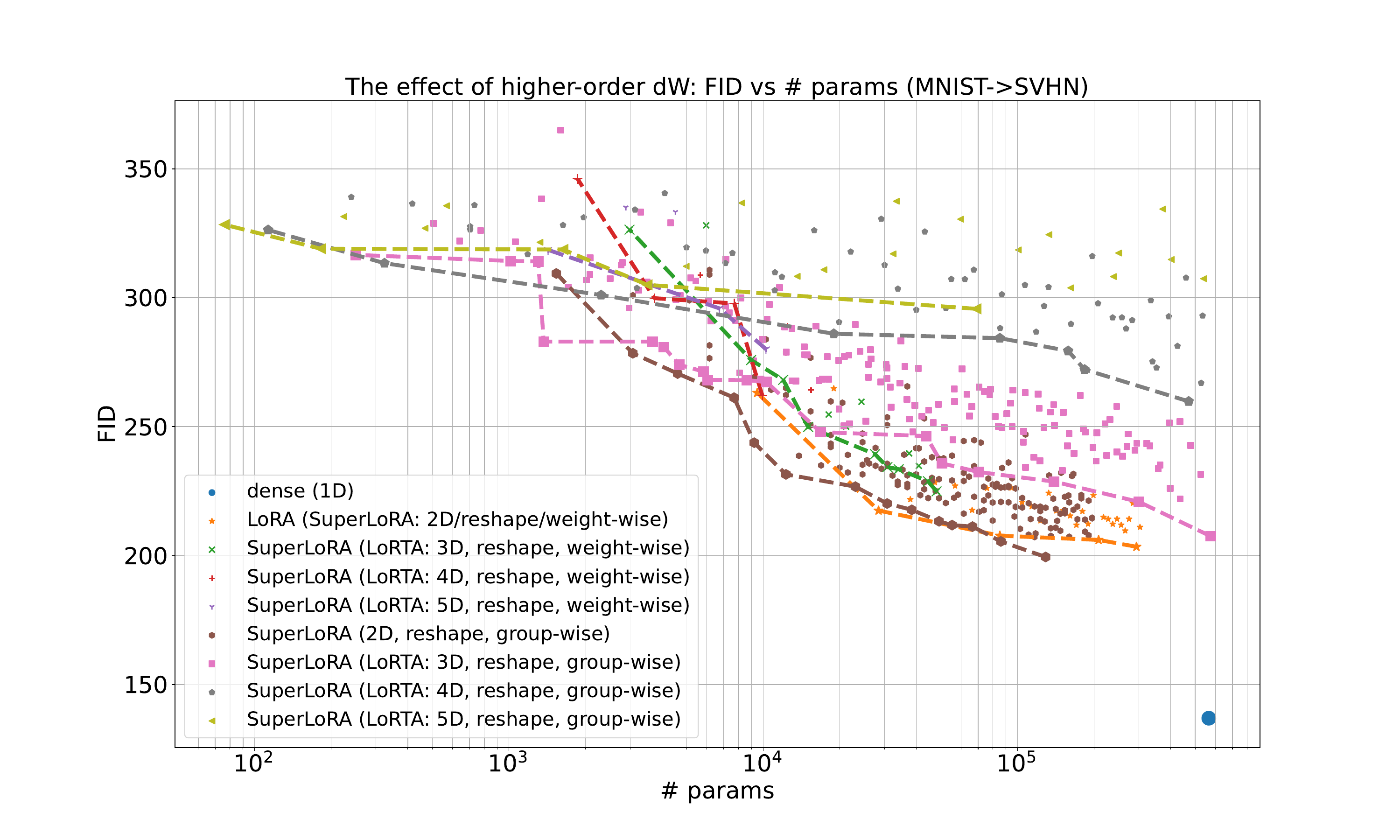}
        \caption{SuperLoRA (LoRTA, FID)}
    \end{subfigure}
    \begin{subfigure}{0.49\linewidth}
        \includegraphics[width=\linewidth]{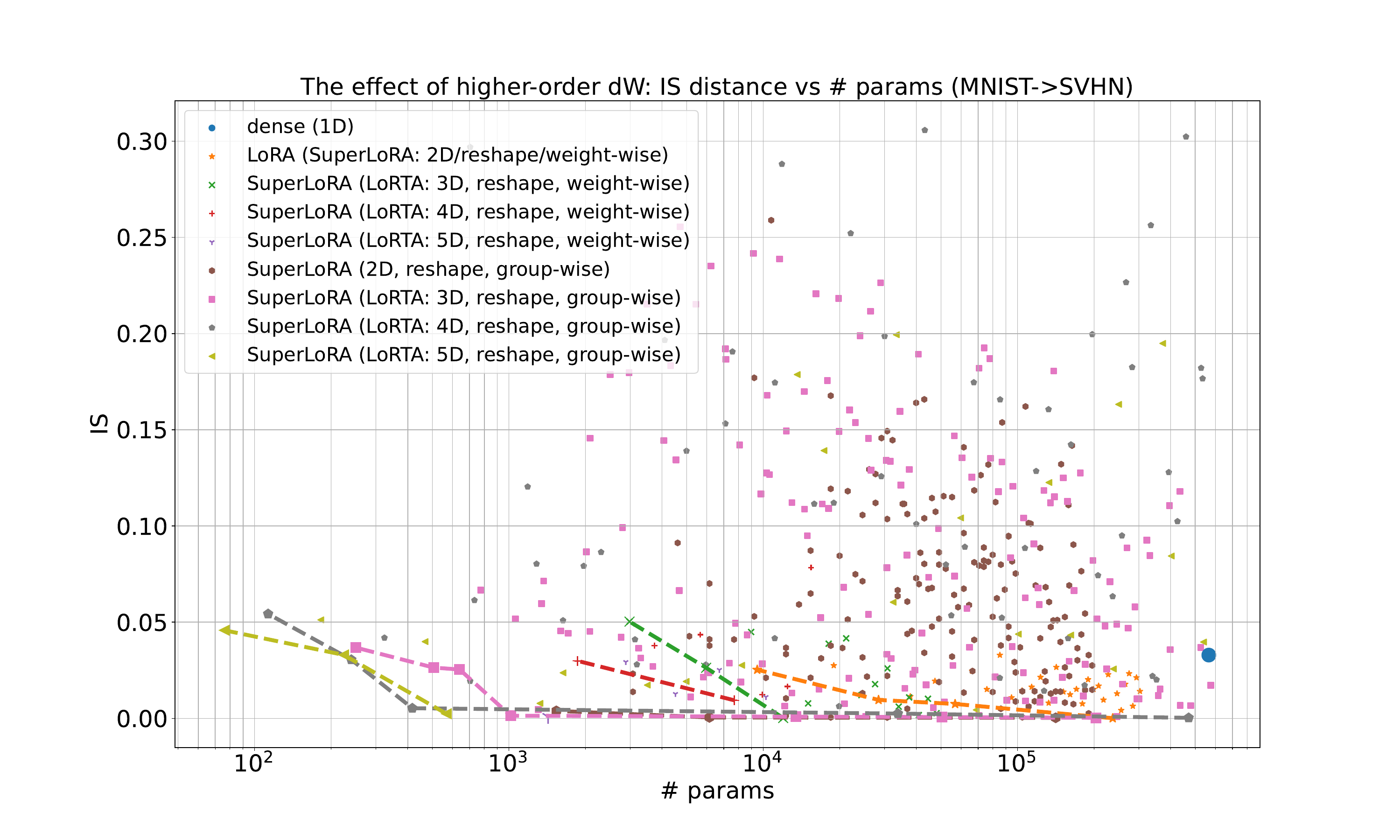}
        \caption{SuperLoRA (LoRTA, IS)}
    \end{subfigure}
   \begin{subfigure}{0.49\linewidth}
        \includegraphics[width=\linewidth]{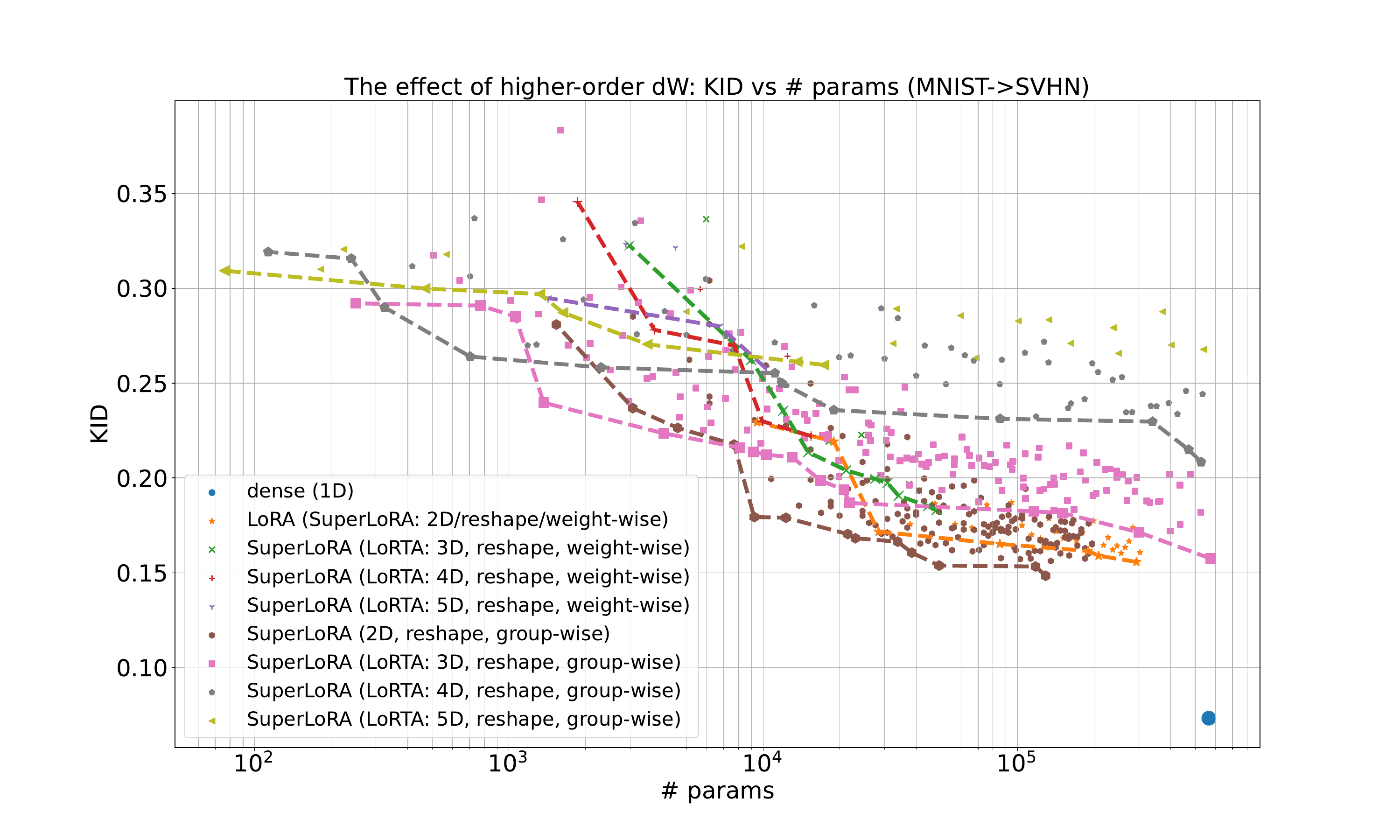}
        \caption{SuperLoRA (LoRTA, KID)}
    \end{subfigure}
    \begin{subfigure}{0.49\linewidth}
        \includegraphics[width=\linewidth]{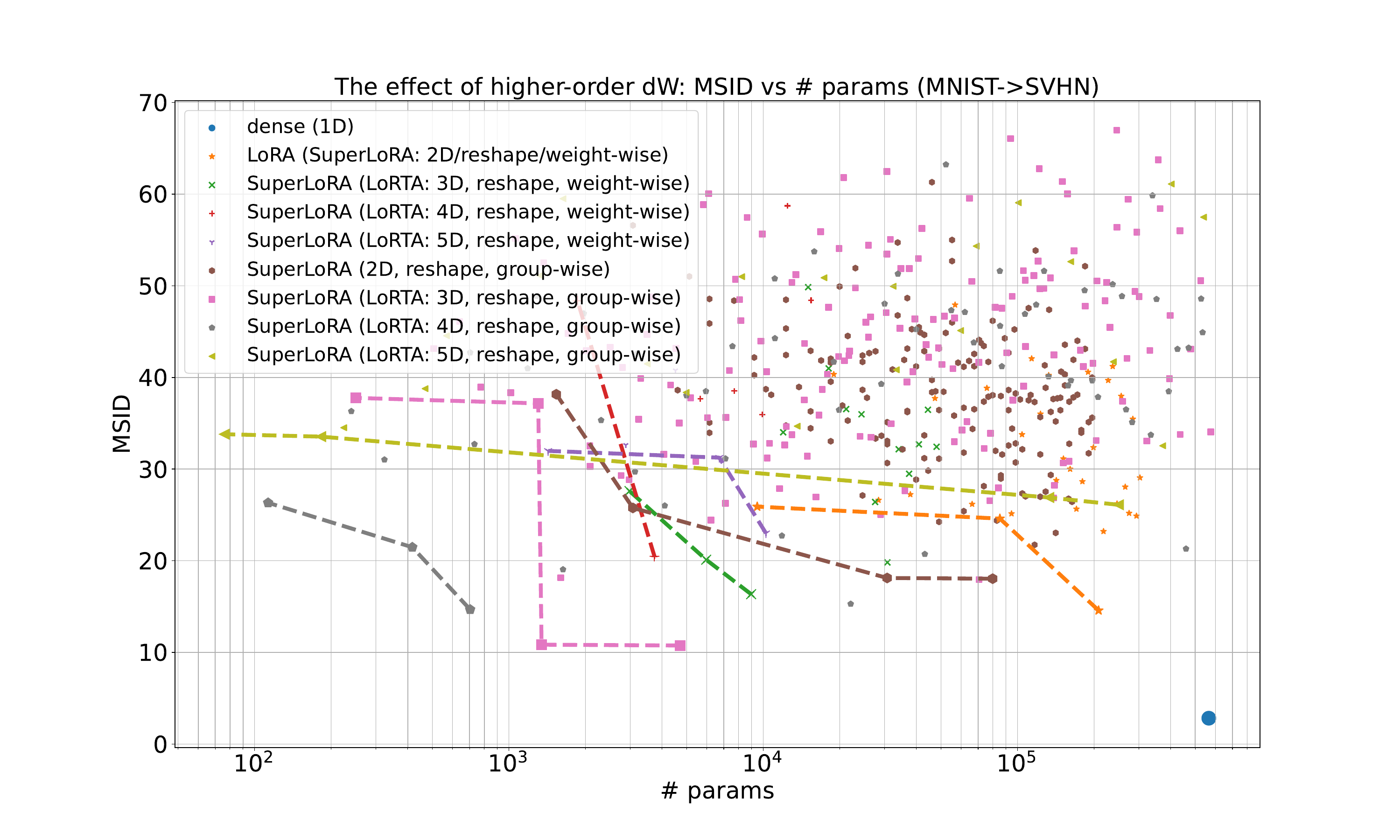}
        \caption{SuperLoRA (LoRTA, MSID)}
    \end{subfigure}    
       \begin{subfigure}{0.49\linewidth}
        \includegraphics[width=\linewidth]{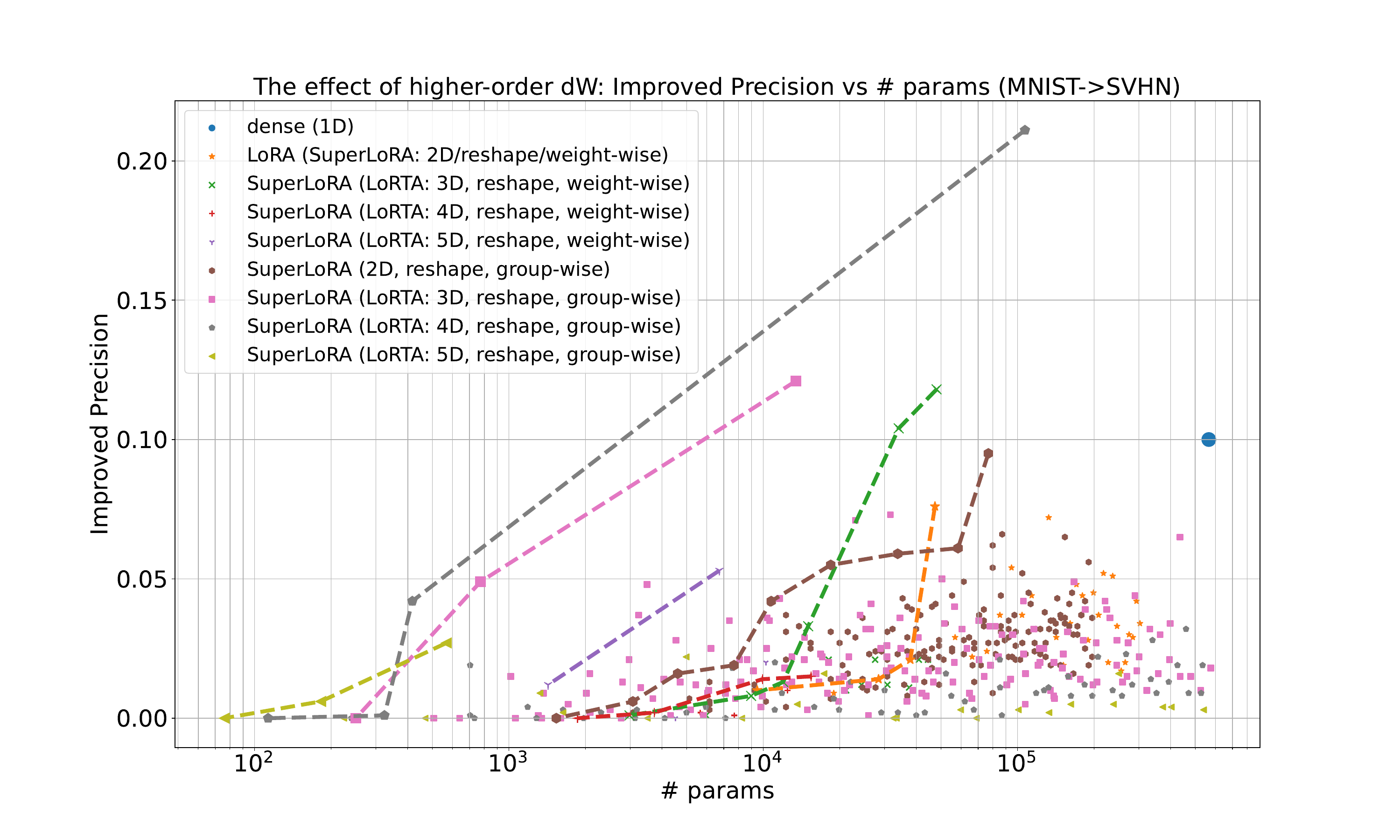}
        \caption{SuperLoRA (LoRTA, Improved Precision)}
    \end{subfigure}
    \begin{subfigure}{0.49\linewidth}
        \includegraphics[width=\linewidth]{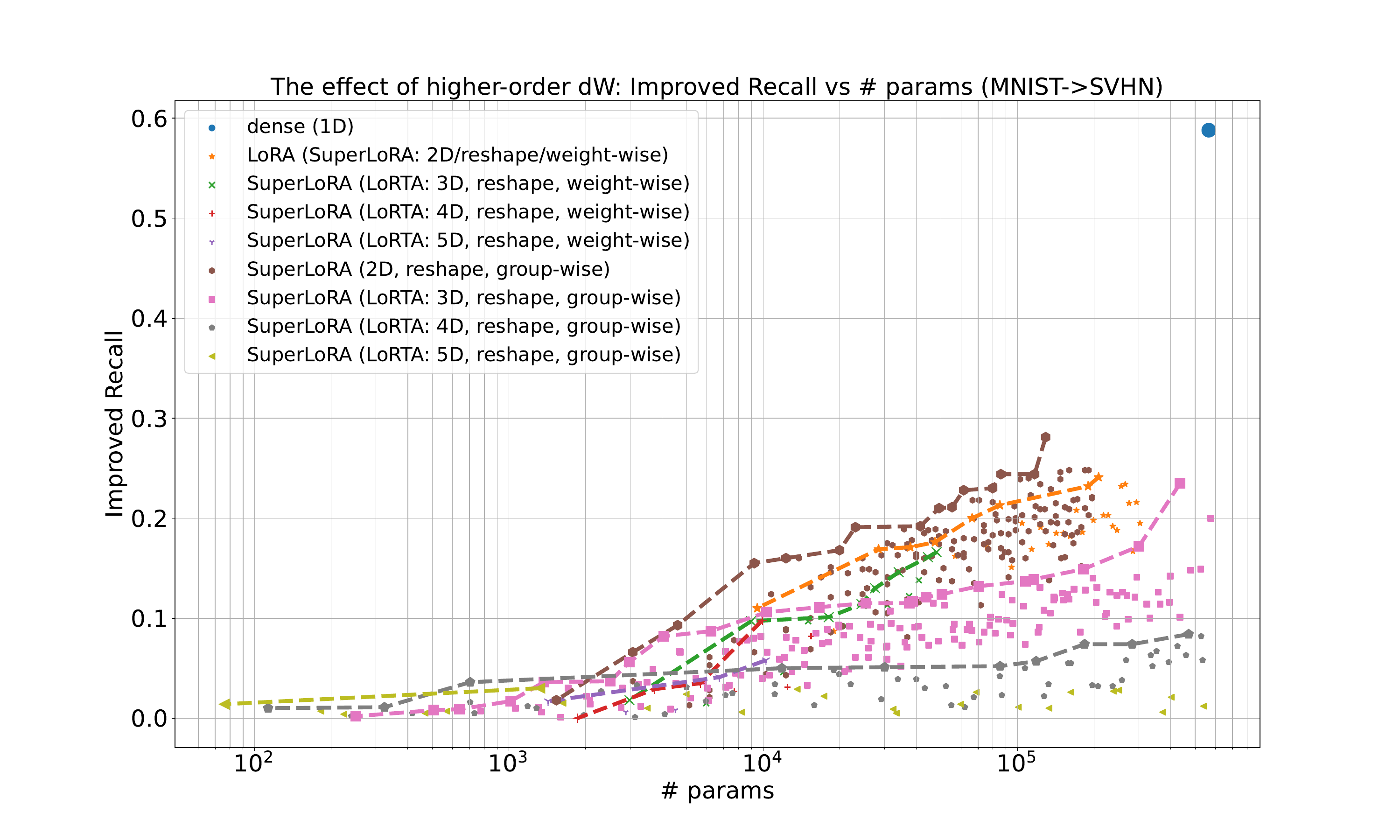}
        \caption{SuperLoRA (LoRTA, Improved Recall)}
    \end{subfigure}    
    \caption{Complete results of LoRTA for transfer learning from MNIST to SVHN.}
        \label{fig:2svhn_lorta}
\end{figure}

From FID and KID in \Cref{fig:2svhn_lorta}, LoRTA pushes required parameters from $10^4$ to $10^2$ compared with LoRA, providing more flexibility when the memory is limited.

\subsection{Effect of groups in LoNKr and LoRTA}
\begin{figure}[t]
    \centering
    \begin{subfigure}{0.49\linewidth}
        \includegraphics[width=\linewidth]{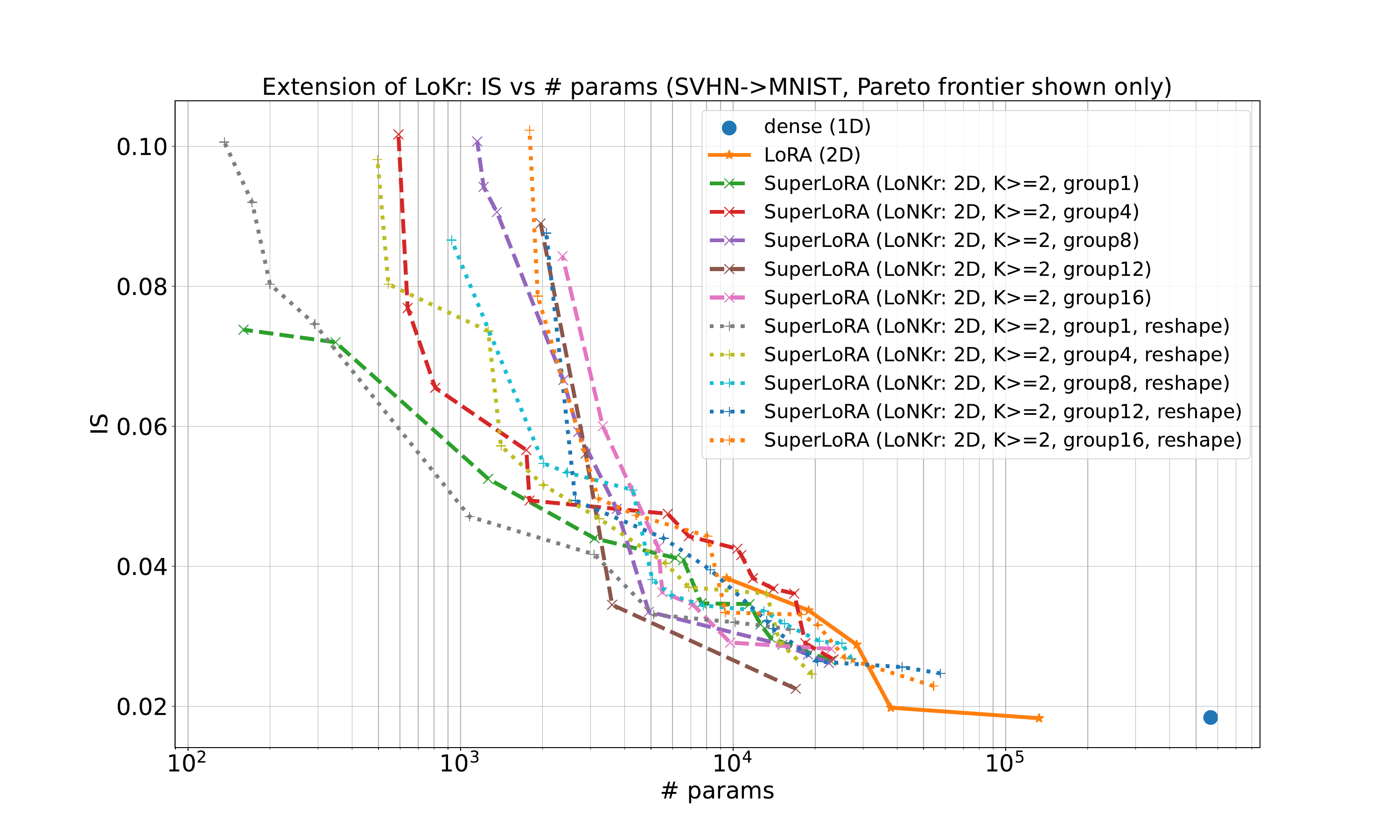}
        \caption{SuperLoRA (LoNKr, Pareto frontier only)}
    \end{subfigure}
    \begin{subfigure}{0.49\linewidth}
        \includegraphics[width=\linewidth]{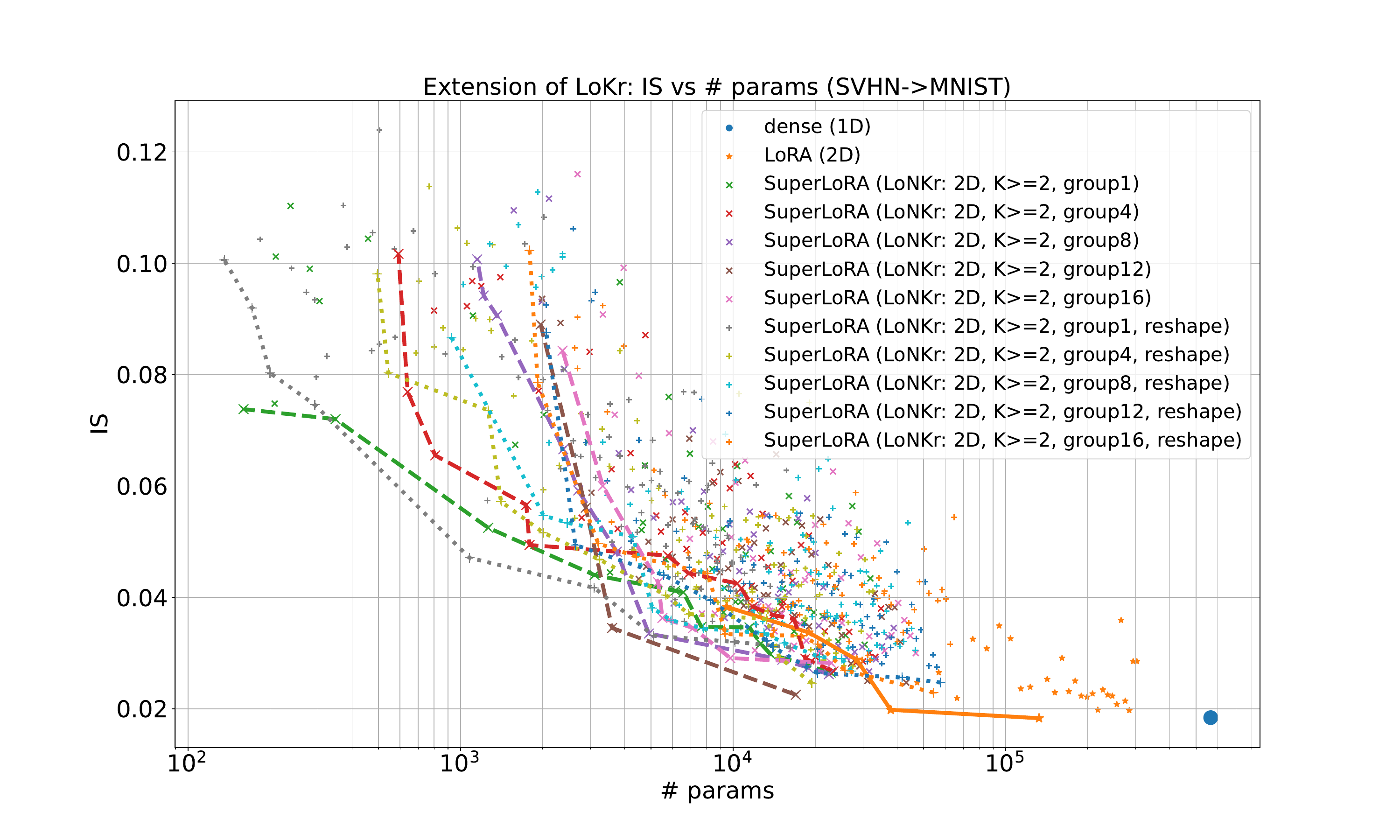}
        \caption{SuperLoRA (LoNKr)}
    \end{subfigure}
    \caption{Effect of groups in LoNKr.}
    \label{fig:lonkr_group}
\end{figure}

\begin{figure}[t]
    \centering
    \begin{subfigure}{0.49\linewidth}
        \includegraphics[width=\linewidth]{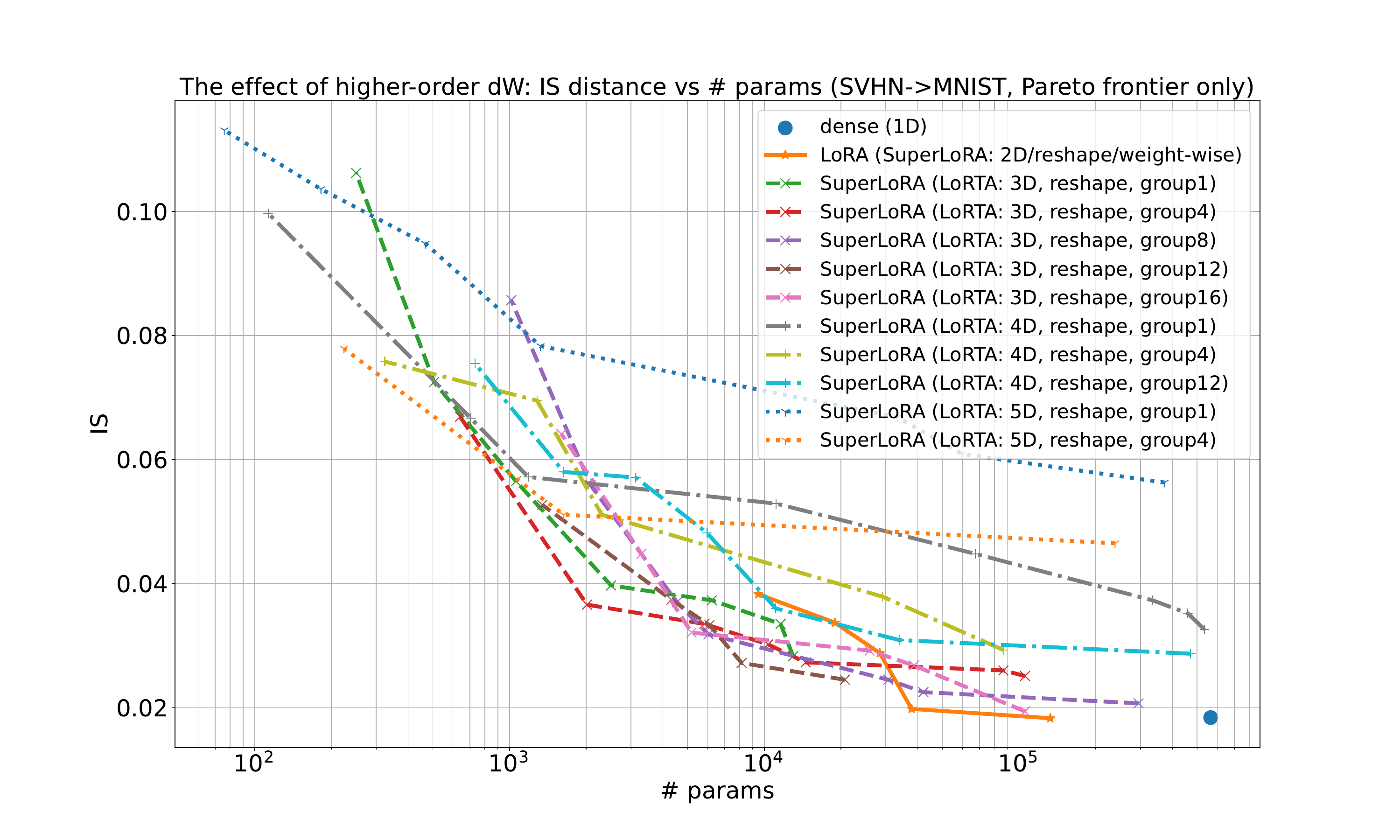}
        \caption{SuperLoRA (LoRTA, Pareto frontier only)}
    \end{subfigure}
    \begin{subfigure}{0.49\linewidth}
        \includegraphics[width=\linewidth]{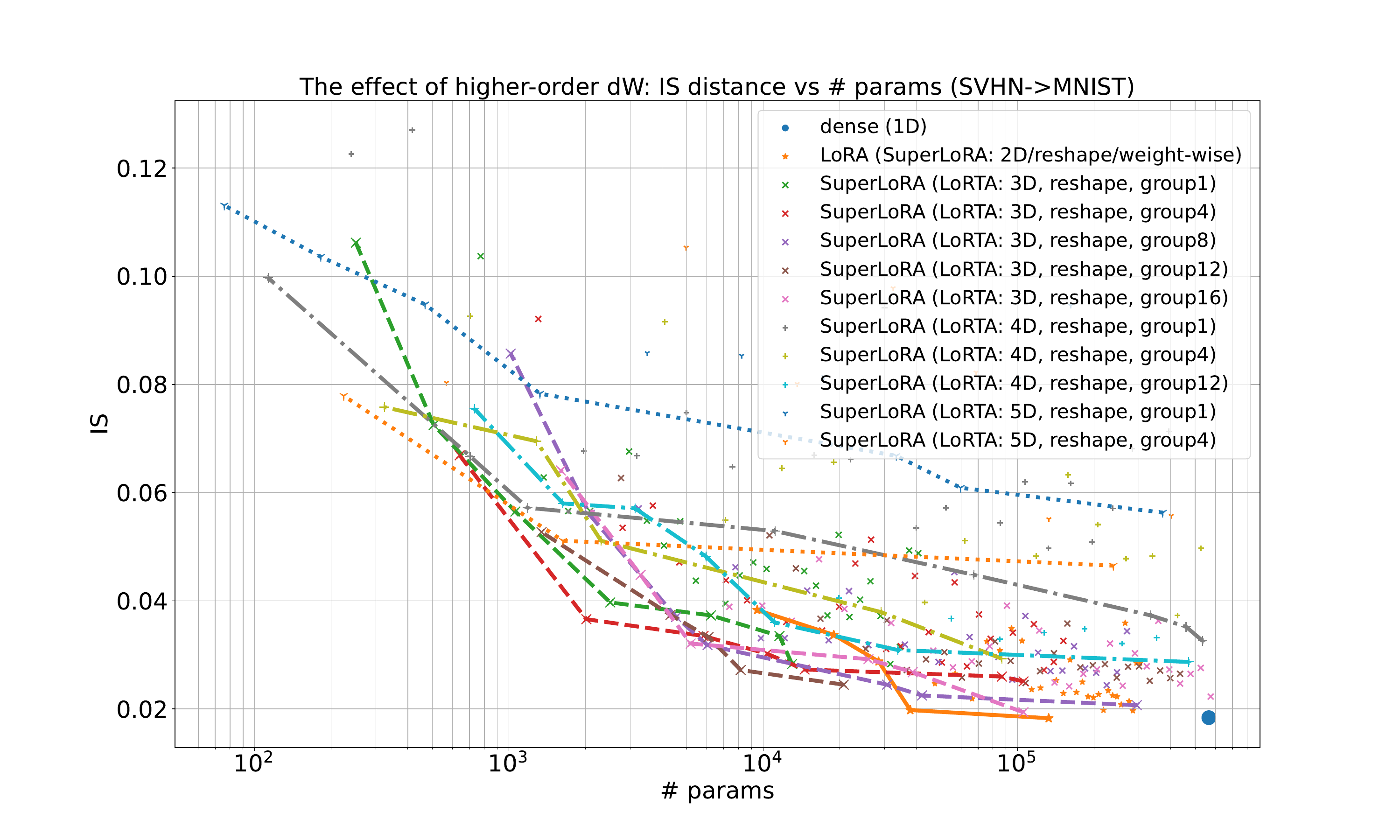}
        \caption{SuperLoRA (LoRTA)}
    \end{subfigure}
    \caption{Effect of groups in LoRTA.}
    \label{fig:lorta_group}
\end{figure}

From \Cref{fig:lonkr_group} and \Cref{fig:lorta_group}, LoNKr and LoRTA behave differently in terms of the number of groups: for LoNKr, fewer groups are better (than more groups) in low-parameter regime, while they are comparable in high-parameter regime. However, LoRTA prefers less groups.

\subsection{Effect of split $K$ in LoNKr}
\begin{figure}[t]
    \centering
    \begin{subfigure}{0.49\linewidth}
        \includegraphics[width=\linewidth]{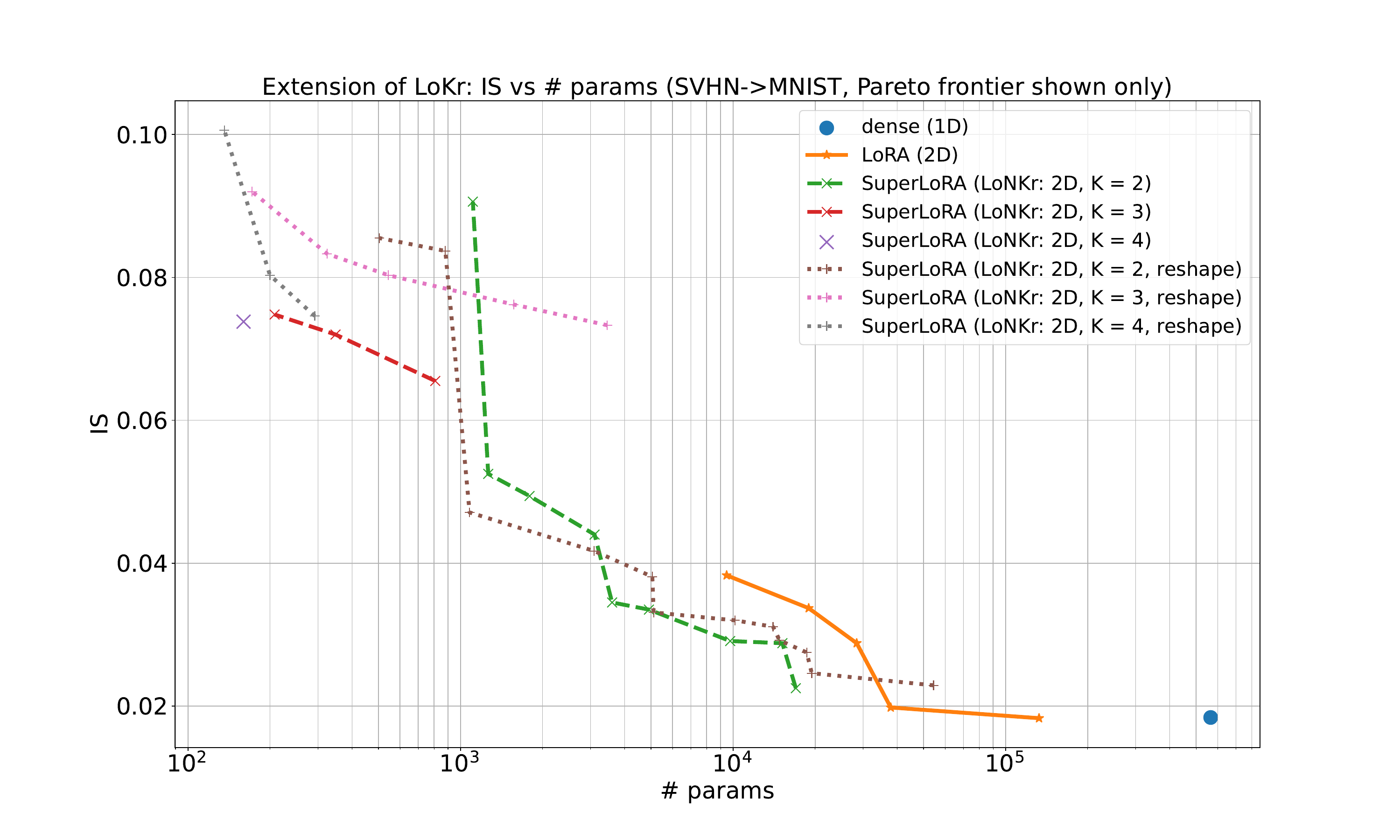}
        \caption{SuperLoRA (LoNKr, Pareto frontier only)}
    \end{subfigure}
    \begin{subfigure}{0.49\linewidth}
        \includegraphics[width=\linewidth]{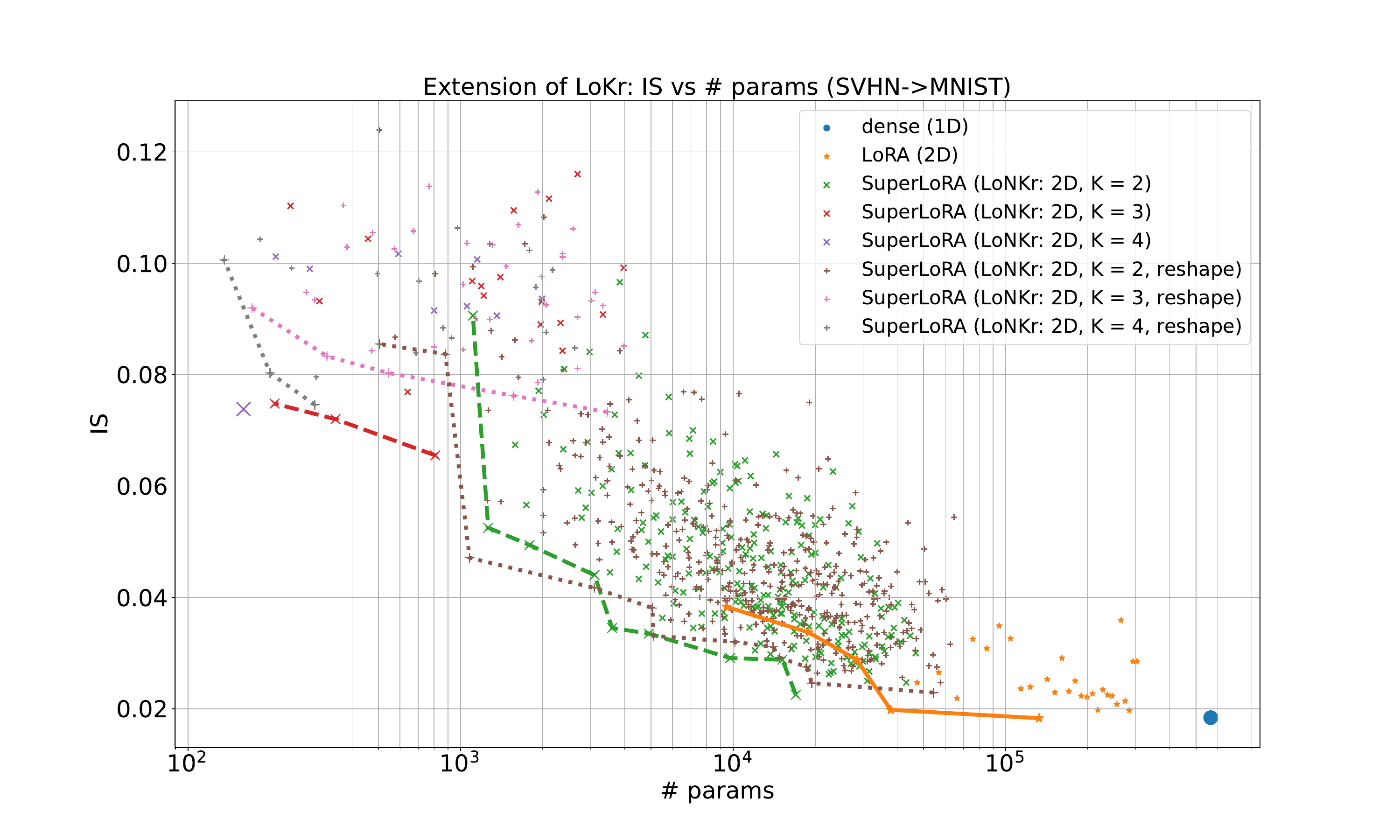}
        \caption{SuperLoRA (LoNKr)}
    \end{subfigure}
    \caption{Effect of K in LoNKr.}
    \label{fig:lonkr_split}
\end{figure}

As shown in \Cref{fig:lonkr_split}, larger $K$ works better than smaller ones in the low-parameter regime.

